\RequirePackage[l2tabu]{nag}		
\documentclass[a4paper,11pt,leqno,openbib]{memoir} 

\usepackage{url}            
\usepackage{booktabs}       
\usepackage{amsfonts}       
\usepackage{nicefrac}       
\usepackage{microtype}      
\usepackage{multirow}
\usepackage{setspace}

\usepackage{datetime}
\usepackage{ifpdf}
\ifpdf
\fi
\ifdraftdoc 
	\usepackage{draftwatermark}				
	\SetWatermarkScale{0.3}
	\SetWatermarkText{\bf Draft: \today}
\fi
\newsubfloat{figure}
\newsubfloat{table}
\settrimmedsize{297mm}{210mm}{*}
\settypeblocksize{634pt}{448.13pt}{*} 
\setulmargins{4cm}{*}{*} 
\setlrmargins{*}{*}{1} 
\setmarginnotes{17pt}{51pt}{\onelineskip} 
\setheadfoot{\onelineskip}{2\onelineskip} 
\setheaderspaces{*}{2\onelineskip}{*} 
\checkandfixthelayout
\usepackage{fouriernc}
\usepackage[T1]{fontenc}

\OnehalfSpacing 
\setsecnumdepth{subsection} 
\maxsecnumdepth{subsubsection}
\usepackage{calc,soul,fourier}
\makeatletter 
\newlength\dlf@normtxtw 
\newsavebox{\feline@chapter} 
\newcommand\feline@chapter@marker[1][4cm]{%
	\sbox\feline@chapter{%
		\resizebox{!}{#1}{\fboxsep=1pt%
			\colorbox{gray}{\color{white}\thechapter}%
		}}%
		\rotatebox{90}{%
			\resizebox{%
				\heightof{\usebox{\feline@chapter}}+\depthof{\usebox{\feline@chapter}}}%
			{!}{\scshape\so\@chapapp}}\quad%
		\raisebox{\depthof{\usebox{\feline@chapter}}}{\usebox{\feline@chapter}}%
} 
\newcommand\feline@chm[1][4cm]{%
	\sbox\feline@chapter{\feline@chapter@marker[#1]}%
	\makebox[0pt][c]{
		\makebox[1cm][r]{\usebox\feline@chapter}%
	}}
\makechapterstyle{daleifmodif}{

	\renewcommand\printchapternum{\null\hfill\feline@chm[2.5cm]\par}

} 
\makeatother 
\chapterstyle{daleifmodif}
\makepagestyle{myvf} 
\makeoddfoot{myvf}{}{\thepage}{} 
\makeevenfoot{myvf}{}{\thepage}{} 
\makeheadrule{myvf}{\textwidth}{\normalrulethickness} 
\makeevenhead{myvf}{\small\textsc{\leftmark}}{}{} 
\makeoddhead{myvf}{}{}{\small\textsc{\rightmark}}
\newcommand{\clearemptydoublepage}{\newpage{\thispagestyle{empty}\cleardoublepage}}
\makeindex
\usepackage{import}

\usepackage{lipsum}					
\usepackage{amsfonts} 					
\usepackage[centertags]{amsmath}			
\usepackage{stmaryrd}					
\usepackage{amssymb}					
\usepackage{amsthm}					
\usepackage{newlfont}					
\usepackage{layouts}					
\usepackage{graphicx}					
\usepackage{longtable,rotating}			
\usepackage[applemac]{inputenc}			
\usepackage{colortbl}					
\usepackage{wasysym}					
\usepackage{mathrsfs}					
\usepackage{float}						
\usepackage{verbatim}					
\usepackage{upgreek }					
\usepackage{latexsym}					
\usepackage[square,numbers,
		     sort&compress]{natbib}		
\usepackage{url}						
\usepackage{etex}						
\usepackage{fixltx2e}					
\usepackage[spanish,english]{babel}		
\usepackage{color}                    				
\usepackage[colorlinks=true,
		     allcolors=black]{hyperref}              
\usepackage{memhfixc}					
\usepackage{enumerate}					
\usepackage{footnote}					
\usepackage{microtype}					
\usepackage{rotfloat}					
\usepackage{alltt}						
\usepackage[version=0.96]{pgf}			
\usepackage{tikz}						
\usetikzlibrary{arrows,shapes,snakes,
		       automata,backgrounds,
		       petri,topaths}				
\newcommand{\pgftextcircled}[1]{                                                                    
    \setbox0=\hbox{#1}%
    \dimen0\wd0%
    \divide\dimen0 by 2%
    \begin{tikzpicture}[baseline=(a.base)]%
        \useasboundingbox (-\the\dimen0,0pt) rectangle (\the\dimen0,1pt);
        \node[circle,draw,outer sep=0pt,inner sep=0.1ex] (a) {#1};
    \end{tikzpicture}
}
\newcommand{\blackged}{\hfill$\blacksquare$}
\newcommand{\whiteged}{\hfill$\square$}
\newcounter{proofcount}
\renewenvironment{proof}[1][\proofname.]{\par
 \ifnum \theproofcount>0 \pushQED{\whiteged} \else \pushQED{\blackged} \fi%
 \refstepcounter{proofcount}
 \normalfont 
 \trivlist
 \item[\hskip\labelsep
       \itshape
   {\bf\em #1}]\ignorespaces
}{%
 \addtocounter{proofcount}{-1}
 \popQED\endtrivlist
}

\let\oldsqrt\sqrt
\def\sqrt{\mathpalette\DHLhksqrt}
\def\DHLhksqrt#1#2{%
\setbox0=\hbox{$#1\oldsqrt{#2\,}$}\dimen0=\ht0
\advance\dimen0-0.2\ht0
\setbox2=\hbox{\vrule height\ht0 depth -\dimen0}%
{\box0\lower0.4pt\box2}}
\newcommand{\mycaption}[2][\@empty]{
	\captionnamefont{\scshape} 
	\changecaptionwidth
	\captionwidth{0.9\linewidth}
	\captiondelim{.\:} 
	\indentcaption{0.75cm}
	\captionstyle[\centering]{}
	\setlength{\belowcaptionskip}{10pt}
	\ifx \@empty#1 \caption{#2}\else \caption[#1]{#2}
}
\newcommand{\mysubcaption}[2][\@empty]{
	\subcaptionsize{\small}
	\hangsubcaption
	\subcaptionlabelfont{\rmfamily}
	\sidecapstyle{\raggedright}
	\setlength{\belowcaptionskip}{10pt}
	\ifx \@empty#1 \subcaption{#2}\else \subcaption[#1]{#2}
}
\usepackage{lettrine}
\newcommand{\initial}[1]{%
	\lettrine[lines=3,lhang=0.33,nindent=0em]{
		\color{gray}
     		{\textsc{#1}}}{}}
            
\newcommand{\mycomment}[1]  {}
\theoremstyle{plain}

\theoremstyle{plain}

\theoremstyle{plain}
\theoremstyle{definition}
\newtheorem{dfn}{Definition}[chapter]
\theoremstyle{plain}

\theoremstyle{plain}

\theoremstyle{plain}

\usepackage{caption}
\usepackage{amsmath}
\usepackage{mathtools}
\usepackage{enumitem}
\usepackage{amsmath}

\newtheorem{conjecture}{Conjecture}
\newtheorem{lemma}{Lemma}
\newtheorem{theorem}{Theorem}
\newtheorem{claim}{Claim}

\def\be{\begin{equation}}
\def\ee{\end{equation}}
\def\beas{\begin{align*}}
\def\eeas{\end{align*}}
\def\bea{\begin{align}}
\def\eea{\end{align}}

\newcommand{\h}{{\mathbf h}}
\newcommand{\x}{{\mathbf x}}
\newcommand{\y}{{\mathbf y}}

\newcommand{\uu}{{\mathbf u}}
\newcommand{\vv}{{\mathbf v}}
\newcommand{\w}{{\mathbf w}}

\newcommand{\e}{{\mathbf e}}
\newcommand{\aaa}{{\mathbf a}}
\newcommand{\bb}{{\mathbf b}}

\newcommand{\dd}{{\mathbf d}}
\newcommand{\ff}{{\mathbf f}}

\newcommand{\p}{{\mathbf p}}
\newcommand{\q}{{\mathbf q}}

\newcommand{\1}{{\mathbf 1}}

\newcommand{\A}{{\mathcal A}}
\newcommand{\B}{{\mathcal B}}

\newcommand{\X}{{\mathcal X}}

\newcommand{\R}{{\mathbb R}}
\newcommand{\N}{{\mathbb N}}

\newcommand{\abs}[1]{\left\lvert#1 \right\rvert}
\newcommand{\norm}[1]{\left\|#1 \right\|}

\newcommand{\inprod}[2]  {\left\langle{#1},{#2}\right\rangle}

\DeclareMathOperator*{\argmax}{argmax}

\newcommand{\mat}[1]{\llbracket#1\rrbracket}
\newcommand{\matflex}[1]{\left\llbracket#1\right\rrbracket}
\newcommand{\sep}[2]{\mathrm{sep}_{(#1)}\left( #2 \right)}
\newcommand{\rank}[1]{\mathrm{rank}\left( #1 \right)}
\newcommand{\state}[2]{\mathrm{states}\left( #1, #2 \right)}
\newcommand{\trajectory}[1]{\mathrm{trajectory}\left( #1 \right)}
\newcommand{\WInone}{W^{\mathrm{I}}}
\newcommand{\WHnone}{W^{\mathrm{H}}}
\newcommand{\WI}[1]{
	W^{\mathrm{I}, #1}
}
\newcommand{\WH}[1]{
	W^{\mathrm{H}, #1}
}
\newcommand{\WO}{
	W^{\mathrm{O}}
}

\def\multiset#1#2{\ensuremath{\left(\kern-.3em\left(\genfrac{}{}{0pt}{}{#1}{#2}\right)\kern-.3em\right)}}

\newcommand{\eg}{\emph{e.g.}}
\newcommand{\ie}{\emph{i.e.}}

\newcommand{\wrt}{w.r.t.}

\begin{document}
%
%
%
%
%
\frontmatter
\pagenumbering{roman}
%
%
%
%
%
%
\begin{titlingpage}
\begin{SingleSpace}
\calccentering{\unitlength} 
\begin{adjustwidth*}{\unitlength}{-\unitlength}
\vspace*{13mm}
\begin{center}
\rule[0.5ex]{\linewidth}{2pt}\vspace*{-\baselineskip}\vspace*{3.2pt}
\rule[0.5ex]{\linewidth}{1pt}\\[\baselineskip]
{\HUGE Depth Enables Long-Term Memory}\\[\baselineskip]
{\HUGE for Recurrent Neural Networks}\\[\baselineskip]
\rule[0.5ex]{\linewidth}{1pt}\vspace*{-\baselineskip}\vspace{3.2pt}
\rule[0.5ex]{\linewidth}{2pt}\\
\vspace{6.5mm}
{\large By}\\
\vspace{6.5mm}
{\large\textsc{Alon Ziv}}\\
\vspace{11mm}
{\large\textsc{Supervisor: Prof. Amnon Shashua}}\\
\vspace{11mm}
~~~~~~\includegraphics[scale=0.2]{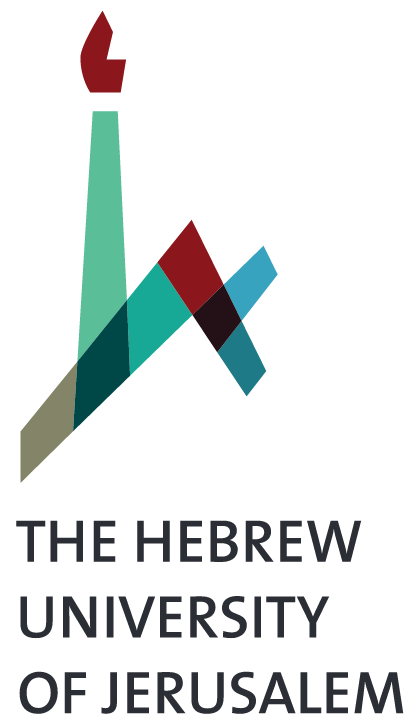}\\
\vspace{6mm}
{\large Faculty of Computer Science and Engineering\\
\textsc{The Hebrew University of Jerusalem}}\\
\vspace{11mm}
\begin{minipage}{10cm}
A dissertation submitted to the Hebrew University of Jerusalem as a partial fulfillment of the requirements of the degree of \textsc{Master of Science} in the Faculty of Computer Science and Engineering.
\end{minipage}\\
\vspace{9mm}
{\large\textsc{September 2019}}
\vspace{12mm}
\end{center}
\begin{flushright}
\end{flushright}
\end{adjustwidth*}
\end{SingleSpace}
\end{titlingpage}
\clearemptydoublepage
%
%
%
%

\chapter*{Abstract}
\initial{A} key attribute that drives the unprecedented success of modern Recurrent Neural Networks (RNNs) on learning tasks which involve sequential data, is their ability to model intricate long-term temporal dependencies. 
However, a well established measure of RNNs long-term memory capacity is lacking, and thus formal understanding of the effect of depth on their ability to correlate data throughout time is limited.
Specifically, existing depth efficiency results on convolutional networks do not suffice in order to account for the success of deep RNNs on data of varying lengths. 
In order to address this, we introduce a measure of the network's ability to support information flow across time, referred to as the \emph{Start-End separation rank}, which reflects the distance of the
function realized by the recurrent network from modeling no dependency between the beginning and end of the input sequence. 
We prove that deep recurrent networks support Start-End separation ranks which are combinatorially higher than those supported by their shallow counterparts.
Thus, we establish that depth brings forth
an overwhelming advantage in the ability of recurrent networks to model
long-term dependencies, and provide an exemplar of quantifying this key attribute which may be readily extended to other RNN architectures of interest, \eg~variants of LSTM networks. We obtain our results by considering a class of
recurrent networks referred to as Recurrent Arithmetic Circuits,
which merge the hidden state with the input via the Multiplicative Integration
operation.
We empirically demonstrate the discussed phenomena on common RNNs through extensive experimental evaluation using the optimization technique of restricting the hidden-to-hidden matrix to being orthogonal.
Finally, we employ the tool of quantum Tensor
Networks to gain additional graphic insights regarding the complexity brought
forth by depth in recurrent networks.

\clearpage
\clearemptydoublepage
%
%
%

\chapter*{Dedication and acknowledgements}
\begin{SingleSpace}
\ \ 

I would like to thank my supervisor, Professor Amnon Shashua, for his insights and support throughout my research. A special thank for Or Sharir and Yoav Levine who guided me into the fascinating world of Recurrent Neural Networks and Deep Learning in general. Finally, I want to thank Noam Wies for very useful discussions along the way.
\end{SingleSpace}
\clearpage
\clearemptydoublepage
\renewcommand{\contentsname}{Table of Contents}
\maxtocdepth{subsection}
\tableofcontents*
\addtocontents{toc}{\par\nobreak \mbox{}\hfill{\bf Page}\par\nobreak}
\clearemptydoublepage
\listoftables
\addtocontents{lot}{\par\nobreak\textbf{{\scshape Table} \hfill Page}\par\nobreak}
\clearemptydoublepage
\listoffigures
\addtocontents{lof}{\par\nobreak\textbf{{\scshape Figure} \hfill Page}\par\nobreak}
\clearemptydoublepage
%
%
\mainmatter
%
%
%
\let\textcircled=\pgftextcircled
\chapter{Introduction}
\label{chapter:intro}

\textit{This document is an extension of \citep{RAC_paper}, a joint work with Yoav Levine, Or Sharir and Amnon Shashua.}
\initial {O}ver the past few years, Recurrent  Neural Networks (RNNs) have become a prominent machine learning architectures for modeling sequential data, having
been successfully employed for language modeling~\citep{sutskever2011generating,pascanu2013difficulty,graves2013generating}, neural
machine translation~\citep{bahdanau2014neural}, online handwritten
recognition~\citep{graves2009novel}, speech recognition~\citep{graves2013speech,amodei2016deep}, and more. 
The success of recurrent networks in learning complex
functional dependencies for sequences of varying lengths, readily implies that
long-term and elaborate dependencies in the given inputs are somehow supported
by these networks. Though connectivity contribution to performance of RNNs has been empirically investigated \citep{zhang2016architectural}, formal understanding of the influence of a recurrent
network's structure on its expressiveness, and specifically on its
ever-improving ability to integrate data throughout time (\eg~translating long
sentences, answering elaborate questions), is lacking.

An ongoing empirical effort to successfully apply recurrent networks to tasks of
increasing complexity and temporal extent, includes augmentations of the
recurrent unit such as Long Short Term Memory (LSTM)
networks~\citep{hochreiter1997long} and their variants (\eg~\citep{gers2000recurrent,cho2014learning}). A parallel avenue, which we focus
on in this work, includes the stacking of layers to form deep recurrent
networks~\citep{schmidhuber1992learning}.
Deep recurrent networks, which exhibit empirical superiority over shallow
ones (see \eg~\citep{graves2013speech}),
implement hierarchical processing of information at every time-step that
accompanies their inherent time-advancing computation. Evidence for a time-scale related effect arises from experiments \citep{hermans2013training}~--~deep recurrent networks appear to model dependencies which correspond to longer time-scales than shallow
ones.
These findings, which imply that depth brings forth a considerable advantage both in complexity and in temporal capacity of recurrent networks, have no
adequate theoretical explanation.

In this work, we theoretically address the above presented issues. Based on the relative
maturity of \emph{depth efficiency} results in neural networks, namely results
that show that deep networks efficiently express functions that would require
shallow ones to have a super-linear size (see \eg~\citep{cohen2016expressive,eldan2016power,telgarsky2015representation}), it
is natural to assume that depth has a similar effect on the expressiveness of
recurrent networks. Indeed, we show that depth efficiency holds for recurrent
networks.

However, the distinguishing attribute of recurrent networks,  is their inherent
ability to cope with varying input sequence length. Thus, once
establishing the above depth efficiency in recurrent networks, a basic question
arises, which relates to the apparent depth enhanced long-term memory in
recurrent networks: {\sl Do the functions which are efficiently expressed by
	deep recurrent networks correspond to dependencies over longer time-scales?}
We answer this question affirmatively, by showing that depth provides a super-linear (combinatorial) boost to
the ability of recurrent networks to model long-term dependencies in their inputs.

This document is divided into three chapters and a series of appendices. In the remainder of this chapter we give the relevant background required for understanding our main results. More specifically, in Section \ref{sec:intro:rnns}~we present the concept of Recurrent Neural Networks (RNNs) and discuss how deep RNNs are constructed. In Section \ref{sec:intro:orthRNNs}~we present Orthogonal RNNs - an optimization technique that plays a major role in the experimental evaluation of Chapter \ref{chap:experiments}, and in Section \ref{sec:intro:tensors}~we go over a few basic concepts in tensor analysis that will be used extensively in the theoretical proofs of Chapter \ref{chapter:theory}.

In Chapter \ref{chapter:theory} we theoretically analyze the effect of multi-layered connectivity on the long-term memory capacity of recurrent networks, from a functional perspective. Our main result, given in Theorem \ref{theorem:main_result}, states that \emph{depth-2} RNNs are combinatorially more expressive than \emph{depth-1} RNNs in their ability to model long-term dependencies in the data. The theorem is proved using a family of recurrent networks named Recurrent Arithmetic Circuits (to be presented in Section \ref{sec:theory:RAC}), a surrogate model that has the exact connectivity of conventional RNNs and differs only in the non-linear operation.

Finally, in Chapter~\ref{chap:experiments} we present numerical evaluations which support the above theoretical findings. Specifically, we perform two experiments that directly test the ability of recurrent networks to model complex long-term temporal dependencies. 
Our results exhibit a clear boost in memory capacity of deeper recurrent networks relative to shallower networks that are given the same amount of resources, and thus directly demonstrate the theoretical trends established in this work.

In Appendix \ref{app:rac_tns}~we present the concept of tensor networks (TNs) and use it to graphically analyze the computation performed by Recurrent Arithmetic Circuits. Apart from taking an integral part in the proof of Theorem \ref{theorem:main_result}, TNs are the central tool used as a basis for Conjecture \ref{conjecture:high_L}~given at the end of Chapter \ref{chapter:theory}, making this appendix crucial for a reader interested in a deeper understanding of our work. 

In Appendix \ref{app:proofs:main_result}~we extend the proof sketch given in Chapter \ref{chapter:theory}~into a formal proof of Theorem \ref{theorem:main_result}~. The main flow of the proof is given, as well as a proof for several technical lemmas and claims that was used, ranging from tensorial analysis through measure theory to combinatorics. 

In Appendix \ref{app:natural_data_experiments}~we present additional experiments that demonstrate the generalization capabilities of deep RNNs on natural data sets. This somehow complements the synthetic experiments given in Chapter \ref{chap:experiments}~which deals with functional expressiveness, and doesn't discuss generalization.  
\section{Recurrent Neural Networks}\label{sec:intro:rnns}
Recurrent Neural Networks (RNNs) are a family of neural networks designed to process sequential data, such as text or audio. The architecture of a recurrent network allows it to function as a discrete-time dynamical system. The hidden state of the system, typically represented as a vector, changes in time according to new observations while taking into account past events. The system emits an output in each time-step, calculated by a linear transformation of the state. 

\begin{figure}
	\centering
	\includegraphics[width=\linewidth]{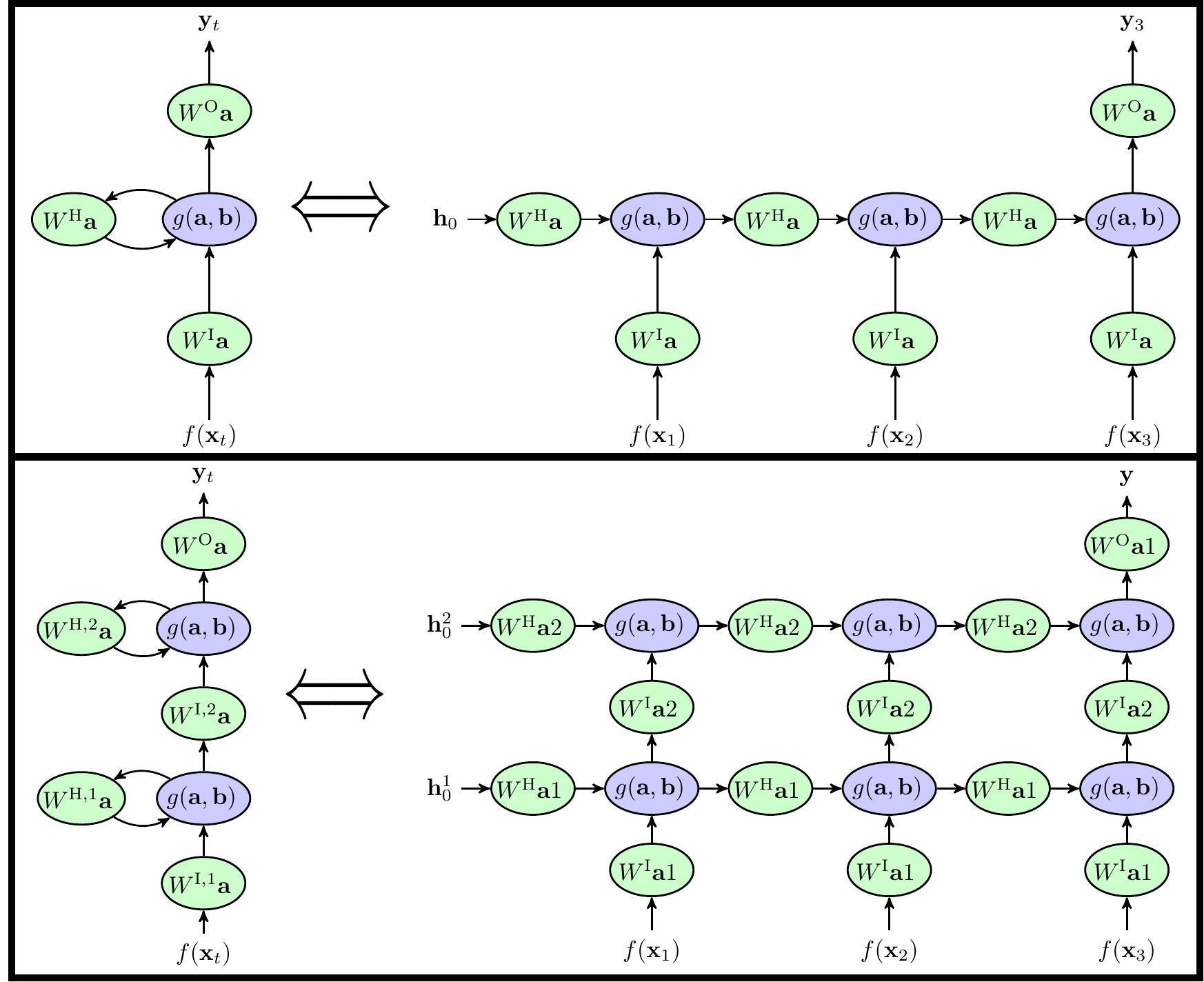}
	\ifdefined\SQUEEZE \vspace{-6mm} \fi
	\caption{Shallow and deep recurrent networks, as described by Equations~\eqref{eq:shallow_rn} and~\eqref{eq:deep_rn}, respectively.}\ifdefined\SQUEEZE \vspace{-6mm} \fi
	
	\label{fig:intro_recurrent_net}
\end{figure}

We formally present below the basic framework of recurrent networks
(top of Figure~\ref{fig:intro_recurrent_net}), which
describes both the common RNNs and the newly introduced RACs, to be discussed in Chapter \ref{chapter:theory}. We focus on the setting of a sequence to sequence classification task into one of the
categories $\{1,...,C\}\equiv[C]$. Denoting the temporal dependence by $t$,
the sequential input to the network is $\{\x^t\in \X\}_{t=1}^T$, and the output is a sequence of class scores
vectors $\{\y^{t,\Theta}\in\R^C\}_{t=1}^T$, where $\Theta$
denotes the parameters of the recurrent network, and $T$ represents the extent of the sequence in time-steps. We assume the input lies in some input space $\X$ that may be discrete (e.g. text data) or continuous (e.g. audio data), and that some initial mapping $\ff:\X\rightarrow\R^M$ is preformed on the input, so that all input types are mapped to vectors $\ff(\x^t)\in \R^M$. The function $\ff(\cdot)$ may be viewed as an encoding, e.g. words to vectors or images to a final dense layer via some trained ConvNet. The output at time $t\in[T]$ of the recurrent network with $R$ hidden channels, depicted at
the top of Figure~\ref{fig:intro_recurrent_net}, is given by:
\begin{align}\label{eq:shallow_rn}
\h^{t} &= g\left(\WHnone \h^{t-1},\WInone \ff(\x^t)\right) \\
\y^{t, \Theta} &= \WO \h^{t}, \nonumber
\end{align}
where $\h^t\in\R^R$ is the hidden state of the network at time $t$  ($\h^0$ is some initial hidden state), $\Theta$
denotes the learned parameters $\WInone \in \R^{R \times M},
\WHnone \in \R^{R \times R}, \WO \in \R^{C \times R}$, which are the \emph{input-to-hidden}, \emph{hidden-to-hidden} and \emph{hidden-to-output} weights matrices respectively, and $g$ is some non-linear operation.
Omitting a bias term for simplicity, the non-linearity of a common RNN is given by: 
\begin{align*}
  g^{\mathrm {RNN}}(\aaa,\bb) = \sigma(\aaa+\bb),
\end{align*}
where $\sigma(\cdot)$ is typically some point-wise non-linearity such as sigmoid, tanh etc. 

Since the hidden state $h^t$ is used both for transition and for emission, its inner representation plays a crucial role in prediction tasks. A discriminatively trained RNN effectively uses its state as a memory unit, trained to hold the most relevant information for both transition and emission. The temporal distance between relevant information in the data determines the length of memory required for solving a task, and the temporal length of consistencies a model is able to learn is referred to as the memory length of the model. According to this terminology, \emph{long-term memory capacity} refers to the ability of a recurrent network to leverage information from distant time-steps for its predictions, an ability that is crucial for solving almost every practical task of sequential data.

\subsection{Depth in Recurrent Networks}\label{sec:intro:depth}
The recent success of Convolutional Neural Networks (CNNs) is strongly attributed to the notion of depth. Generally speaking, depth refers to the principle of taking a non-linear parametric function, described by a computation graph, and composing its graph with itself over and over to create a more complex parametric function. A series of remarkable results in image recognition (\citep{Krizhevsky:2012wl}, \citep{he2016deep} and more), achieved by multi-layered CNNs, gave rise to the hypothesis that deep hierarchical models are exponentially more expressive than their shallow counterparts. This hypothesis was theoretically proven by ~\citep{NIPS2011_4350} and ~\citep{cohen2016deep}, for arithmetic variants of CNNs. 

Inspired by the central role depth takes in the architectural design of CNNs, we naturally ask the following question: \emph{What does depth bring forth to Recurrent Neural Networks?}\\
Before answering this question, we note that there are several different ways to introduce depth into the connectivity of a recurrent network. Moreover, as \citep{pascanu2013construct} highlighted, every recurrent network is deep when unfolded in time, as the computation path between an input at time-step $t$ to an output at time-step $k > t$ crosses several non-linear layers. But since the three basic computation units - \emph{input-to-hidden}, \emph{hidden-to-hidden} and  \emph{hidden-to-output} - are shallow \emph{w.r.t.} a single time event, we refer to conventional RNNs as shallow models.

Arguably the most natural way to deepen an RNN is by stacking together multiple hidden states, as done in the early works of \citep{schmidhuber1992learning} and \citep{elhihi1996}. We will focus on this form of depth, to be described in detail in the following paragraph. Alternative forms of deep RNNs can be constructed by replacing each of the basic computation units described above by a deep feed-forward network, as done in \citep{pascanu2013construct}.

Having chosen the stacking of hidden states as the deepening scheme, the extension of our setting to deep recurrent networks becomes natural. We follow the common approach (see e.g. \citep{hermans2013training}) where each layer acts as a recurrent network which receives the hidden state of the previous layer as its input. The output at time-step $t$ of the depth $L$ recurrent network with $R$
hidden channels in each layer, depicted at the bottom of
Figure~\ref{fig:intro_recurrent_net}, is constructed
by the following:
\begin{align}
\h^{t,l} &= g\left(\WH{l} \h^{t-1,l}, \WI{l} \h^{t,l-1}\right) \nonumber\\
\h^{t,0} &\equiv \ff(\x^t) \label{eq:deep_rn} \\
\y^{t,L, \Theta} &= \WO \h^{t,L}, \nonumber
\end{align}
where $\h^{t,l}\in\R^R$ is the state of the depth $l$ hidden unit at time-step $t$ ($\h^{0,l}$ is some initial hidden state per layer),
and $\Theta$ denotes the learned parameters. Specifically,
$\WI{l} \in \R^{R \times R} ~(l>1), \WH{l}\in\R^{R\times R}$
are the input and hidden weights matrices at depth $l$, respectively. For $l=1$,
the weights matrix which multiplies the inputs vector has the appropriate
dimensions: $\WI{1}\in \R^{R \times M}$. The output weights matrix is
$\WO\in \R^{C \times R}$ as in the shallow case, representing a final
calculation of the scores for all classes $1$ through $C$ at every time-step.

\section{Orthogonal RNNs}\label{sec:intro:orthRNNs}

RNNs are known to be difficult for optimization \citep{pascanu2013difficulty}. Training an RNN often fails completely due to gradient issues known as the \emph{Vanishing Gradient} and the \emph{Exploding Gradient}, to be discussed in Section \ref{subsec:gradient_issues}. A similar issue that could prevent a successful training is a situation where the state either extremely contracts or explodes.\footnote{For bounded activation functions, such as the hyperbolic tangent, explosion refers to the situation where the elements of the state saturates either to 1 or to -1.}

The theoretical analysis of this work, to be presented in Chapter \ref{chapter:theory}, was performed on Recurrent Arithmetic Circuits (see Section \ref{sec:theory:RAC}), an RNN variant that has the exact connectivity of conventional "vanilla" RNNs, and differs only in its non-linear operation. Thus, in order to properly demonstrate such theory empirically, a solution for the mentioned optimization issues has to be given while preserving the basic connectivity. For this purpose, we found the idea of training vanilla-RNNs with orthogonal transformations, to be discussed in the remainder of this section, the most suitable, as opposed to the alternative of using gating-based solutions such as LSTMs \citep{hochreiter1997long}.

A recent line of work (\citep{arjovsky2016unitary}, \citep{wisdom2016full}, \citep{jing2016tunable}, \citep{pmlr-v80-helfrich18a}) suggests to use orthogonal matrices for the \emph{hidden-to-hidden} transformation of the RNN. A squared matrix $A$ is orthogonal if $A^{T}A=AA^{T}=I$. The transformation performed by an orthogonal matrix is an isometry, \ie~norm preserving, in the sense that $\|Ax\|_{2}=\|x\|_{2}$. Accordingly, the integration of orthogonal matrices into the computation graph of an RNN is a straightforward way to prevent abrupt changes in the state's scale during training or prediction. Moreover, 
as we will see next, combining an orthogonal transformation with an identity based non-linearity such as ReLU (\citep{nair2010rectified}) or modReLU (\citep{arjovsky2016unitary}) formulates a remedy also for gradient issues. To conclude, the rational behind using orthogonal matrices for the \emph{hidden-to-hidden} transformation of an RNN is twofold: 
\begin{enumerate}[label=(\alph*)]
    \item To prevent the contraction of the state.
    \item As a remedy for gradient issues.
\end{enumerate}

\subsection{Orthogonality as a Remedy for Gradient Issues}\label{subsec:gradient_issues}

\emph{The following analysis is based on section 2 of ~\citep{arjovsky2016unitary}.} \\ 
Following the definitions given in equations \eqref{eq:shallow_rn} - \eqref{eq:g_rnn}, and denoting $W := W^{H}$, the transition of a common RNN from the state at time-step $t$ to the state at time-step $t+1$ can be described as
\begin{align*}
  &z^{t+1}=Wh^{t}+W^{I}f(x^{t+1})&\\
  &h^{t+1}=\sigma\left(z^{t+1}\right)&
\end{align*}

Say that an RNN is trained to minimize some non-negative loss function $l$ over a dataset of sequences of length $T$. Let $t \ll T$ be some time-step at the beginning of the sequence.  The \emph{Vanishing Gradient} issue refers to the phenomenon where during training the norm of $\frac{dl}{dh^{t}}$, the gradient of the loss \wrt~the state at time-step $t$, extremely contracts and effectively goes to zero. Similarly, \emph{Exploding Gradient} refers to a situation where $\|\frac{dl}{dh^{t}}\|\rightarrow\infty$.

Narrowing the discussion into the case where $\sigma$ is the ReLU activation, and $W$ is orthogonal, we will give an upper bound for the norm $\|\frac{dl}{dh^{t}}\|$ that guarantees that it can't explode.

Due to the diagonal structure of the Jacobian of element-wise functions such as ReLU, and using the chain rule, we see that the Jacobian of the state \wrt~to the state at the previous time-step takes the following form: 
\begin{align*}
  \frac{dh^{t+1}}{dh^{t}}=\frac{d\sigma\left(z^{t+1}\right)}{dz^{t+1}}\frac{dz^{t+1}}{dh^{t}}=diag\left(\sigma'\left(z^{t+1}\right)\right)W^{T}
\end{align*}
Consequently, the gradient of the loss \wrt~the state at time $t$ takes the form
\begin{align*}
\frac{dl}{dh^{t}}=\frac{dl}{dh^{T}}\prod_{k=t}^{T-1}\frac{dh^{k+1}}{dh^{k}}=\frac{dl}{dh^{T}}\prod_{k=t}^{T-1}(diag\left(\sigma'\left(z^{k+1}\right)\right)W^{T})=\frac{dl}{dh^{T}}\prod_{k=t}^{T-1}(D_{k+1}W^{T})
\end{align*}
where $D := diag\left(\sigma'\left(z^{k+1}\right)\right)$. 

We would like to calculate the norm of $\frac{dl}{dh^{t}}$. Considering the spectral norm for matrices and the $l_{2}$ norm for vectors, and using the two inequalities
\begin{align*}
\|Av\|\leq\|A\|\|v\|\\
\|AB\|\leq\|A\|\|B\|    
\end{align*}
we get the following upper bound:
\begin{align*}
\|\frac{dl}{dh^{t}}\|\leq\|\frac{dl}{dh^{T}}\|\prod_{k=t}^{T-1}\|D_{k+1}W^{T}\|\overset{\text{\ensuremath{W} is orthogonal}}{=}\|\frac{dl}{dh^{T}}\|\prod_{k=t}^{T-1}\|D_{k+1}\|.
\end{align*}
Since the spectral norm of a diagonal matrix with all diagonals positive is the maximal element, and since the derivative of ReLU is either 1 or 0, then
$\|D_{k}\|=\text{\ensuremath{\underset{}{max}(diag\left(\sigma'\left(z^{k}\right)\right))\in\left\{ 0,1\right\} }}$. Moreover, unless all values of $z^{k}$ are negative, then in fact $\|D_{k}\|=1$, and in this case 
\begin{align*}
    \|\frac{dl}{dh^{t}}\|\leq\|\frac{dl}{dh^{T}}\|
\end{align*} 
which proves that the situation where  $\|\frac{dl}{dh^{t}}\| \gg \|\frac{dl}{dh^{T}}\|$ is prevented, and consequently the gradient  $\frac{dl}{dh^{t}}$ can't explode. 

\subsection{Its all about Parameterization}
\label{subsec:orth_parameterizations}
After clarifying the rational behind using orthogonal matrices as a building block in recurrent networks, in this section we would like to dive into the details of how optimization of orthogonal RNNs is done in practice. During optimization, orthogonality can be imposed in several different ways with different levels of constraints on the \emph{hidden-to-hidden} matrix. 

In the simplest form, used by \citep{leSimpleWayToInitialize} and  \citep{henaff_OrthogonalRNNs}, $W^{H}$ is initialized as an orthogonal matrix, by some sampling scheme of random orthogonal matrices, and then trained without any constraints on its orthogonality. Though proved to be a useful initialization scheme, this approach doesn't guarantee orthogonality of $W^{H}$ beyond the first few training iterations. Taking the orthogonality one step further, we note a technique used by \citep{brock2016_neuralPhotoEditing} in the context of generative models, in which the regularization term $\|WW^{T} - I\|$ is added to the objective function. This rather soft constraint encourages orthogonality of the learned matrix $W$ and doesn't narrow the hypothesis space directly. 

The techniques mentioned above indeed encourage $W^{H}$ at being orthogonal, but this is done in a somewhat soft manner and the door is still open for more thorough approaches. This leads us to the highest level of orthogonality, where the \emph{hidden-to-hidden} matrix is explicitly restricted at being orthogonal. Implementing this preconditioning on $W^{H}$ is not trivial, as the naive approach of projecting $W^{H}$ onto the manifold of orthogonal matrices in each training iteration is computationally expensive, and an intricate parameterization scheme is needed for efficiency. \citep{arjovsky2016unitary}, \citep{wisdom2016full} and \citep{jing2016tunable} tackled the broader case of restricting $W^{H}$ at being unitary, each in a different parameterization approach, while \citep{pmlr-v80-helfrich18a} dealt directly with orthogonal matrices, with a novel paramterization scheme of its own. Among this works, we find the works of \citep{wisdom2016full} and \citep{pmlr-v80-helfrich18a} the most relevant, as they allow straightforward optimization over the full orthogonal manifold. 

Full Capacity Unitary RNNs (Full Capacity uRNNs), presented in \citep{wisdom2016full}, are recurrent networks optimized with a parametrization scheme that keeps $W^{H}$ on the Steifel Manifold through a multiplicative update rule, based on the work of \citep{tagare2011}. The Steifel manifold is the set of matrices $W$ satisfies $WW^T=I$. Restricting ourselves to the set of squared $R\times R$ matrices, the Steifel manifold is the set $\{W\in \R^{R\times R}:\ WW^T=I\}$. Let $W := W^{H}$ be the \emph{hidden-to-hidden} matrix and let $l$ be the loss function being minimized. Equipping the Steifel Manifold with the canonical inner product $\left\langle Z_{1},Z_{2}\right\rangle =tr\left(Z_{1}^{T}\left(I-\frac{1}{2}WW^{T}\right)Z_{2}\right)$ (see \citep{tagare2011}), the gradient along the Steifel manifold, of $l$ \wrt~$W$ arranged as a matrix, can be expressed as $AW$, where
\begin{align*}
    &A=G^TW-W^TG&\\
    &G=\frac{dl}{dW}&
\end{align*}
Accordingly, \citep{wisdom2016full} suggests the following update rule for $W$:
\begin{align*}
    &G=\frac{dl}{dW}&\\
    &A=G^TW-W^TG&\\
    &\Tilde{W}=(I+\frac{\eta}{2}A)^{-1}(I-\frac{\eta}{2}A)W&
\end{align*}
where $\frac{dl}{dW}$ is the gradient of $l$ \wrt~$W$ arranged as a matrix, $\eta$ is the learning rate, and $\Tilde{W}$ is the updated matrix.
This approach has proven to be useful empirically, achieving impressive results in synthetic RNN benchmarks such as the Copying Memory Problem (to be discussed in Chapter \ref{chap:experiments}), but suffers from numerical instabilities, causing $W^{H}$ to deviate from the Steifel Manifold as training proceeds.

\citep{pmlr-v80-helfrich18a} claimed that the numerical instability of Full Capacity uRNNs is a result of the update rule being multiplicative. As a remedy, they presented Scaled Cayley Orthogonal RNNs (scoRNN), which are recurrent networks with orthogonal \emph{hidden-to-hidden} matrix, optimized by an additive update rule via the Cayley Transform to the set of skew-symmetric matrices. The Cayley Transform, defined by 
\begin{align*}
    A\mapsto\left(I+A\right)^{-1}\left(I-A\right),
\end{align*}
is a bijection between the set of orthogonal matrices with no -1 eigenvalues and the set of skew-symmetric matrices A, \ie~matrices that satisfy $A^T=-A$. In other words, every orthogonal matrix W that does not have -1 as its eigenvalue, can be expressed as 
\begin{align*}
    W=\left(I+A\right)^{-1}\left(I-A\right)
\end{align*}
for the skew-symmetric matrix
\begin{align*}
    A=\left(I+W\right)^{-1}\left(I-W\right)
\end{align*}
We note that:
\begin{enumerate}[label=(\alph*)]
    \item The set of skew-symmetric matrices is closed under addition.
    \item Every orthogonal matrix can be represented as $W=\left(I+A\right)^{-1}\left(I-A\right)D$ where $D$ is a diagonal matrix with diagonal values $\in\left\{ 1,-1\right\}$.
\end{enumerate}
\citep{pmlr-v80-helfrich18a} suggests the idea of optimizing $A$, the skew-symmetric matrix obtained by applying the Cayley Transform on $W^{H}$, rather than optimizing $W^{H}$ directly. In specific, they show that the gradient of the loss function \wrt~$A$ arranged as a matrix is also skew-symmetric and thus, by property (a), a conventional additive gradient descent step would keep $A$ skew-symmetric. After $A$ is being updated, $W^{H}$ is reconstructed by applying back the Cayley Transform on $A$. Using these facts, \citep{pmlr-v80-helfrich18a} suggests the following update rule for $W := W^{H}$:
\begin{align*}
    &V=\left(I+A\right)^{-T}\frac{dl}{dW}\left(D+W^{T}\right)&\\
    &\frac{dl}{dA}=V^{T}-V&\\
    &\Tilde{A}=A-\eta\frac{dl}{dA}&\\
    &\Tilde{W}=\left(I+\Tilde{A}\right)^{-1}\left(I-\Tilde{A}\right)D&
\end{align*}
where $\frac{dl}{dW}$ is the gradient of $l$ \wrt~$W$ arranged as a matrix, $D$ is a diagonal matrix with diagonals $\in\{0,1\}$ chosen as an hyper-parameter and $\eta$ is the learning rate.

\begin{figure}
	\centering
	\includegraphics[scale=0.5]{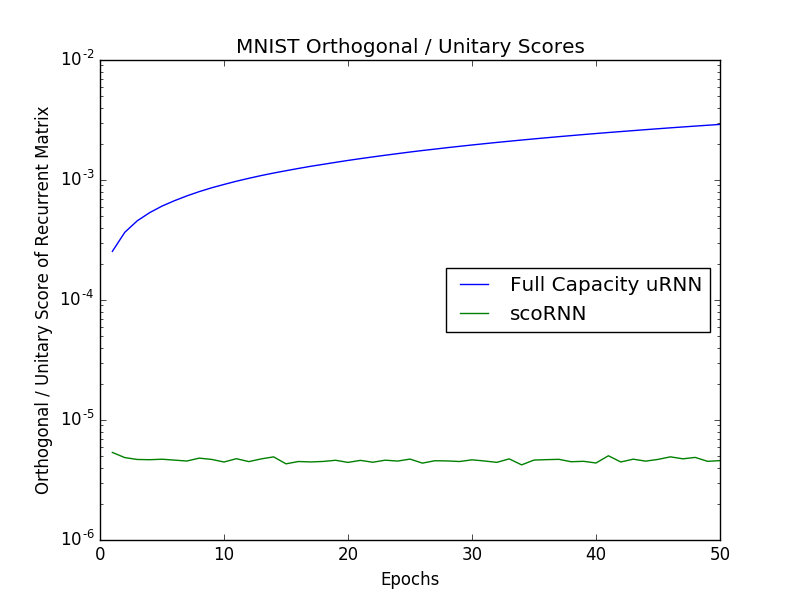}
	\ifdefined\SQUEEZE \vspace{-6mm} \fi
	\caption{Taken from \citep{pmlr-v80-helfrich18a} section 5: Deviation from orthogonality $\|WW^T-I\|$ of Full Capacity uRNN and scoRNN during training on the Permuted Pixel-by-Pixel MNIST (see Section \ref{sec:exp:perm_mnist} for the task definition).}
	\ifdefined\SQUEEZE \vspace{-6mm} \fi
	\label{fig:orthogonality_during_training}
\end{figure}

Aiming at finding the simplest approach for a successful optimization of conventional RNNs in our experiments, to be presented in Chapter \ref{chap:experiments}, we've tried all approaches presented in this section, including the soft approaches, the Full Capacity uRNN and the scoRNN. The full-orthogonal approaches performed significantly better than the soft alternatives. When using Full Capacity uRNNs, with $W^{H}$ restricted at being orthogonal rather than unitary, we encountered the same numerical issues reported at \citep{pmlr-v80-helfrich18a} (see figure \ref{fig:orthogonality_during_training}), and while optimizing with scoRNNs these issues didn't occur. Moreover, scoRNNs performed better than Full Capacity RNNs in each of the problems chosen for this work (see Chapter \ref{chap:experiments} and Appendix \ref{app:natural_data_experiments}), and so we ultimately used scoRNN as the optimization scheme for the experimental evaluation. 

Though all approaches presented in this section has the same goal of keeping $W^{H}$ orthogonal, they differ in their parameterization scheme, and significantly differ in performance, suggesting that parameterization of the \emph{hidden-to-hidden} matrix plays a crucial role in any practical setting of training a recurrent network.

\section{Basic Concepts in Tensor Theory}\label{sec:intro:tensors}
In this section we lay out basic concepts in tensor theory required for
the theoretical analysis given in Chapter~\ref{chapter:theory}. The core concept of a \emph{tensor} may be thought of as a
multi-dimensional array. The \emph{order} of a
tensor is defined to be the number of indexing entries in the array,
referred to as \emph{modes}. The \emph{dimension} of a tensor in a particular
mode is defined as the number of values taken by the index in that
mode. If $\A$ is a tensor of order $T$ and dimension $M_i$ in each mode
$i\in[T]$, its entries are denoted $\A_{d_1...d_T}$, where the index in each
mode takes values $d_i\in [M_i]$.

\subsection{Tensor Product}\label{intro:tensors:tensor_product}
A fundamental operator in tensor analysis is the \emph{tensor product}, which
we denote by $\otimes$. It is an operator that intakes two tensors
$\A\in\R^{M_1{\times\cdots\times}M_P}$ and
$\B\in\R^{M_{P+1}{\times\cdots\times}M_{P+Q}}$ (orders $P$ and $Q$ respectively), and returns a tensor
$\A\otimes\B\in\R^{M_1{\times\cdots\times}M_{P+Q}}$  (order $P+Q$) defined by:
$(\A\otimes\B)_{d_1{\ldots}d_{P+Q}}=\A_{d_1{\ldots}d_P}\cdot\B_{d_{P+1}{\ldots}d_{P+Q}}$.

\subsection{Grid Tensors}\label{intro:tensors:grid_tensors}
Grid tensors~\citep{hackbusch2012tensor}, which will be used as a central tool in
Section~\ref{sec:corr:grid}, are a form of function discretization. Essentially, the function is
evaluated for a set of points on an exponentially large grid in the
input space and the outcomes are stored in a tensor. Formally, fixing a set of \emph{template} vectors
$\x^{(1)},\ldots,\x^{(M)} \in \X$, the points on the grid are the set
$\{(\x^{(d_1)},\ldots,\x^{(d_T)})\}_{d_1,\ldots,d_T=1}^M$. Given a function
$y(\x^1,\ldots,\x^T)$, the set of its values on the grid arranged in the form of a tensor are
called the grid tensor induced by $y$, denoted
$\A(y)_{d_1,\ldots,d_T} \equiv y(\x^{(d_1)},\ldots,\x^{(d_T)})$.

\subsection{Matricization}\label{intro:tensors:matricization}

An additional concept we will make use of is the \emph{matricization of $\A$ \wrt~the partition $(S,E)$}, denoted $\mat{\A}_{S,E}\in\R^{M^{\nicefrac{T}{2}}\times M^{\nicefrac{T}{2}}}$, which is essentially the arrangement of the tensor elements as a matrix whose rows correspond to $S$ and columns to $E$. The formal definition is given below:

\begin{dfn}\label{def:matricization}
    Suppose $\A\in\R^{M{\times\cdots\times}M}$
    is a tensor of order $T$, and let $(I,J)$ be a partition of $[T]$, \ie~$I$
    and~$J$ are disjoint subsets of $[T]$ whose union gives~$[T]$.
    The \emph{matricization of $\A$ \wrt~the partition $(I,J)$}, denoted
    $\mat{\A}_{I,J}$, is the $M^{\abs{I}}$-by-$M^{\abs{J}}$ matrix holding the entries of $\A$ such that $\A_{d_1{\ldots}d_T}$ is placed in row index $1+\sum_{t=1}^{\abs{I}}(d_{i_t}-1)M^{\abs{I}-t}$ and column index $1+\sum_{t=1}^{\abs{J}}(d_{j_t}-1)M^{\abs{J}-t}$.
\end{dfn} 

\subsection{The Tensor-Train Decomposition}\label{intro:tensors:tt_decomposition}

A tensor $\A$ of order $T$ and dimension $M$ in each mode has $M^T$ elements in total. Due to the exponential dependent on the order $T$ it is often impractical to store the entire tensor explicitly in memory. Fortunately, in many practical situations the tensor of interest is not fully-ranked, \ie~there exist degenerate degrees of freedom, and thus it can be represented compactly, with an amount of memory substantially smaller than $M^T$. These compact representations, known as \emph{tensor decompositions}, were used in recent deep learning works both for theoretical analyses (see e.g. \citep{cohen2016expressive}) and for practical purposes (see e.g. \cite{novikov2015tensorizingNN}, \citep{khrulkov2018expressive}). Among the different approaches, we will focus on the the \emph{tensor train} (TT) decomposition (\citep{oseledets2011tensor}). 

An expression of the form $\A_{d_1...d_T} = G_{1}(d_1)G_{2}(d_2)...G_{T}(d_T)$, where $G_{t}(d_t)$ ($t\in [T],\ d_{t}\in [M]$) are matrices, is a \emph{TT-decomposition} of an order-$T$ tensor $\A$ with dimension $M$ in each mode. The order-3 tensors $G_{t} \in \R^{R_{t-1}, M, R_{t}}$ are referred to as the \emph{TT-cores} and the dimensions $R_{t}$ are called the \emph{TT-ranks}. We would note the substantial reduction of parameters \wrt~the full description of the $M^T$ tensor elements. Here, $\A$ is represented with only $O(T\cdot M\cdot R^{2})$ parameters where $R=max_{t=0,...,T}(R_{t})$. 

In Section \ref{sec:corr:grid}~we will show how the computation performed by a shallow Recurrent Arithmetic Circuit (see Section \ref{sec:theory:RAC}) can be described in the language of TT-decompositions. Due to the recurrent structure of the computation, the TT-cores in this case are shared across the different modes, \ie~the same core $G \in \R^{R, M, R}$ is used for each mode, and thus the decomposition takes the form $\A_{d_1...d_T} = G(d_1)G(d_2)...G(d_T)$. 

\clearpage
%
%
\let\textcircled=\pgftextcircled
\chapter{Theoretical Analysis}
\label{chapter:theory}
\textit{The content of this chapter is an edited version of \citep{RAC_paper}, a joint work with Yoav Levine, Or Sharir and Amnon Shashua.}
\initial{I}n order to take-on the questions presented in Chapter \ref{chapter:intro}, we introduce in Section~\ref{sec:theory:RAC} a
recurrent network referred to as a recurrent arithmetic circuit (RAC)
that shares the architectural features of RNNs, and differs from them in
the type of non-linearity used in the calculation. This type of connection
between state-of-the-art machine learning algorithms and arithmetic circuits
(also known as Sum-Product Networks~\citep{poon2011sum}) has well-established
precedence in the context of neural networks. \citep{NIPS2011_4350} prove a
depth efficiency result on such networks, and \citep{cohen2016expressive}
theoretically analyze the class of Convolutional Arithmetic Circuits which
differ from common ConvNets in the exact same fashion in which RACs differ from
more standard RNNs. Conclusions drawn from such analyses were empirically shown
to extend to common ConvNets (\citep{sharir2018expressive,cohen2017inductive,cohen2017boosting,levine2018deep}).
Beyond their connection to theoretical models, RACs are similar to
empirically successful recurrent network architectures. The modification which
defines RACs resembles that of Multiplicative RNNs used by
\citep{sutskever2011generating} and of Multiplicative Integration networks used
by \citep{wu2016multiplicative}, which provide a substantial performance boost
over many of the existing RNN models.
In order to obtain our results, we make a connection between RACs and the Tensor Train (TT) decomposition \citep{oseledets2011tensor}, which suggests that Multiplicative RNNs may be related to a generalized TT-decomposition, similar to the way \citep{cohen2016convolutional} connected ReLU ConvNets to generalized tensor decompositions.

We move on to introduce in Section~\ref{sec:theory:separation_rank} the notion of \emph{Start-End separation rank} as a measure of the recurrent network's ability to model elaborate long-term dependencies. In order to analyze the long-term dependencies modeled by a
function defined over a sequential input which extends $T$ time-steps, we partition the
inputs to those which arrive at the first $\nicefrac{T}{2}$ time-steps
(``Start'') and the last $\nicefrac{T}{2}$ time-steps (``End''), and ask how far
the function realized by the recurrent network is from being separable \wrt~this
partition. Distance from separability is measured through the notion of
separation rank~\citep{beylkin2002numerical}, which can be viewed as a
surrogate of the $L^2$ distance from the closest separable function. For a given
function, high Start-End separation rank implies that the function induces
strong dependency between the beginning and end of the input sequence, and vice
versa.

Finally, in Section~\ref{sec:theory:bounds} we directly address the depth enhanced long-term memory
question above, by examining depth $L=2$ RACs and proving that functions realized by these deep networks enjoy Start-End
separation ranks that are combinatorially higher than those of shallow networks, implying that indeed these functions can model more elaborate input dependencies
over longer periods of time.
An additional reinforcing result is that the
Start-End separation rank of the deep recurrent network grows combinatorially with
the sequence length, while that of the shallow recurrent network is
\emph{independent} of the sequence length. Informally, this implies that vanilla
shallow recurrent networks are inadequate in modeling dependencies of long input
sequences, since in contrast to the case of deep recurrent networks, the modeled
dependencies achievable by shallow ones do not adapt to the actual length of
the input. In an effort to extend our separation results to RNNs with depth $L > 2$, we present and motivate a quantitative conjecture by which the Start-End
separation rank of recurrent networks grows combinatorially with the network depth. A proof of this conjecture, which provides an even deeper insight regarding the advantages of depth in recurrent networks, is left as an open problem.

\section{Recurrent Arithmetic Circuits}\label{sec:theory:RAC}
In this section, we introduce a class of recurrent networks referred to as
Recurrent Arithmetic Circuits (RACs), which shares the architectural features of
standard RNNs. As demonstrated below, the operation of RACs on sequential data
is identical to the operation of RNNs, where a hidden state mixes information
from previous time-steps with new incoming data (see
Figure~\ref{fig:intro_recurrent_net}). The two classes differ only in the type of
non-linearity used in the calculation, as described by
Equations~\eqref{eq:g_rnn}-\eqref{eq:g_rac}. In the following sections, we utilize
the algebraic properties of RACs for proving results regarding their ability to
model long-term dependencies of their inputs.

	For common RNNs, as presented in Chapter \ref{chapter:intro} the non-linearity is given by:
	\begin{equation}\label{eq:g_rnn}
	g^{\mathrm {RNN}}(\aaa,\bb) = \sigma(\aaa+\bb),
	\end{equation}
	where $\sigma(\cdot)$ is typically the hyperbolic tangent. For the newly introduced class of RACs, $g$ is given by:
	\begin{equation} \label{eq:g_rac}
	g^{\mathrm {RAC}}(\aaa,\bb) = \aaa \odot \bb,
	\end{equation}
	where the operation $\odot$ stands for element-wise multiplication between	vectors, for which the resultant vector upholds
	$(\aaa \odot \bb)_i = a_i \cdot b_i$. This form of merging the input and the hidden state by multiplication rather than addition is referred to as Multiplicative Integration~\citep{wu2016multiplicative}.
	
	In the case of deep recurrent networks, as defined in \ref{eq:deep_rn}, the non-linear operation $g$ determines the type of the deep recurrent network,
	where a common deep RNN is obtained by choosing $g=g^{\mathrm {RNN}}$
	[Equation~\eqref{eq:g_rnn}], and a deep RAC is obtained for
	$g=g^{\mathrm {RAC}}$~[Equation~\eqref{eq:g_rac}].
	
	We consider the newly presented class of RACs to be a good surrogate of common
	RNNs. Firstly, there is an obvious structural resemblance between the two
	classes, as the recurrent aspect of the calculation has the exact same form in
	both networks (Figure~\ref{fig:intro_recurrent_net}). In fact, recurrent networks that
	include Multiplicative Integration similarly to RACs (and include additional non-linearities), have been shown to
	outperform many of the existing RNN models~\citep{sutskever2011generating,
		wu2016multiplicative}. Secondly, as mentioned above, arithmetic circuits have
	been successfully used as surrogates of convolutional networks. The fact that
	\citep{cohen2016convolutional} laid the foundation for extending the proof
	methodologies of convolutional arithmetic circuits to common ConvNets with ReLU
	activations, suggests that such adaptations may be made in the recurrent
	network analog, rendering the newly proposed class of recurrent networks all
	the more interesting. Finally, RACs have recently been shown to operate well in practical settings \citep{khrulkov2018expressive}.
	In the following sections, we make use of the algebraic properties of RACs in
	order to obtain clear-cut observations regarding the benefits of depth in recurrent
	networks.

\section{A Functional Measure of Long-Term Memory}\label{sec:theory:separation_rank}
In this section, we establish means for quantifying the ability of recurrent
	networks to model long-term temporal dependencies in the sequential input data.
	We begin by introducing the Start-End
	separation-rank of the function realized by a
	recurrent network as a measure of the amount of information flow across time
	that can be supported by the network. We then tie the Start-End separation rank to the
	algebraic concept of grid tensors (see Section~\ref{intro:tensors:grid_tensors}), which will allow us
	to employ tools and results from tensorial analysis in order to show that depth
	provides a powerful boost to the ability of recurrent networks to model
	elaborate long-term temporal dependencies.
	\ifdefined\SQUEEZE \vspace{-2mm} \fi
	
	\subsection{The Start-End Separation Rank}\label{sec:corr:sep}
	\ifdefined\SQUEEZE \vspace{-2mm} \fi
	
	We define below the concept of the \emph{Start-End separation rank} for
	functions realized by recurrent networks after $T$ time-steps, \ie~functions that take as input $X=(\x^1,\ldots,\x^T)\in\X^T$. The separation rank
	quantifies a function's distance from separability with respect to two
	disjoint subsets of its inputs. Specifically, let $(S,E)$ be a partition of
	input indices, such that $S=\{1,\ldots,\nicefrac{T}{2}\}$ and
	$E=\{\nicefrac{T}{2} + 1, \ldots, T\}$ (we consider even values of $T$
	throughout this document for convenience of presentation). This implies that
	$\{\x^s\}_{s\in S}$ are the first $T/2$ (``Start'') inputs to the network, and
	$\{\x^e\}_{e\in E}$ are the last $T/2$ (``End'') inputs to the network. For a
	function $y:\X^T\to\R$, the \emph{Start-End separation rank} is defined as
	follows:
	\begin{align}
	\sep{S,E}{y} \equiv\min\left\{K\in\N\cup\{0\}:\exists{g^s_1{\ldots}g^s_K:\X^{T/2}\to\R,~g^e_1{\ldots}g^e_K:\X^{T/2}\to\R}~~s.t.\right.
	\quad~\label{eq:sep_rank}\\
	\left.y(\x^1,\ldots,\x^T) =\sum\nolimits_{\nu=1}^{K}g^s_{\nu}(\x^{1},\ldots,\x^{T/2})g^e_\nu(\x^{T/2+1},\ldots,\x^{T})
	\right\}.
	\nonumber
	\end{align}
	In words, it is the minimal number of summands that together give $y$, where
	each summand is \emph{separable \wrt~$(S,E)$}, \ie~is equal to a product of two
	functions~--~one that intakes only inputs from the first $T/2$ time-steps, and
	another that intakes only inputs from the last $T/2$ time-steps.
	
	The separation rank \wrt~a general partition of the inputs was
	introduced in \citep{beylkin2002numerical} for high-dimensional numerical analysis, and was employed for various
	applications, \eg~chemistry~\citep{harrison2003multiresolution},
	particle engineering~\citep{hackbusch2006efficient}, and machine
	learning~\citep{beylkin2009multivariate}. \citep{cohen2017inductive} connect
	the separation rank to the $L^2$~distance of the function from the set of separable
	functions, and use it to measure dependencies modeled by deep
	convolutional networks. \citep{levine2018deep} tie the separation rank to the
	family of quantum entanglement measures, which quantify dependencies in
	many-body quantum systems.
	
	In our context, if the Start-End separation rank of a function realized by a
	recurrent network is equal to~$1$, then the function is separable, meaning it
	cannot model any interaction between the inputs which arrive at the beginning
	of the sequence and the inputs that follow later, towards the end of the
	sequence. Specifically, if $\sep{S,E}{y}=1$ then there exist
	$g^s:\X^{T/2}\to\R$ and $g^e:\X^{T/2}\to\R$ such that
	$y(\x^1,\ldots,\x^T)=g^s(\x^1,\ldots,\x^{T/2})g^e(\x^{T/2+1},\ldots,\x^{T})$,
	and the function~$y$ cannot take into account consistency between the values of
	$\{\x^1,\ldots,\x^{T/2}\}$ and those of $\{\x^{T/2+1},\ldots,\x^{T}\}$. In a
	statistical setting, if~$y$ were a probability density function, this would
	imply that $\{\x^1,\ldots,\x^{T/2}\}$ and $\{\x^{T/2+1},\ldots,\x^{T}\}$ are
	statistically independent. The higher $\sep{S,E}{y}$ is, the farther~$y$ is
	from this situation, \ie~the more it models dependency between the beginning
	and the end of the inputs sequence. Stated differently, if the recurrent
	network's architecture restricts the hypothesis space to functions with low
	Start-End separation ranks, a more elaborate long-term temporal dependence, which
	corresponds to a function with a higher Start-End separation rank, cannot be
	learned.
	
	In Section~\ref{sec:theory:bounds} we show that deep RACs support Start-End
	separations ranks which are combinatorially larger than those supported by
	shallow RACs, and are therefore much better fit to model long-term temporal
	dependencies. To this end, we employ in the following sub-section the
	algebraic tool of \emph{grid tensors} that will allow us to evaluate the Start-End separation ranks of deep and
	shallow RACs.

	\subsection{Bounding the Start-End Separation Rank via Grid Tensors}\label{sec:corr:grid}
	\ifdefined\SQUEEZE \vspace{-1mm} \fi
	
	We consider the
	function realized by a shallow RAC with $R$ hidden channels, which
	computes the score of class $c\in[C]$ at time $T$. This function, which is
	given by a recursive definition in Equations~\eqref{eq:shallow_rn} and~\eqref{eq:g_rac},
	can be alternatively written in the following closed form:
	\begin{equation}
	y^{T,1, \Theta}_{c}\left(\x^1,\ldots,\x^T\right)
	= \sum\nolimits_{d_1{\ldots}d_T=1}^M\left(\A^{T,1, \Theta}_c\right)_{d_1,\ldots,d_T}\prod\nolimits_{i=1}^{T} f_{d_i}(\x^i),
	\label{eq:score}
	\end{equation}
	where the order $T$ tensor $\A^{T,1, \Theta}_c$, which lies at the heart of
	the above expression, is referred to as the \emph{shallow RAC weights tensor},
	since its entries are polynomials in the network weights $\Theta$.
	Specifically, denoting the rows of the input weights matrix, $\WInone$,
	by $\aaa^{\mathrm{I},\alpha}\in\R^M$ (or element-wise:
	$a^{\mathrm{I},\alpha}_j=\WInone_{\alpha,j}$), the rows of the hidden
	weights matrix, $\WHnone$, by $\aaa^{\mathrm{H},\beta}\in\R^R$ (or
	element-wise: $a^{\mathrm{H},\beta}_j=\WHnone_{\beta,j}$), and the rows
	of the output weights matrix, $\WO$, by
	$\aaa^{\mathrm{O},c}\in\R^R,~c\in[C]$ (or element-wise:
	$a^{\mathrm{O},c}_j=\WO_{c,j}$), the shallow RAC weights tensor can
	be gradually constructed in the following fashion:
	\begin{align}
	\underbrace{\phi^{1,\beta}}_{\text{order $1$ tensor}} &= \sum\nolimits_{\alpha=1}^{R}~ a_\alpha^{\mathrm{H},\beta}~~
	\aaa^{\mathrm{I},\alpha}
	\nonumber \\
	\underbrace{\phi^{2,\beta}}_{\text{order $2$ tensor}} &= \sum\nolimits_{\alpha=1}^{R}~ a_\alpha^{\mathrm{H},\beta}~~
	\phi^{\mathrm{1},\alpha} ~\otimes \aaa^{\mathrm{I},\alpha}
	\nonumber \\[-3mm]
	&~~~\cdots
	\nonumber\\[-1mm]
	\underbrace{\phi^{t,\beta}}_{\text{order $t$ tensor}} &= \sum\nolimits_{\alpha=1}^{R} a_\alpha^{\mathrm{H},\beta}
	\phi^{t-1,\alpha} \otimes
	\aaa^{\mathrm{I},\alpha}
	\nonumber\\[-3mm]
	&~~~\cdots
	\nonumber\\[-1mm]
	\underbrace{\A^{T,1, \Theta}_c}_{\text{order $T$ tensor}} &= \sum\nolimits_{\alpha=1}^{R}
	a_\alpha^{\mathrm{O},c}
	\phi^{T-1,\alpha} \otimes
	\aaa^{\mathrm{I},\alpha},
	\label{eq:tt_decomp}
	\end{align}
	having set $\h^0=\left(W^{\textrm H}\right)^{\dagger}\1$, where $\dagger$ is the pseudoinverse operation. In the above equation, the tensor products, which appear inside the sums, are
	directly related to the Multiplicative Integration property of RACs
	[Equation~\eqref{eq:g_rac}]. The sums originate in the multiplication of the hidden
	states vector by the hidden weights matrix at every time-step
	[Equation~\eqref{eq:shallow_rn}].
	The construction of the shallow RAC weights tensor, presented in Equation~\eqref{eq:tt_decomp}, is referred to as a Tensor Train (TT) decomposition (see Section~\ref{intro:tensors:tt_decomposition}) of
	TT-rank $R$ in the tensor analysis community~\citep{oseledets2011tensor} and is analogously described by a
	Matrix Product State (MPS) Tensor Network (see \citep{orus2014practical}) in the quantum physics community.
	See Appendix~\ref{app:rac_tns} for the Tensor Networks construction of deep and shallow RACs, which provides graphical insight regarding the complexity brought forth by depth in recurrent networks.

	The grid tensors (see Section~\ref{intro:tensors:grid_tensors}) of functions realized by recurrent networks, will allow us to
	calculate their separations ranks and establish definitive conclusions
	regarding the benefits of depth for these networks.
	Having presented the tensorial structure of the function realized by a shallow
	RAC, as given by Equations~\eqref{eq:score} and~\eqref{eq:tt_decomp} above, we are now
	in a position to tie its Start-End separation rank to its grid tensor, as
	formulated in the following claim:
	\begin{claim}\label{claim:grid_sep_shallow}
		Let $y^{T,1, \Theta}_c$ be a function realized by a shallow RAC
		(top of Figure~\ref{fig:intro_recurrent_net}) after $T$
		time-steps, and let $\A^{T,1, \Theta}_c$ be its shallow RAC weights tensor,
		constructed according to Equation~\eqref{eq:tt_decomp}.
		Assume that the network's initial mapping functions~$\{f_{d}\}_{d=1}^M$ are
		linearly independent, and that they, as well as the
		functions~$g^s_\nu,g^e_\nu$ in the definition of Start-End separation rank
		[Equation~\eqref{eq:sep_rank}], are measurable and square-integrable.
		Then, there exist template vectors $\x^{(1)}, \ldots, \x^{(M)} \in \X$
		such that the following holds:
		\begin{equation}
		\sep{S,E}{y^{T,1, \Theta}_c}=\rank{\mat{\A^{T,1, \Theta}_c}_{S,E}}=\rank{\mat{\A(y^{T,1, \Theta}_c)}_{S,E}},
		\end{equation}
		where $\A(y^{T,1, \Theta}_c)$ is the grid tensor of $y^{T,1, \Theta}_c$
		\wrt~the above template vectors, and $\mat{\A}_{S,E}$ denotes the matricization (See section \ref{intro:tensors:matricization}) of a tensor $\A$ \wrt~the Start-End partition.
	\end{claim}
	\ifdefined\SQUEEZE \vspace{-4mm}\fi
	
	\begin{proof}
		We first note that though square-integrability may seem as a limitation at first glance (for
		example neurons $f_{d}(\x)=\sigma(\w_d^\top\x+b_d)$ with sigmoid or ReLU
		activation $\sigma(\cdot)$, do not meet this condition), in practice our inputs are bounded (\eg~image pixels by holding
		intensity values, etc). Therefore, we may view these functions as having compact support,
		which, as long as they are continuous (holds in all cases of interest),
		ensures square-integrability.
		
		We begin by proving the equality
		$\sep{S,E}{y_c^{T,1, \Theta}} = \rank{\mat{\A_c^{T,1, \Theta}}_{S,E}}$. As shown in
		\citep{cohen2017inductive}, for any function
		$f:{\cal{X}}\times \cdots \times {\cal{X}} \to R$ which follows the structure of
		Equation~\eqref{eq:score} with a general weights tensor $\A$, assuming that
		$\{f_d\}_{d=1}^M$ are linearly independent, measurable, and
		square-integrable (as assumed in Claim~\ref{claim:grid_sep_shallow}), it
		holds that $\sep{S,E}{f} = \rank{\mat{\A}_{S,E}}$. Specifically, for
		$f=y_c^{T,1, \Theta}$ and $\A = \A_c^{T,1, \Theta}$ the above equality holds.
		
		It remains to prove that there exists template vectors for which
		\begin{align*}
		    \rank{\mat{\A_c^{T,1, \Theta}}_{S,E}} = \rank{\mat{\A(y_c^{T,1, \Theta})}_{S,E}}.
		\end{align*}
		For any given set of template vectors $\x^{(1)},\ldots,\x^{(M)}\in\X$, we define
		the matrix $F \in \R^{M \times M}$ such that $F_{ij} = f_j(\x^{(i)})$, for which
		it holds that:
		\begin{align*}
		\A(y_c^{T,1, \Theta})_{k_1,\ldots,k_T} &= \sum_{d_1,\ldots,d_T = 1}^M \left(\A_c^{T,1, \Theta} \right)_{d_1,\ldots,d_T} \prod_{i=1}^T f_{d_i}(\x^{(k_i)}) \\
		&= \sum_{d_1,\ldots,d_T = 1}^M \left(\A_c^{T,1, \Theta} \right)_{d_1,\ldots,d_T} \prod_{i=1}^T F_{k_i d_i}.
		\end{align*}
		The right-hand side in the above equation can be regarded as a linear
		transformation of $\A_c^{T,1, \Theta}$ specified by the tensor operator
		$F \otimes \cdots \otimes F$, which is more commonly denoted by
		$(F \otimes \cdots \otimes F)(\A_c^{T,1, \Theta})$. According to lemma 5.6 in
		\citep{hackbusch2012tensor}, if $F$ is non-singular then
		$\rank{\mat{(F \otimes \cdots \otimes F)(\A_c^{T,1, \Theta})}_{S,E}} =
		\rank{\mat{\A_c^{T,1, \Theta}}_{S,E}}$. To conclude
		the proof, we simply note that \citep{cohen2016convolutional}
		showed that if $\{f_d\}_{d=1}^M$ are linearly independent then there exists
		template vectors for which $F$ is non-singular.
	\end{proof}
	
	The above claim establishes an equality between the Start-End separation rank
	and the rank of the matrix obtained by the corresponding grid tensor
	matricization (see Section~\ref{intro:tensors:matricization}), denoted $\mat{\A(y^{T,1, \Theta}_c)}_{S,E}$, with respect to a
	specific set of template vectors. Note that the limitation to specific
	template vectors does not restrict our results, as grid tensors are
	merely a tool used to bound the separation rank. The additional equality to the rank of the
	matrix obtained by matricizing the shallow RAC weights tensor, will be of use to
	us when proving our main results below (Theorem~\ref{theorem:main_result}).
	
	Due to the inherent use of data duplication in the computation preformed by a
	deep RAC (see Appendix~\ref{app:rac_tns:deep} for further details), it cannot be
	written in a closed tensorial form similar to that of Equation~\eqref{eq:score}.
	This in turn implies that the equality shown in
	Claim~\ref{claim:grid_sep_shallow} does not hold for functions realized by deep
	RACs. The following claim introduces a fundamental relation between a function's
	Start-End separation rank and the rank of the matrix obtained by the
	corresponding grid tensor matricization. This relation, which holds for all functions, is
	formulated below for functions realized by deep RACs:
	\begin{claim}\label{claim:grid_sep_deep}
		Let $y^{T,L, \Theta}_c$ be a function realized by a depth $L$ RAC
		(bottom of Figure~\ref{fig:intro_recurrent_net}) after $T$
		time-steps. Then, for any set of template vectors $\x^{(1)},\ldots,\x^{(M)} \in \X$ it
		holds that:
		\begin{equation}
		\sep{S,E}{y^{T,L, \Theta}_c}\geq \rank{\mat{\A(y^{T,L, \Theta}_c)}_{S,E}},
		\end{equation}
		where $\A(y^{T,L, \Theta}_c)$ is the grid tensor of $y^{T,L, \Theta}_c$ \wrt~the above template vectors, and $\mat{\A(y^{T,L, \Theta}_c)}_{S,E}$ is the matricization (See section \ref{intro:tensors:matricization}) of $\A(y^{T,L, \Theta}_c)$ \wrt~the Start-End partition.
	\end{claim}
	\ifdefined\SQUEEZE \vspace{-4mm} \fi
	\begin{proof}
		If $\sep{S,E}{y_c^{T,L, \Theta}} = \infty$ then the inequality is trivially
		satisfied. Otherwise, assume that
		$\sep{S,E}{y_c^{T,L, \Theta}} = K \in \N$, and let $\{g_i^s, g_i^e\}_{i=1}^K$
		be the functions of the respective decomposition to a sum of separable
		functions, i.e. that the following holds:
		\begin{align*}
		y_c^{T,L, \Theta}(\x^1,\ldots,\x^T)
		&= \sum_{\nu=1}^K g_\nu^s(\x^1,\ldots,\x^{T/2})
		\cdot g_\nu^e(\x^{T/2+1},\ldots,\x^T).
		\end{align*}
		Then, by definition of the grid tensor, for any template vectors $\x^{(1)},\ldots,\x^{(M)}\in \X$ the following
		equality holds:
		\begin{align*}
		\A(y_c^{T,L, \Theta})_{d_1,\ldots,d_N} &=
		\sum_{\nu = 1}^K g_\nu^s(\x^{(d_1)},\ldots,\x^{(d_{T/2})})
		\cdot g_\nu^e(\x^{(d_{T/2+1})},\ldots,\x^{(d_T)}) \\
		&\equiv \sum_{\nu=1}^K V^\nu_{d_1,\ldots,d_{T/2}} U^\nu_{d_{T/2+1},\ldots,d_T},
		\end{align*}
		where $V^\nu$ and $U^\nu$ are the tensors holding the values of $g_\nu^s$
		and $g_\nu^e$, respectively, at the points defined by the template vectors.
		Under the matricization according to the $(S,E)$ partition, it holds that
		$\mat{V^\nu}_{S,E}$ and $\mat{U^\nu}_{S,E}$ are column and row vectors,
		respectively, which we denote by $\vv_\nu$ and $\uu_\nu^T$. It follows that the
		matricization of the grid tensor is given by:
		\begin{align*}
		\mat{\A(y_c^{T,L, \Theta})}_{S,E} &= \sum_{\nu=1}^K \vv_\nu \uu_\nu^T,
		\end{align*}
		which means that
		$\rank{\mat{\A(y_c^{T,L, \Theta})}_{S,E}}\leq K=\sep{S,E}{y_c^{T,L, \Theta}}$.
	\end{proof}
	\vspace{2mm}
	
	Claim~\ref{claim:grid_sep_deep} will allow us to provide a lower bound on the
	Start-End separation rank of functions realized by deep RACs, which we show to
	be significantly higher than the Start-End separation rank of functions realized
	by shallow RACs (to be obtained via Claim~\ref{claim:grid_sep_shallow}). Thus, in the next section, we employ the above presented tools to
	show that a compelling enhancement of the Start-End separation
	rank is brought forth by depth in recurrent networks.
\section{Analyzing the Long-Term Memory of Recurrent Networks }\label{sec:theory:bounds}
In this section, we present the main theoretical contributions of this work.
	In Section~\ref{sec:results:proof}, we formally present a result which clearly separates between the memory capacity of a deep ($L=2$) recurrent network and a shallow ($L=1$) one.
	Following the formal presentation of results in
	Theorem~\ref{theorem:main_result}, we discuss some of their implications and
	then conclude by sketching a proof outline for the theorem (full proof is
	relegated to Appendix~\ref{app:proofs:main_result}). In Section~\ref{sec:results:conjecture}, we present a quantitative conjecture regarding the enhanced memory capacity of deep recurrent networks of general depth $L$, which relies on the inherent combinatorial properties of the recurrent network's computation. We leave the formal proof of this conjecture for future work.
	
	\ifdefined\SQUEEZE\vspace{-2mm}\fi
	\subsection{Separating Between Shallow and Deep Recurrent Networks}\label{sec:results:proof}
	\ifdefined\SQUEEZE\vspace{-2mm}\fi
	
	Theorem~\ref{theorem:main_result} states, that the dependencies modeled between
	the beginning and end of the input sequence to a recurrent network, as measured
	by the Start-End separation rank (see Section~\ref{sec:corr:sep}), can be considerably
	more complex for deep networks than for shallow ones:

	\begin{theorem}\label{theorem:main_result}\label{THEOREM:MAIN_RESULT}
		Let $y^{T,L, \Theta}_c$ be the function computing the $c$'th class score after
		$T$ time-steps of an RAC with $L$ layers, $R$ hidden channels per layer,
		weights denoted by $\Theta$, and initial hidden states $\h^{0,l},~l\in[L]$ (Figure~\ref{fig:intro_recurrent_net} with
		$g=g^{\mathrm {RAC}}$). Assume that the network's initial mapping functions~$\{f_{d}\}_{d=1}^M$ are
		linearly independent and square integrable. Let $\sep{S,E}{y^{T,L, \Theta}_c}$ be the Start-End
		separation rank of $y^{T,L, \Theta}_c$ [Equation~\eqref{eq:sep_rank}]. Then, the
		following holds almost everywhere, \ie~for all values of the parameters $\Theta\times\h^{0,l}$ but a set of
		Lebesgue measure zero:
		
		\begin{enumerate}
			\ifdefined\SQUEEZE \vspace{-2mm}\fi
			\item $\sep{S,E}{y^{T,L, \Theta}_c} = \min\left\{R, M^{\nicefrac{T}{2}} \right\}$, for $L=1$ (shallow network).
			\ifdefined\SQUEEZE	\vspace{-3mm}\fi
			\item $\sep{S,E}{y^{T,L, \Theta}_c} \geq~\multiset{\min\{M,R\}}{T/2} $, for $L =2$ (deep network),
		\end{enumerate}
		\ifdefined\SQUEEZE	\vspace{-3mm}\fi
		where $\multiset{\min\{M,R\}}{T/2}$ is the multiset coefficient, given in the binomial form by
		$\binom{\min\{M,R\}+T/2-1}{T/2}$.
	\end{theorem}
	\ifdefined\SQUEEZE\vspace{-2mm}\fi
	
	The above theorem readily implies that depth entails an enhanced ability of
	recurrent networks to model long-term temporal dependencies in the sequential
	input.
	Specifically, Theorem~\ref{theorem:main_result} indicates depth efficiency -- it
	ensures us that upon randomizing the weights of a deep RAC with $R$ hidden
	channels per layer, with probability $1$ the function realized by it after $T$
	time-steps may only be realized by a shallow RAC with a number of hidden
	channels that is combinatorially large. Stated alternatively, this means that
	almost all functional dependencies which lie in the hypothesis space of deep
	RACs with $R$ hidden channels per layer, calculated after $T$ time-steps, are
	inaccessible to shallow RACs with less than a super-linear number of
	hidden channels. Thus, a shallow recurrent network would require an impractical
	amount of parameters if it is to implement the same
	function as a deep recurrent network.
	
	The established role of the Start-End separation rank as a dependency measure between the
	beginning and the end of the sequence (see Section~\ref{sec:corr:sep}), implies
	that these functions, which are realized by almost any deep network and can
	never be realized by a shallow network of a reasonable size, represent more
	elaborate dependencies over longer periods of time.
	The above notion is strengthened by the fact that the Start-End separation rank
	of deep RACs increases with the sequence length $T$, while the Start-End
	separation rank of shallow RACs is independent of it.
	This indicates that shallow recurrent networks are much more restricted in
	modeling long-term dependencies than the deep ones, which enjoy a combinatorially
	increasing Start-End separation rank as time progresses.
	Below, we present an outline of the proof for Theorem~\ref{theorem:main_result}
	(see Appendix~\ref{app:proofs:main_result} for the full proof):
	
	\vspace{2mm}
	\ifdefined\SQUEEZE\vspace{-3mm}\fi
	\begin{proof}[Proof sketch of Theorem~\ref{theorem:main_result}]
		\leavevmode
		\ifdefined\SQUEEZE	\vspace{-1mm}\fi
		\begin{enumerate}
			\item For a shallow network, Claim~\ref{claim:grid_sep_shallow} establishes
			that the Start-End separation rank of the function realized by a shallow
			($L=1$) RAC is equal to the rank of the matrix obtained by matricizing
			the corresponding shallow RAC weights tensor [Equation~\eqref{eq:score}]
			according to the Start-End partition: $\sep{S,E}{y^{T,1, \Theta}_c}= \rank{\mat{\A^{T,1, \Theta}_c}_{S,E}}$.
			Thus, it suffices to prove that $\rank{\mat{\A^{T,1, \Theta}_c}_{S,E}}=R$
			in order to satisfy bullet
			(1) of the theorem, as the rank is trivially upper-bounded by the
			dimension of the matrix, $M^{\nicefrac{T}{2}}$. To this end, we call
			upon the TT-decomposition of $\A^{T,1, \Theta}_c$, given by
			Equation~\eqref{eq:tt_decomp}, which
			corresponds to the MPS Tensor Network presented in Appendix~\ref{app:rac_tns}. We rely on a recent result by \citep{levine2018deep},
			who state that the rank of the matrix obtained by matricizing any tensor
			according to a partition $(S,E)$, is equal to a minimal cut separating $S$
			from $E$ in the Tensor Network graph representing this tensor. The
			required equality follows from the fact that the TT-decomposition in
			Equation~\eqref{eq:tt_decomp} is of TT-rank $R$, which in turn implies that the
			min-cut in the appropriate Tensor Network graph is equal to $R$.
			\item For a deep network, Claim~\ref{claim:grid_sep_deep} assures us that the
			Start-End separation rank of the function realized by a depth $L=2$ RAC
			is lower bounded by the rank of the matrix obtained by the corresponding
			grid tensor matricization: $\sep{S,E}{y^{T,2, \Theta}_c}\geq \rank{\mat{\A(y^{T,2, \Theta}_c)}_{S,E}}$.
			Thus, proving that
			$\rank{\mat{\A(y^{T,2, \Theta}_c)}_{S,E}} \geq \multiset{\min\{M,R\}}{T/2}$ for
			all of the values of parameters $\Theta\times\h^{0,l}$ but a set of Lebesgue measure
			zero, would satisfy the theorem.
			We use a lemma proved in
			\citep{sharirtractable}, which states that since the entries of
			$\A(y^{T,2, \Theta}_c)$ are polynomials in the deep recurrent network's
			weights, it suffices to find a single example for which the rank of the
			matricized grid tensor is greater than the desired lower bound. Finding
			such an example would indeed imply that for almost all of the values of
			the network parameters, the desired inequality holds.
			
			We choose a weight assignment such that the resulting matricized grid
			tensor resembles a matrix obtained by raising a rank-$\bar{R}\equiv\min\{M,R\}$ matrix to the
			Hadamard power of degree $T/2$. This operation, which raises each
			element of the original rank-$\bar{R}$ matrix to the power of $T/2$, was shown
			to yield a matrix with a rank upper-bounded by the multiset
			coefficient $\multiset{\bar{R}}{T/2}$ (see \eg~\citep{amini2012low}). We show
			that our assignment results in a matricized grid tensor with a rank
			which is not only upper-bounded by this value, but actually achieves it. Under our assignment, the matricized grid tensor takes the form: 
			\begin{equation*}
			\mat{\A(y^{T,2, \Theta}_c)}_{S,E}=
			\sum_{\substack{\p\in \state{{\bar{R}}}{\nicefrac{T}{2}}}}
			U_{S\p} \cdot
			V_{\p E},
			\end{equation*}
			where the set  \begin{equation*}	\state{{\bar{R}}}{\nicefrac{T}{2}} \equiv \{\p \in (\N \cup \{0\})^{\bar{R}} |\sum_{i=1}^{\bar{R}} p_i = \nicefrac{T}{2}\}
			\end{equation*}
			can be viewed as the set of all possible states of a bucket containing $\nicefrac{T}{2}$
			balls of ${\bar{R}}$ colors, where $p_r$ for $r\in[\bar{R}]$ specifies the number of balls
			of the $r$'th color. By definition: $\abs{\state{{\bar{R}}}{\nicefrac{T}{2}}}=\multiset{\bar{R}}{\nicefrac{T}{2}}$ and $\abs{S}=\abs{E}=M^{\nicefrac{T}{2}}$, therefore the matrices $U$ and $V$ uphold: $U\in\R^{M^{\nicefrac{T}{2}}\times\multiset{\bar{R}}{\nicefrac{T}{2}}}$; $V\in\R^{\multiset{\bar{R}}{\nicefrac{T}{2}}\times M^{\nicefrac{T}{2}}}$, and for the theorem to follow we must show that they both are of rank $\multiset{\bar{R}}{\nicefrac{T}{2}}$ (note that $\multiset{\bar{R}}{\nicefrac{T}{2}}\leq\multiset{M}{\nicefrac{T}{2}}<M^{\nicefrac{T}{2}}$).
			
			We observe the sub-matrix $\bar{U}$ defined by the subset of the rows of $U$ such that we select the row $d_1\ldots d_{\nicefrac{T}{2}}\in S$ only if it upholds
			that $\forall j, d_j \leq d_{j+1}$. Note that there are exactly $\multiset{\bar{R}}{\nicefrac{T}{2}}$ such rows, thus $\bar{U}\in\R^{\multiset{\bar{R}}{\nicefrac{T}{2}}\times\multiset{\bar{R}}{\nicefrac{T}{2}}}$ is a square matrix.
			Similarly we observe a sub-matrix of $V$ denoted $\bar{V}$, for which we select the column $d_{\nicefrac{T}{2}+1}\ldots d_T \in E$ only if it upholds
			that $\forall j, d_j \leq d_{j+1}$, such that it is also a square matrix. Finally, by employing a variety of technical lemmas, we show that the determinants of these square matrices are non vanishing under the given assignment, thus satisfying the theorem. 
			
		\end{enumerate}
		\ifdefined\SQUEEZE	\vspace{-3mm}\fi
	\end{proof}

	\subsection{Increase of Memory Capacity with Depth}\label{sec:results:conjecture}
	\ifdefined\SQUEEZE\vspace{-1mm}\fi
	
	Theorem~\ref{theorem:main_result} provides a lower bound of
	$\multiset{R}{\nicefrac{T}{2}}$ on the Start-End separation rank of depth $L=2$
	recurrent networks, combinatorially separating deep recurrent networks from shallow ones.
	By a trivial assignment of weights in higher layers, the Start-End separation rank of even
	deeper recurrent networks ($L>2$) is also lower-bounded by this expression, which does not
	depend on $L$. In the following, we conjecture that a tighter lower bound holds for
	networks of depth $L>2$, the form of which implies that the memory capacity of deep recurrent networks
	grows combinatorially with the network depth:
	
	\begin{conjecture}\label{conjecture:high_L}\label{CONJECTURE:HIGH_L}
		Under the same conditions as in Theorem~\ref{theorem:main_result}, for all values of
		$\Theta\times\h^{0,l}$ but a set of Lebesgue measure zero, it holds for any $L$ that:
		\ifdefined\SQUEEZE	\vspace{-2mm}\fi
		\begin{align*}
		\sep{S,E}{y^{T,L, \Theta}_c}&\geq \min\left\{\multiset{\min\{M,R\}}{\multiset{T/2}{L-1}}, M^{\nicefrac{T}{2}} \right\}.
		\end{align*}
	\end{conjecture}
	
	We motivate Conjecture~\ref{conjecture:high_L} by investigating the combinatorial nature
	of the computation performed by a deep RAC. By constructing Tensor Networks which correspond to deep RACs, we attain an informative visualization of this combinatorial perspective.
	In Appendix~\ref{app:rac_tns}, we provide full details of this Tensor Networks construction and present
	the formal motivation for the conjecture in Appendix~\ref{app:rac_tns:conjecture}. Below, we qualitatively outline
	our approach.
	
\begin{figure}
		\centering
		\includegraphics[width=1\linewidth]{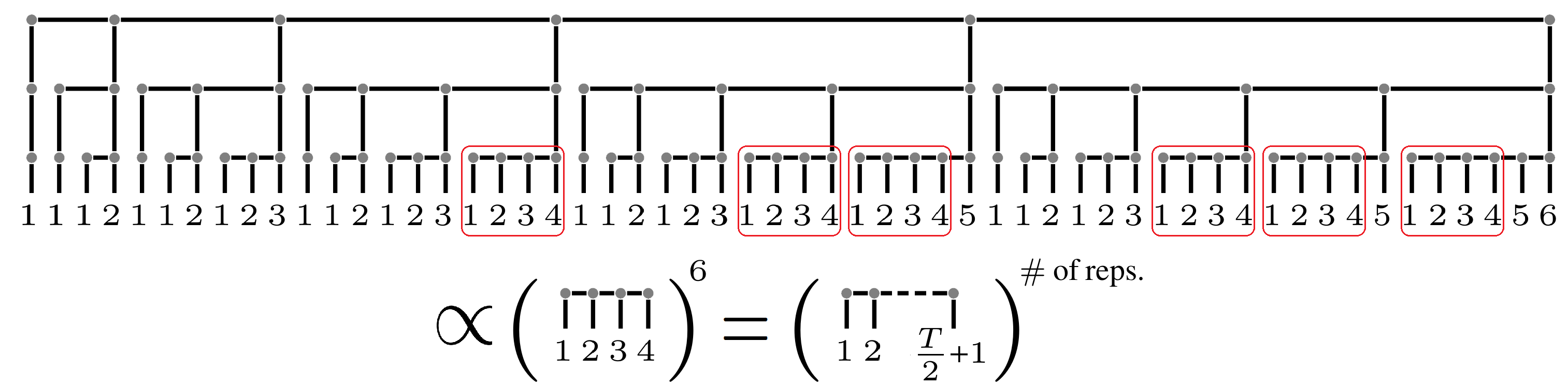}
		\caption{Tensor Network representing the computation of a depth $L=3$ RAC after $T=6$ time-steps. See construction in Appendix~\ref{app:rac_tns}. The number of repetitions of the basic unit cell connecting `Start' and `End' inputs in the Tensor Network graph gives rise to the lower bound in Conjecture~\ref{conjecture:high_L}.}
		\label{fig:conjecture}
		\ifdefined\SQUEEZE	\vspace{-5mm}\fi
	\end{figure}

	A Tensor Network is essentially a graphical tool for representing algebraic operations
	which resemble multiplications of vectors and matrices, between higher order tensors.
	Figure~\ref{fig:conjecture} (top) shows an example of the Tensor Network representing the computation of a depth $L=3$ RAC after
	$T=6$ time-steps. This well-defined computation graph hosts the values of the weight
	matrices at its nodes. The inputs $\{x^1,\ldots, x^T\}$ are marked by their corresponding time-step $\{1,\ldots, T\}$, and are integrated in a depth dependent and time-advancing manner (see further discussion regarding this form in Appendix~\ref{app:rac_tns:deep}), as portrayed in the example of Figure~\ref{fig:conjecture}. We highlight in red the basic unit in
	the Tensor Network which connects ``Start" inputs $\{1,\ldots,\nicefrac{T}{2}\}$ and ``End" inputs $\{\nicefrac{T}{2}+1,\ldots,T\}$. In order to estimate
	a lower bound on the Start-End separation rank of a depth $L>2$ recurrent network, we employ a
	similar strategy to that presented in the proof sketch of the $L=2$ case (see
	Section~\ref{sec:results:proof}). Specifically, we rely on the fact that it is sufficient
	to find a specific instance of the network parameters $\Theta\times\h^{0,l}$ for which
	$\mat{\A(y^{T,L, \Theta}_c)}_{S,E}$ achieves a certain rank, in order for this rank to
	bound the Start-End separation rank of the network from below.
	
	Indeed, we find a specific assignment of the network weights, presented in
	Appendix~\ref{app:rac_tns:conjecture}, for which the Tensor Network effectively takes the form of the
	basic unit connecting ``Start" and ``End", raised to the power of the number of its
	repetitions in the graph (bottom of Figure~\ref{fig:conjecture}). This basic unit corresponds to a simple computation
	represented by a grid tensor with Start-End matricization of rank $R$. Raising such a
	matrix to the Hadamard power of any $p\in\mathbb{Z}$, results in a matrix with a rank
	upper bounded by $\multiset{R}{p}$, and the challenge of
	proving the conjecture amounts to proving that the upper bound is tight in this case.
	In Appendix~\ref{app:rac_tns:conjecture}, we prove that the number of repetitions of the basic unit connecting ``Start" and ``End" in the deep RAC Tensor Network graph, is exactly equal to
	$\multiset{T/2}{L-1}$ for any depth $L$. For example, in the $T=6,L=3$
	network illustrated in Figure~\ref{fig:conjecture}, the number of repetitions indeed corresponds to
	$p=\multiset{3}{2}=6$. It is noteworthy that for $L=1,2$~ the bound in
	Conjecture~\ref{conjecture:high_L} coincides with the bounds that were proved for these depths in
	Theorem~\ref{theorem:main_result}.
	
	Conjecture~\ref{conjecture:high_L} indicates that beyond the proved combinatorial advantage
	in memory capacity of deep networks over shallow ones, a further combinatorial separation
	may be shown between recurrent networks of different depths. We leave the proof of this
	result, which can reinforce and refine the understanding of advantages brought forth by
	depth in recurrent networks, as an open problem.
	In the following, we empirically investigate the theoretical outcomes presented in this section.

\clearpage
%
%
\let\textcircled=\pgftextcircled
\chapter{Experimental Evaluation}
\label{chap:experiments}
\textit{All experiments presented in this chapter can be reproduced using the source code publicly available in \url{https://github.com/HUJI-Deep/Long-Term-Memory-of-Deep-RNNs}.}
\initial{I}n this chapter, we provide an empirical demonstration supporting the theoretical findings of Chapter ~\ref{chapter:theory}. 
The results above are formulated for the class of RACs (presented in Section~\ref{sec:theory:RAC}), and the experiments presented hereinafter demonstrate their extension to more commonly used RNN architectures. 
As noted in Chapter~\ref{chapter:intro}, the advantage of deep recurrent networks over shallow ones is well established empirically, as the best results on various sequential tasks have been achieved by stacking recurrent layers~\citep{cirecsan2010deep,mohamed2012acoustic,graves2013speech}. 
Below, we focus on two tasks which highlight the `long-term memory' demand of recurrent networks, and show how depth empowers the network's ability to express the appropriate distant temporal dependencies. 

We address two synthetic problems. The first is the Copying Memory Task, to be
described in Section~\ref{sec:exp:copy}, which was previously used to test
proposed solutions to the gradient issues of backpropagation through
time~\citep{hochreiter1997long,martens2011learning,arjovsky2016unitary,wisdom2016full,jing2016tunable}.
We employ this task as a test for the recurrent network's expressive ability to
`remember' information seen in the distant past. 
The second task is referred to as the Start-End Similarity Task, to be described
in Section~\ref{sec:exp:sim}, which is closely related to the Start-End
separation rank measure proposed in Section~\ref{sec:theory:separation_rank}. 
In both experiments we use a successful RNN variant referred to as Scaled Cayley Orthogonal Recurrent Neural Network (scoRNN)~\citep{pmlr-v80-helfrich18a}, which was
shown to enable efficient optimization without the need to use gating units such as in LSTM networks. Moreover, scoRNNs are known to perform exceptionally well on the Copying Memory Task.
Specifically, we use scoRNN with $\rho$ (see \citep{pmlr-v80-helfrich18a} Section 3.1) set to \emph{$\nicefrac{R}{2}$}. Under the notations we presented
in section~\ref{sec:theory:RAC} and portrayed in Figure~\ref{fig:intro_recurrent_net},
scoRNNs employ $g^\mathrm{RNN}(\aaa, \bb) = \sigma(\aaa + \bb)$, where
$\sigma(\cdot)$ is the modReLU function (\citep{pmlr-v80-helfrich18a} Section 4.1),
and the matrices $W^{\mathrm{H},l}$ are restricted to being orthogonal.

Throughout both experiments we use RMSprop~\citep{tieleman2012lecture} as the
optimization algorithm, where we took the best of several learning rates in the range $[1e-5,1e-3]$, and with the default value $0.9$ for the moving average discount factor. Our focus is on the expressive ability of each network, rather than on generalization, so instead of holding a fixed training set we perform each optimization step on a batch of fresh 128 examples drawn directly from the data distribution. We train each configuration for a maximum of 200K iterations, and we use an early stopping criteria defined by the loss over a fixed validation set of 5K examples. The final evaluation is done on a held-out test set of size 10K.

The methodology we employ in the experiments below is aimed at testing the
following practical hypothesis, which is commensurate with the theoretical
outcomes in Section~\ref{sec:theory:bounds}: \emph{Given a certain resource budget for
	a recurrent network that is intended to solve a `long-term memory problem',
	adding recurrent layers is significantly preferable to increasing the number of
	channels in existing layers}. Specifically, we train RNNs of depths $1$, $2$, and
$3$ over increasingly hard variants of each problem (requiring longer-term
memory), and report the maximal amount of memory capabilities for each
architecture in Figures~\ref{fig:copy} and~\ref{fig:sim}. 

\section{Copying Memory Task}\label{sec:exp:copy}

In the Copying Memory Task, the network is required to memorize a sequence of characters of a fixed length $m$, and then to reproduce it after a long lag of $B$ time-steps, known as the \emph{delay time}. The input sequence is composed of characters drawn from a given alphabet $\left\{ a_{i}\right\} _{i=1}^{n}$, and two special symbols: a \emph{blank} symbol denoted by `\texttt{\_}', and a \emph{trigger} symbol denoted by `\texttt{:}'. The input begins with a string of $m$ \emph{data characters} randomly drawn from the alphabet, and followed by $B$ occurrences of blank symbols. On the $m$'th before last time-step the trigger symbol is entered, signaling that the data needs to be presented. Finally the input ends with an additional $m-1$ blank characters. In total, the sequence length is $T=B+2m$. The correct sequential output of this task is referred to as the target. The target character in every time-step is always the blank character, except for the last $m$ time-steps, in which the target is the original $m$ data characters of the input. For example, if $m=3$ and $B=5$, then a legal input-output pair could be  ``\texttt{ABA\_\_\_\_\_:\_\_}'' and ``\texttt{\_\_\_\_\_\_\_\_ABA}'', respectively.

\begin{figure}
	\centering
	\includegraphics[width=\linewidth]{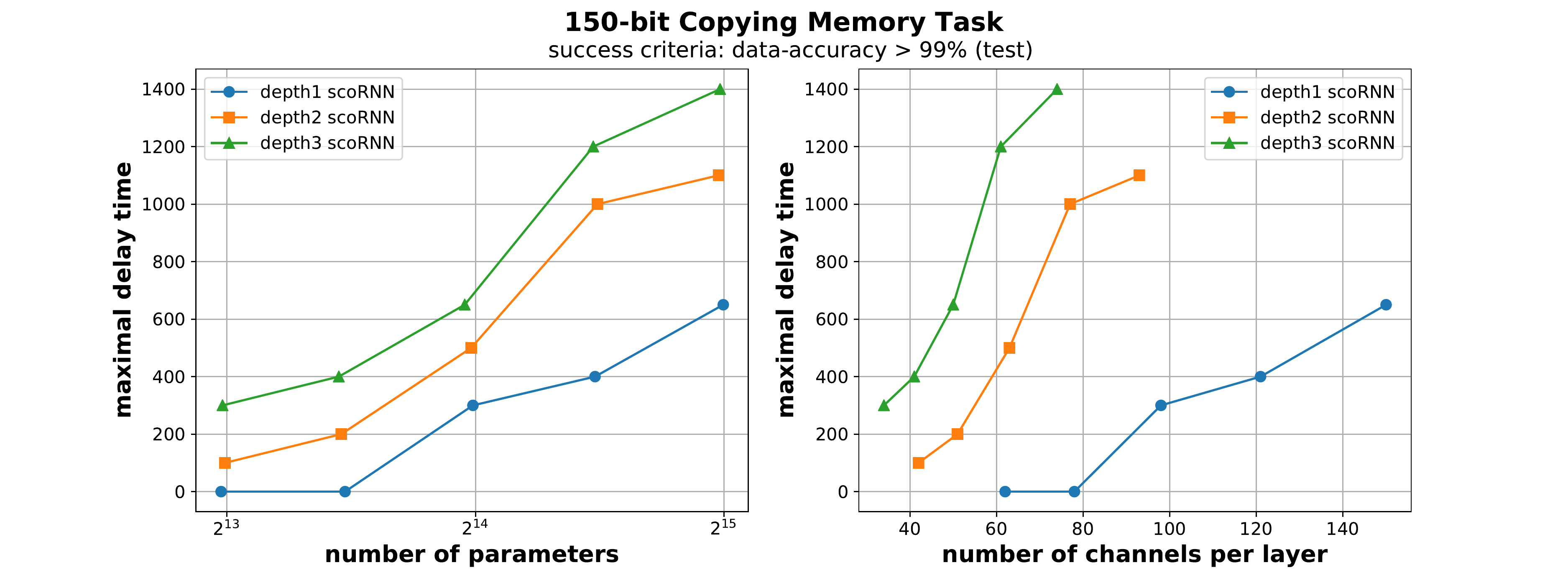}
	\includegraphics[width=\linewidth]{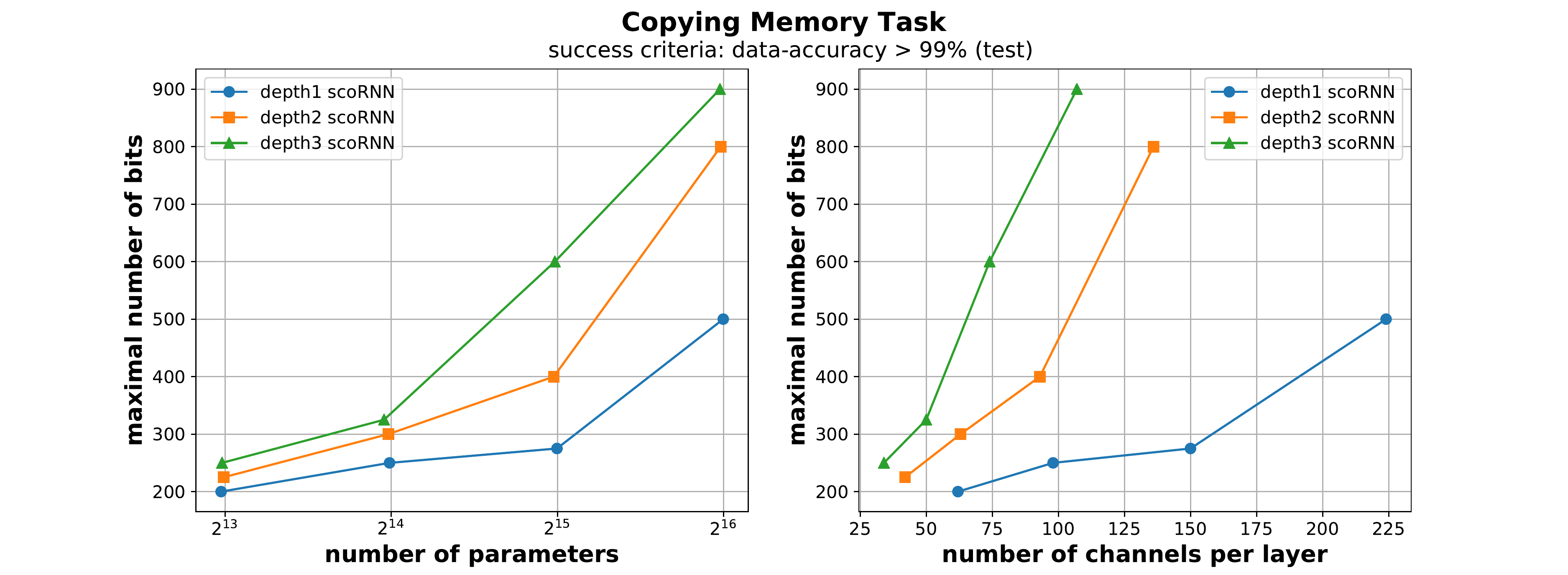}
	\caption{
		Results of the Copying Memory Task, as defined in Section~\ref{sec:exp:copy}. The results are shown for networks of depths $1,2,3$ and sizes $2^{13}$~-~$2^{16}$ (measured in the number of parameters). 
		We define success in the Copying Memory Task as achieving a data-accuracy $> 99\%$, \ie~being able to reproduce each character of the copied data after a given delay time with more than $99\%$ accuracy. For each network architecture, the plots report the longest delay time (top) and the maximal number of bits (bottom) for which the architecture has been successful on the test set as a function of network size (left) and number of channels per layer (right). In the first experiment (up) we tested the performance on delay times up to $1500$, sampling delay times of $0,50,100,150,...,1000$ and then in intervals of $100$, while the second experiment (bottom) was evaluated using data lengths of 30 to 200, sampled in intervals of 5. The advantage of deepening the network is evident, as for each tested network size, the recurrent network of depth $3$ outperforms the recurrent network of depth $2$, which outperforms the recurrent network of depth $1$. For a case of limited amount of resources \wrt~the task hardness, which occurs in the smaller network sizes, shallower networks cannot reproduce the given sequence for any delay time larger than zero, where deeper networks succeed. The displayed results clearly highlight the augmenting contribution of depth to the recurrent network's long-term memory capacity.
    \label{fig:copy}}
\end{figure}

In essence, the data length $m$ and alphabet size $n$ control the number of bits to be memorized, and the delay time controls the time these bits need to stay in memory~--~together these parameters control the hardness of the task. Previous works have used values such as $m=10$ and $n=8$~\citep{arjovsky2016unitary} or similar, which amount to memorizing $30$ bits of information, for which it was demonstrated that even shallow recurrent networks are able to solve this task for delay times as long as $B=1000$ or more. To allow us to properly separate between the performance of networks of different depths, we consider much harder variants of the problem. The flexibility in how to control the problem hardness allows us to scrutinize two different perspectives of long-term memory, namely the distance in time of the information to be memorized and the memory capacity, by performing two experiments. In the first experiment (top of Figure \ref{fig:copy}) we fix the data length and the alphabet size to $m=30$ and $n=32$ respectively, which requires memorizing $150$ bits of information, and train each architecture again and again, each time with a larger delay time $B$, until its expressive limit is found (see Figure \ref{fig:copy}~for further details of the success criteria). In the second experiment, the data length $m$ is used as the varying parameter, while the alphabet size and the delay time are kept fixed to $n=32$ and $B=0$ respectively. Our method is still to seek for the expressive limit of each network architecture, but this time we measure it by the number of bits to be memorized, essentially representing the memory capacity (bottom of Figure \ref{fig:copy}).

We present the results for this task in Figure~\ref{fig:copy}, where we compare the performance for networks of depths $1,2,3$ and of size in the range of $2^{13}$~-~$2^{16}$, measured in the number of parameters. Our measure of performance in the Copy Memory Task is referred to as the \emph{data-accuracy}, calculated as $\frac{1}{N}\frac{1}{m}\sum_{j=1}^{N}\sum_{t=m+B+1}^{T}\mathbf{1}[\hat{O}_t^j=O_t^j]$, where $N$ is the sample size, $O_t^j$ the correct output character at time $t$ for example $j$, and $\hat{O}_t^j=\argmax_{i\in[n+2]}{\y_t^j}$ the predicted character. The data-accuracy effectively reflects the per-character data reproduction ability, therefore it is defined only over the final $m$ time-steps when the memorized data is to be reproduced. 
In the top of Figure~\ref{fig:copy}~we display for each network the longest delay time for which it is able to solve the task, and in the bottom of the same figure we display the maximal number of bits each architecture is able to memorize, both demonstrating a clear advantage of depth for this task. In other words, given an amount of resources, it is advantageous to allocate them in a stacked layer fashion for this long-term memory based task.

\section{Start-End Similarity Task}\label{sec:exp:sim}
The Start-End Similarity Task directly tests the
recurrent network's expressive ability to integrate
between the two halves of the input sequence. In this
task, the network needs to determine how similar the two halves are. 
The input is a sequence of $T$ characters
$\{x_t\}_{t=1}^T$ drawn from a given alphabet
$\left\{ a_{i}\right\} _{i=1}^{n}$ and an additional special \emph{blank} symbol denoted by '\texttt{\_}'. The first $\nicefrac{T}{2}$ characters are denoted by `Start' and the rest by `End', similarly to previous sections. 
Consider the process of generating a single example. Two strings of a fixed size \emph{$m < \nicefrac{T}{2}$}, denoted by \emph{$s_{1}$} and \emph{$s_{2}$}, are drawn from the set of alphabetic characters. These two strings are then placed contiguously in the input sequence, \emph{s.t.} \emph{$s_{1}$} begins in a random position in 'Start' and \emph{$s_{2}$} begins in the corresponding position in 'End'. More specifically, the first character of \emph{$s_{1}$} is placed in the random index \emph{$t$} $\sim Uni\{1,2,...,\nicefrac{T}{2}-m\}$ and the first character of \emph{$s_{2}$} is then placed in the index \emph{$t+\nicefrac{T}{2}$}. Every other position is occupied by the blank character '\texttt{\_}'

Considering pairs of characters in the same
relative position in `Start' and `End', \ie~the alphabetic pairs
$\left(x_{t},x_{t+\nicefrac{T}{2}}\right)$, we divide
each input sequence into one of the following classes: 

\begin{itemize}
	\item \textbf{1-similar}: `Start' and `End' contains the same $m$ length sub-string, \eg~``\texttt{\_\_\_ABBA\_\_\_\_\_\_ABBA\_\_\_}''.
	\item \textbf{0.5-similar}: `Start' and `End' have
	exactly $\nicefrac{m}{2}$ matching pairs of
	characters (a randomly positioned half of the
	sub-string is identical, and the other half is not), \eg~``\texttt{\_\_\_\_\_ABBA\_\_\_\_\_\_BABA\_}''.
	\item \textbf{0-similar}: no pair of alphabetic characters
	$\left(x_{t},x_{t+\nicefrac{T}{2}}\right)$ match, \eg~``\texttt{\_\_\_ABBA\_\_\_\_\_\_BAAB\_\_\_}''. 
\end{itemize}

\begin{figure}
	\centering
	\includegraphics[width=\linewidth]{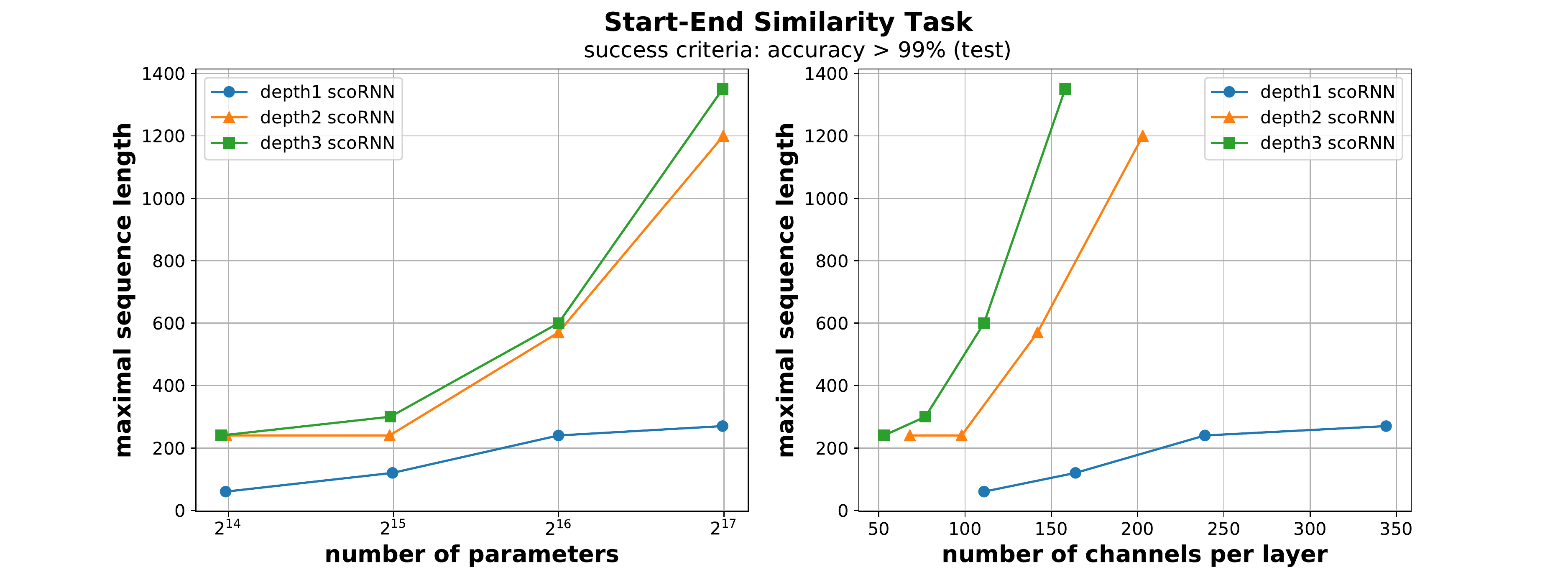}
	\caption{Results of the Start-End Similarity Task, as defined in Section~\ref{sec:exp:sim}. The results are shown for networks of depths $1,2,3$ and sizes $2^{14}$~-~$2^{17}$. We define success on the Start-End similarity task as test accuracy $> 99\%$. For each network architecture, the plots report the longest input sequence length for which the architecture has been successful as a function of network size (left) and number of channels per layer (right). We tested the performance on sequences with 60 to 1500 characters in intervals of 30. It can be seen that for every given network size, a deeper network can model long-term dependencies more successfully than a shallower one. For example, a depth-$3$ network succeeds at solving the Start-End Similarity task for $T=1350$ while a depth-$1$ network succeeds only for $T=270$.\label{fig:sim}}
\end{figure}

The task we examine is a classification task of a dataset 
distributed uniformly over these three classes. Here, the
recurrent networks are to produce a meaningful output
only in the last time-step, determining in which class
the input was, \ie~how similar the beginning of the input
sequence is to its end. The hardness of the problem is controlled by the alphabet size $n$, the length of the alphabetic sub-strings $m$, and the input sequence length $T$. We fixed $m=30$ and $n=32$, and used different values of $T$ to increasingly make the problem harder.
Figure~\ref{fig:sim} shows the performance for networks of depths $1,2,3$ and sizes $2^{14}$~-~$2^{17}$, measured in the number of parameters, on the Start-End Similarity Task. The clear advantage of depth is portrayed in this task as well, empirically demonstrating the enhanced ability of deep recurrent networks to model long-term elaborate dependencies in the input string.

Overall, the empirical results presented in this chapter reflect well our theoretical findings, presented in section ~\ref{sec:theory:bounds}.
\clearpage
\let\textcircled=\pgftextcircled
\chapter{Conclusion}

\initial{T}he notion of depth efficiency, by which deep networks efficiently express functions that would require shallow networks to have a super-linear size, is well established in the context of convolutional networks.
However, recurrent networks differ from convolutional networks, as they are suited by design to tackle inputs of varying lengths.
Accordingly, depth efficiency alone does not account for the remarkable performance of deep recurrent networks on long input sequences.
In this work, we identified a fundamental need for a quantifier of `time-series expressivity', quantifying the memory capacity of recurrent networks, which can account for the empirically undisputed advantage of depth in hard sequential tasks.
In order to meet this need, we proposed a measure of the ability of recurrent networks to model long-term temporal dependencies, in the form of the Start-End separation rank.
The separation rank was used to quantify dependencies in convolutional networks, and has roots in the field of quantum physics.
The proposed Start-End separation rank measure adjusts itself to the temporal extent of the input series, and quantifies the ability of the recurrent network to correlate the incoming sequential data as time progresses.

We analyzed the class of Recurrent Arithmetic Circuits, which are closely related to successful RNN architectures, and proved that the Start-End separation rank of deep RACs increases combinatorially with the number of channels and as the input sequence extends, while that of shallow RACs increases linearly with the number of channels and is independent of the input sequence length.
These results, which demonstrate that depth brings forth an overwhelming advantage in the ability of recurrent networks to model long-term dependencies, were achieved by combining tools from the fields of measure theory, tensorial analysis, combinatorics, graph theory and quantum physics. 
The above presented empirical evaluations support our theoretical findings, and provide a demonstration of their relevance for commonly used classes of recurrent networks.

Such analyses may be readily extended to other architectural features employed in modern recurrent networks.
Indeed, the same time-series expressivity question may now be applied to the different variants of LSTM networks, and the proposed notion of Start-End separation rank may be employed for quantifying their memory capacity.
We have demonstrated that such a treatment can go beyond unveiling the origins of the success of a certain architectural choice, and leads to new insights.
The above established observation that dependencies achievable by vanilla shallow recurrent network do not adapt at all to the sequence length, is an exemplar of this potential.

Moreover, practical recipes may emerge by such theoretical analyses.
The experiments preformed in \citep{hermans2013training}, suggest that shallow layers of recurrent networks are related to short time-scales, \eg~in speech: phonemes, syllables, words, while deeper layers appear to support dependencies of longer time-scales, \eg~full sentences, elaborate questions.
These findings open the door to further depth related investigations in recurrent networks, and specifically the role of each layer in modeling temporal dependencies may be better understood.
\citep{levine2018deep} establish theoretical observations which translate into practical conclusions regarding the number of hidden channels to be chosen for each layer in a deep convolutional network.
The conjecture presented in this work, by which the Start-End separation rank of recurrent networks grows combinatorially with depth, can similarly entail practical recipes for enhancing their memory capacity. Such analyses can lead to a profound understanding of the contribution of deep layers to the recurrent network's memory. Indeed, we view this work as an important step towards novel methods of matching the recurrent network architecture to the temporal dependencies in a given sequential dataset.
\clearpage
%
%
\appendix
%
%

\chapter{Tensor Network Representation of Recurrent Arithmetic circuits}
\label{app:rac_tns}


\initial{I}n this appendix, we expand our algebraic view on recurrent networks and make use
of a graphical approach to tensor decompositions referred to as Tensor Networks
(TNs). The tool of TNs is mainly used in the many-body quantum physics literature
for a graphical decomposition of tensors, and has been recently connected to the
deep learning field by~\citep{levine2018deep}, who constructed a deep
convolutional network in terms of a TN. The use of TNs in machine learning has
appeared in an empirical context, where \citep{NIPS2016_6211} trained a Matrix
Product State (MPS) TN to preform supervised learning tasks on the MNIST
dataset~\citep{lecun1998mnist}. The constructions presented in this section
suggest a separation in expressiveness between recurrent networks
of different depths, as formulated by Conjecture~\ref{conjecture:high_L}.

We begin in Appendix~\ref{app:rac_tns:tns_intro} by providing a brief introduction to TNs.
Next, we present in Appendix~\ref{app:rac_tns:shallow} the TN which corresponds to the calculation of a shallow RAC, and tie it to a common TN architecture referred to as a \emph{Matrix Product State} (MPS) (see overview in e.g. \citep{orus2014practical}), and equivalently to the \emph{tensor train} (TT) decomposition (see Section~\ref{intro:tensors:tt_decomposition}).
Subsequently, we present in Appendix~\ref{app:rac_tns:deep} a TN construction of a deep RAC, and emphasize the characteristics of this construction that are the origin of the enhanced ability of deep RACs to model elaborate temporal dependencies.
Finally, in Appendix~\ref{app:rac_tns:conjecture}, we make use of the above TNs construction in order to formally motivate Conjecture~\ref{conjecture:high_L}, according to which the Start-End separation rank of RACs grows combinatorially with depth.

\section{Introduction to Tensor Networks}\label{app:rac_tns:tns_intro}

\begin{figure}
	\centering
	\includegraphics[width=\linewidth]{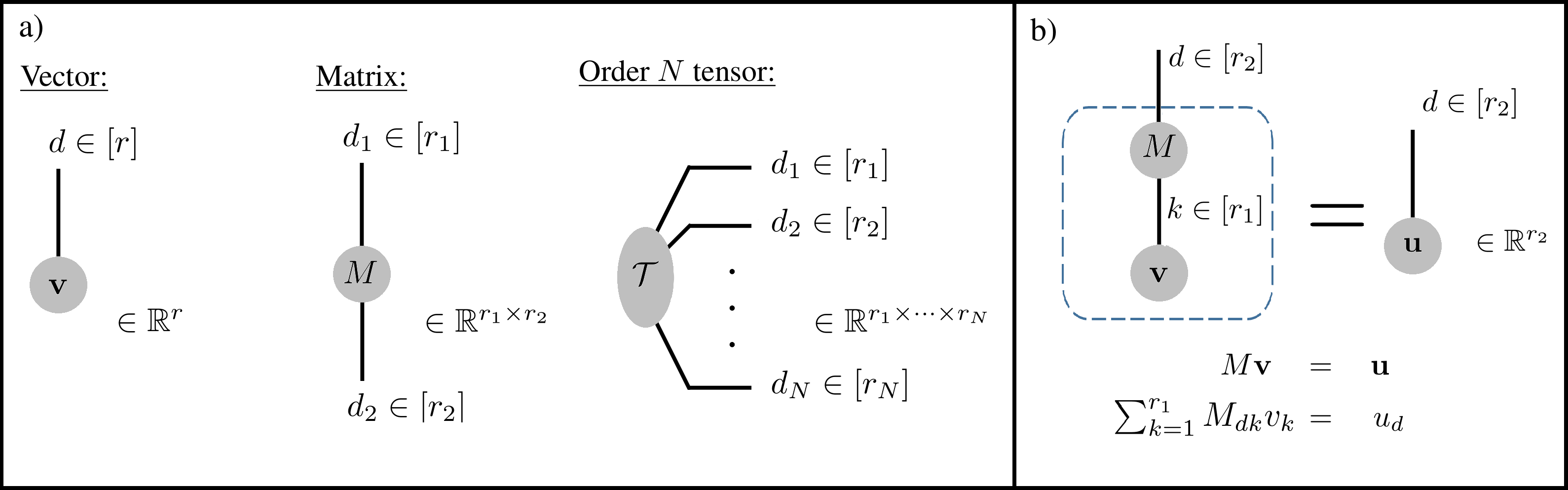}
	\caption{
		A quick introduction to Tensor Networks (TNs). a) Tensors in the TN are
		represented by nodes. The degree of the node corresponds to the order of
		the tensor represented by it. b) A matrix multiplying a vector in TN
		notation. The contracted index $k$, which connects two nodes, is summed
		upon, while the open index $d$ is not. The number of open indices equals
		the order of the tensor represented by the entire network. All of the
		indices receive values that range between $1$ and their bond dimension.
		The contraction is marked by the dashed line.
	}
	\label{fig:tns_intro}
\end{figure}

A TN is a weighted graph, where each node corresponds to a tensor whose order is
equal to the degree of the node in the graph. Accordingly, the edges emanating
out of a node, also referred to as its legs, represent the different modes of
the corresponding tensor. The weight of each edge in the graph, also referred to
as its bond dimension, is equal to the dimension of the appropriate tensor mode.
In accordance with the relation between mode, dimension and index of a tensor
presented in Section~\ref{sec:corr:grid}, each edge in a TN is represented by
an index that runs between $1$ and its bond dimension.
Figure~\ref{fig:tns_intro}a shows three examples:
(1) A vector, which is a tensor of order $1$, is represented by a node with one
leg. (2) A matrix, which is a tensor of order $2$, is represented by a node with
two legs. (3) Accordingly, a tensor of order $N$ is represented in the TN as a
node with $N$ legs.

We move on to present the connectivity properties of a TN. Edges which connect
two nodes in the TN represent an operation between the two corresponding
tensors. A index which represents such an edge is called a contracted index, and
the operation of contracting that index is in fact a summation over all of the
values it can take. An index representing an edge with one loose end is called
an open index. The tensor represented by the entire TN, whose order is equal to
the number of open indices, can be calculated by summing over all of the
contracted indices in the network. An example for a contraction of a simple TN
is depicted in Figure~\ref{fig:tns_intro}b. There, a
TN corresponding to the operation of multiplying a vector $\vv \in \R^{r_1}$ by
a matrix $M\in \R^{r_2 \times r_1}$ is performed by summing over the only
contracted index, $k$. As there is only one open index, $d$, the result of
contracting the network is an order $1$ tensor (a vector): $\uu \in \R^{r_2}$
which upholds $\uu = M\vv$. Though we use below the contraction of indices in
more elaborate TNs, this operation can be essentially viewed as a generalization
of matrix multiplication.

\section{Shallow RAC Tensor Network}
\label{app:rac_tns:shallow}

\begin{figure}
	\centering
	\includegraphics[width=\linewidth]{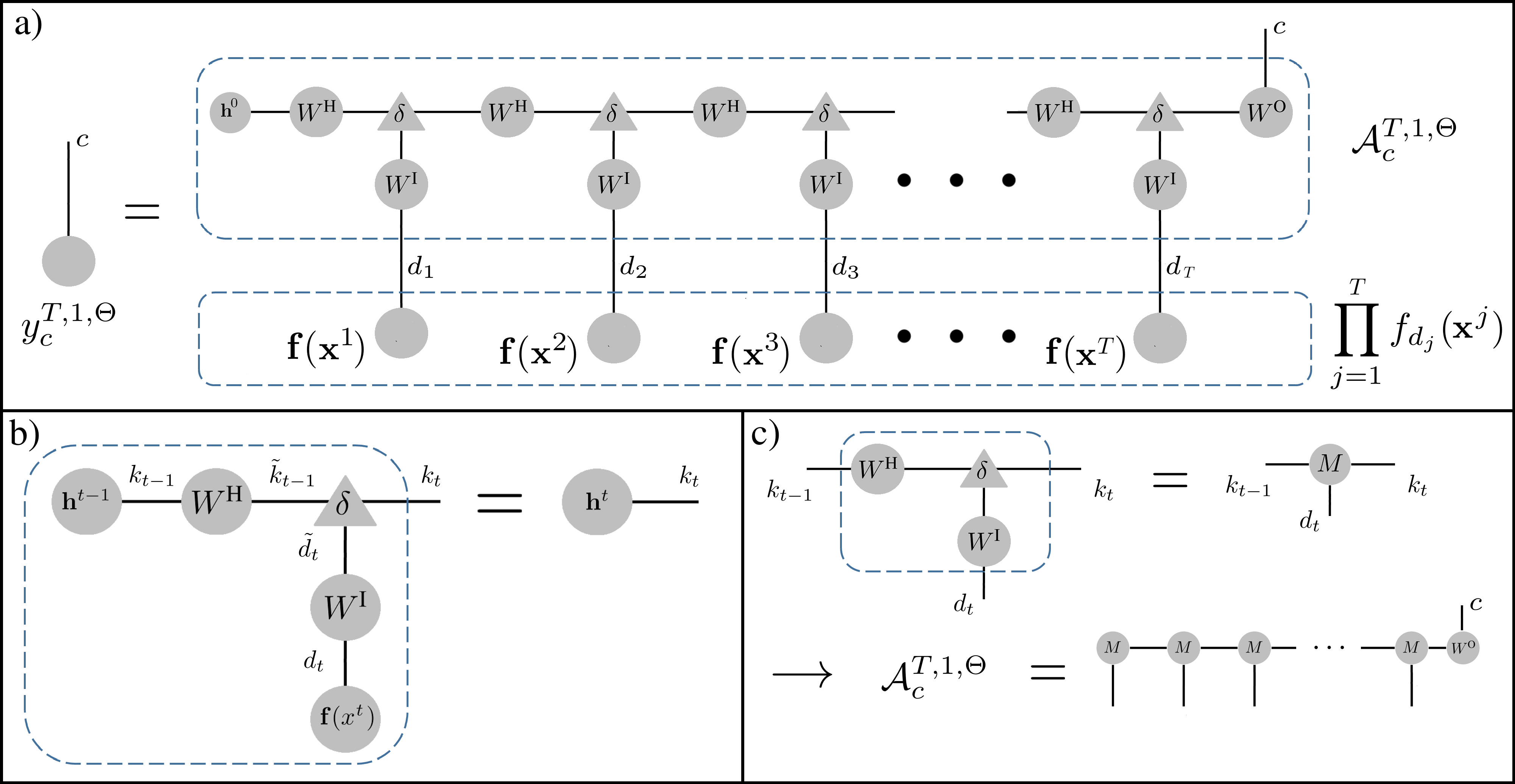}
	
	\caption{
		a) The Tensor Network representing the calculation performed by a shallow RAC.
		b) A Tensor Network construction of the recursive relation given in Equation~\eqref{eq:shallow_rn}.
		c) A presentation of the shallow RAC weights tensor in a standard MPS form.
	}
	\label{fig:shallow_rac_tn}
\end{figure}

The computation of the output at time $T$ that is preformed by the shallow
recurrent network given by Equations~\eqref{eq:shallow_rn} and~\eqref{eq:g_rac}, or
alternatively by Equations~\eqref{eq:score} and~\eqref{eq:tt_decomp}, can be written in
terms of a TN. Figure~\ref{fig:shallow_rac_tn}a
shows this TN, which given some initial hidden state $\h_0$, is essentially a
temporal concatenation of a unit cell that preforms a similar computation at
every time-step, as depicted in
Figure~\ref{fig:shallow_rac_tn}b. For any time
$t<T$, this unit cell is composed of the input weights matrix, $W^\textrm{I}$,
contracted with the inputs vector, $\ff(\x^t)$, and the hidden weights matrix,
$W^\textrm{H}$, contracted with the hidden state vector of the previous time-step, $\h^{t-1}$. The final component in each unit cell is the $3$ legged
triangle representing the order $3$ tensor $\delta\in\R^{R\times R \times R}$,
referred to as the \emph{$\delta$ tensor}, defined by:
\begin{equation}
\begin{array}{c}
\delta_{i_1i_2i_3}\equiv\left\{ \begin{array}{c}
1,\quad ~~~~i_1=i_2=i_3\\
0,\quad ~~~~~~otherwise
\end{array}\right.,\\
\end{array}\label{eq:deltadef}
\end{equation}
with $i_j\in[R]~\forall j\in[3]$, \ie~its entries are equal to $1$ only on the
super-diagonal and are zero otherwise. The use of a triangular node in the TN is
intended to remind the reader of the restriction given in Equation~\eqref{eq:deltadef}.
The recursive relation that is defined by the unit cell, is given by the TN in
Figure~\ref{fig:shallow_rac_tn}b:
\begin{align}
h^t_{k_t}=
\sum_{k_{t-1},\tilde{k}_{t-1},\tilde{d}_{t}=1}^{R}
\sum_{d_{t}=1}^{M}
W^{\textrm{ H}}_{\tilde{k}_{t-1}k_{t-1}}h^{t-1}_{k_{t-1}}
W^{\textrm{ I}}_{\tilde{d}_td_t}f_{d_t}(\x^t)
\delta_{\tilde{k}_{t-1}\tilde{d}_tk_t}=
\nonumber \\
~~~~~~~~~~~
\sum_{\tilde{k}_{t-1}\tilde{d}_t=1}^{R}
(W^{\textrm{ H}}\h^{t-1})_{\tilde{k}_{t-1}}
(W^{\textrm{ I}}\ff(\x^t))_{\tilde{d}_t}
\delta_{\tilde{k}_{t-1}\tilde{d}_tk_t}
=
(W^{\textrm{ H}}\h^{t-1})_{k_t}
(W^{\textrm{ I}}\ff(\x^t))_{k_t},
\label{eq:buildingblock1}
\end{align}
where $k_t\in [R]$. In the first equality, we simply follow the TN prescription
and write a summation over all of the contracted indices in the left hand side
of Figure~\ref{fig:shallow_rac_tn}b, in the
second equality we use the definition of matrix multiplication, and in the last
equality we use the definition of the $\delta$ tensor. The component-wise
equality of Equation~\eqref{eq:buildingblock1} readily implies
$\h^t=(W^{\textrm{H}}\h^{t-1})\odot(W^{\textrm{ I}}\ff(\x^t))$, reproducing the
recursive relation in Equations~\eqref{eq:shallow_rn} and~\eqref{eq:g_rac}, which defines
the operation of the shallow RAC. From the above treatment, it is evident that
the restricted $\delta$ tensor is in fact the component in the TN that yields
the element-wise multiplication property. After $T$ repetitions of the unit cell
calculation with the sequential input $\{\x^t\}_{t=1}^{T }$, a final
multiplication of the hidden state vector $\h^T$ by the output weights matrix
$W^\textrm{O}$ yields the output vector $\y^{T,1,\Theta}$.

The tensor network which represents the order $T$ shallow RAC weights tensor
$\A^{T,1, \Theta}_c$, which appears in Equations~\eqref{eq:score}
and~\eqref{eq:tt_decomp}, is given by the TN in the upper part of
Figure~\ref{fig:shallow_rac_tn}a. In
Figure~\ref{fig:shallow_rac_tn}c, we show that
by a simple contraction of indices, the TN representing the shallow RAC weights
tensor $\A^{T,1, \Theta}_c$ can be drawn in the form of a standard MPS TN. This
TN allows the representation of an order $T$ tensor with a linear (in $T$)
amount of parameters, rather than the regular exponential amount ($\A^{T,1, \Theta}_c$ has $M^T$
entries). The decomposition which corresponds to this MPS TN is known as the Tensor
Train (TT) decomposition of rank $R$ in the tensor analysis community, its
explicit form given in Equation~\eqref{eq:tt_decomp}.

The presentation of the shallow
recurrent network in terms of a TN allows the employment of the min-cut
analysis, which was introduced by \citep{levine2018deep} in the context of
convolutional networks, for quantification of the information flow across time
modeled by the shallow recurrent network. This was indeed preformed in our proof
of the shallow case of Theorem~\ref{theorem:main_result} (see Appendix~\ref{app:proofs:main_result:shallow} for further details).
We now move on to present the
computation preformed by a deep recurrent network in the language of TNs.

\section{Deep RAC Tensor Network}
\label{app:rac_tns:deep}

The construction of a TN which matches the calculation of a deep recurrent
network is far less trivial than that of the shallow case, due to the seemingly
innocent property of reusing information which lies at the heart of the
calculation of deep recurrent networks. Specifically, all of the hidden states
of the network are reused, since the state of each layer at every time-step is
duplicated and sent as an input to the calculation of the same layer in the next
time-step, and also as an input to the next layer up in the same time-step (see
bottom of Figure~\ref{fig:intro_recurrent_net}). The required
operation of duplicating a vector and sending it to be part of two different
calculations, which is simply achieved in any practical setting, is actually
impossible to represent in the framework of TNs. We formulate this notion in the
following claim:
\begin{claim} \label{claim:no_clone}
	Let $v\in\R^P,P\in\N$ be a vector. $v$ is represented by a node with one leg
	in the TN notation. The operation of duplicating this node, \ie~forming two
	separate nodes of degree $1$, each equal to $v$, cannot be achieved by any
	TN.
\end{claim}
\begin{proof}
	We assume by contradiction that there exists a Tensor Network $\phi$ which operates on any vector $v\in\R^P$ and clones it to two separate nodes of degree $1$, each equal to $v$, to form an overall TN representing $v\otimes v$.
	Component wise, this implies that $\phi$ upholds $\forall v\in\R^P:~\sum_{i=1}^P \phi_{ijk}v_i=v_jv_k$.
	By our assumption, $\phi$ duplicates the standard basis elements of $\R^P$, denoted $\{\hat{\e}^{(\alpha)}\}_{\alpha=1}^P$, meaning that $\forall \alpha\in[P]$:
	\begin{equation}\label{eq:no_clone}
	\sum_{i=1}^P \phi_{ijk}\hat{e}^{(\alpha)}_i=\hat{e}^{(\alpha)}_j\hat{e}^{(\alpha)}_k.
	\end{equation}
	By definition of the standard basis elements, the left hand side of Equation~\eqref{eq:no_clone} takes the form $\phi_{\alpha jk}$ while the right hand side equals $1$ only if $j=k=\alpha$, and otherwise $0$. Utilizing the $\delta$-tensor notation presented in Equation~\eqref{eq:deltadef}, in order to successfully clone the standard basis elements, Equation~\eqref{eq:no_clone} implies that $\phi$ must uphold $\phi_{\alpha jk}=\delta_{\alpha jk}$. However, for $v=\mathbf{1}$, \ie~$\forall j\in[P]:~v_j=1$, a cloning operation does not take place when using this value of $\phi$, since $\sum_{i=1}^P \phi_{ijk}v_i=\sum_{i=1}^P \delta_{ijk}=\delta_{jk}\neq 1= v_iv_j$, in contradiction to $\phi$ duplicating any vector in $\R^P$.
\end{proof}

\begin{figure}
	\centering
	\includegraphics[width=\linewidth]{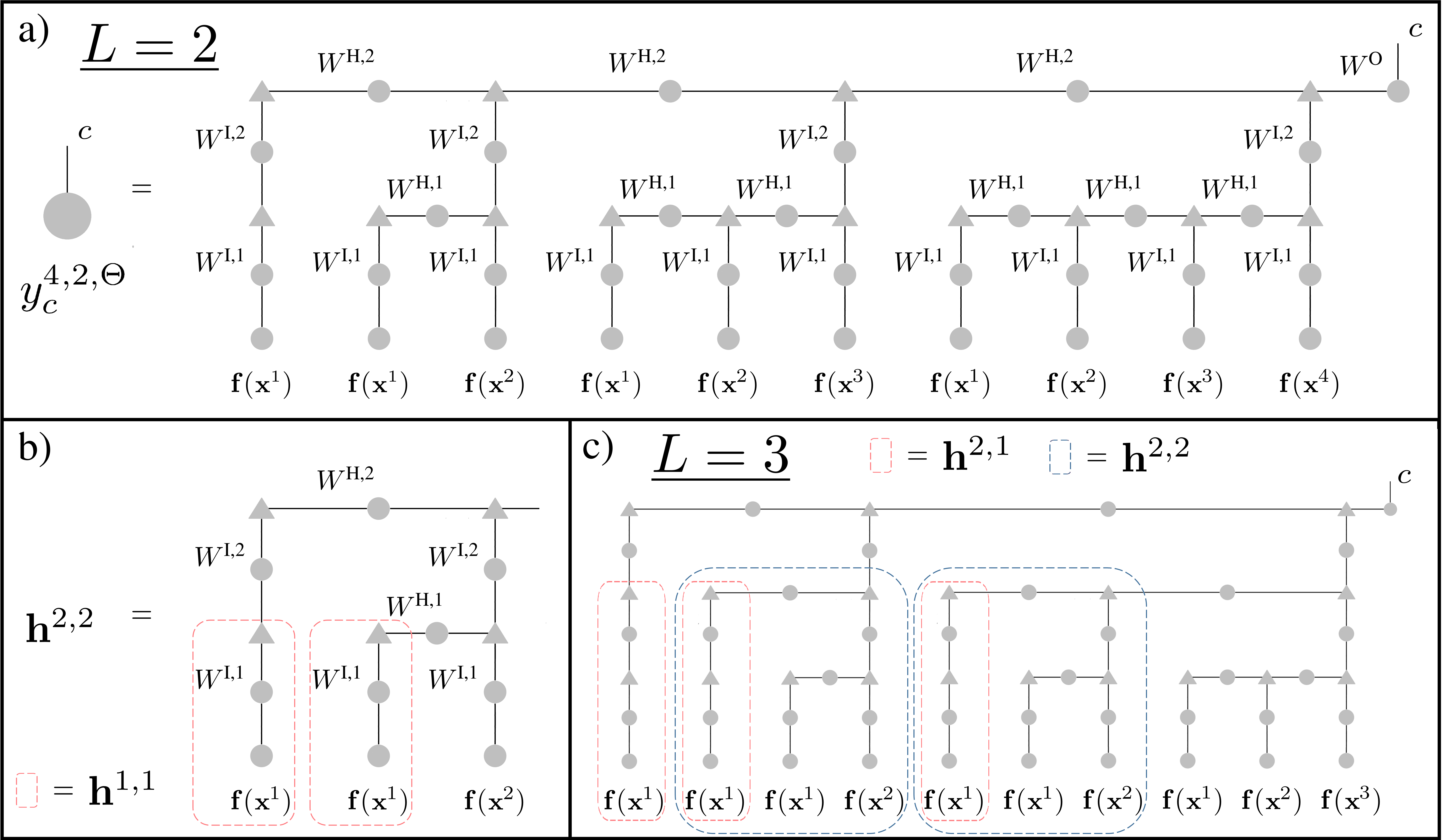}
	\caption{
		a) The Tensor Network representing the calculation preformed by a depth $L=2$ RAC after 4 time-steps.
		b) A Tensor Network construction of the hidden state $\h^{2,2}$ [see Equation~\eqref{eq:dup_example}], which involves duplication of the hidden state $ \h^{1,1}$ that is achieved by duplicating the input $x^1$.
		c) The Tensor Network representing the calculation preformed by a depth $L=3$ RAC after 3 time-steps. Here too, as in any deep RAC, several duplications take place.
	}
	\label{fig:deep_rac_tn}
\end{figure}

Claim~\ref{claim:no_clone} seems to pose a hurdle in our pursuit of a TN
representing a deep recurrent network. Nonetheless, a form of such a TN may be
attained by a simple `trick'~--~in order to model the duplication that is
inherently present in the deep recurrent network computation, we resort to
duplicating the input data itself. By this technique, for every duplication that
takes place along the calculation, the input is inserted into the TN multiple
times, once for each sequence that leads to the duplication point. This
principle, which allows us to circumvent the restriction imposed by
Claim~\ref{claim:no_clone}, yields the elaborate TN construction of deep RACs
depicted in Figure~\ref{fig:deep_rac_tn}.

It is important to note that these TNs, which grow exponentially in size as the
depth $L$ of the recurrent network represented by them increases, are merely a
theoretical tool for analysis and not a suggested implementation scheme for deep
recurrent networks. The actual deep recurrent network is constructed according to the
simple scheme given at the bottom of
Figure~\ref{fig:intro_recurrent_net}, which grows only
linearly in size as the depth $L$ increases, despite the corresponding TN
growing exponentially. In fact, this exponential `blow-up' in the size of the
TNs representing the deep recurrent networks is closely related to their ability
to model more intricate dependencies over longer periods of time in comparison
with their shallower counterparts, which was established in
Section~\ref{sec:theory:bounds}.

Figure~\ref{fig:deep_rac_tn} shows TNs which correspond to depth $L=2,3$ RACs.
Even though the TNs in Figure~\ref{fig:deep_rac_tn} seem rather convoluted and
complex, their architecture follows clear recursive rules. In
Figure~\ref{fig:deep_rac_tn}a, a depth $L=2$
recurrent network is presented, spread out in time onto $T=4$ time-steps.
To understand the logic underlying the input duplication process, which in turn
entails duplication of entire segments of the TN, we focus on the calculation of
the hidden state vector $\h^{2,2}$ that is presented in
Figure~\ref{fig:deep_rac_tn}b. When the first
inputs vector, $\ff(\x^1)$, is inserted into the network, it is multiplied by
$W^{{\textrm I},1}$ and the outcome is equal to $\h^{1,1}$. Note that in this
figure, the initial condition for each layer $l\in L$, $\h^{l,0}$, is chosen
such that a vector of ones will be present in the initial element-wise
multiplication: $(\h^{0,l})^T={\mathbf{1}}^T(W^{{\textrm H},l})^{\dagger}$,where $\dagger$ denotes the pseudoinverse operation.

Next, $\h^{1,1}$ is used in two different places, as an inputs vector to layer $L=2$
at time $t=1$, and as a hidden state vector in layer $L=1$ for time $t=2$
calculation. Our input duplication technique inserts $\ff(\x^1)$ into the network
twice, so that the same exact $\h^{1,1}$ is achieved twice in the TN, as marked
by the red dotted line in
Figure~\ref{fig:deep_rac_tn}b. This way, every copy
of $\h^{1,1}$ goes to the appropriate segment of the calculation, and indeed the
TN in Figure~\ref{fig:deep_rac_tn}b holds the
correct value of $\h^{2,2}$:
\begin{equation} \label{eq:dup_example}
\h^{2,2} = \left(
\vphantom{(W^{{\textrm H},1}\h^{1,1})\odot(W^{{\textrm I},1}\ff(\x^2))}
W^{{\textrm H},2}W^{{\textrm I},2}\h^{1,1}\right)\odot \left(W^{{\textrm I},2}((W^{{\textrm H},1}\h^{1,1})\odot(W^{{\textrm I},1}\ff(\x^2)))\right).
\end{equation}

The extension to deeper layers leads us to a fractal structure of the TNs,
involving many self similarities, as in the $L=3$ example given in
Figure~\ref{fig:deep_rac_tn}c. The duplication of
intermediate hidden states, marked in red and blue in this example, is the
source of the apparent complexity of this $L=3$ RAC TN.
Generalizing the above $L=1,2,3$ examples, a
TN representing an RAC of general depth $L$ and of $T$ time-steps, would involve
in its structure $T$ duplications of TNs representing RACs of depth $L-1$, each
of which has a distinct length in time-steps $i$, where $i\in [T]$. This fractal structure leads to an increasing with
depth complexity of the TN representing the depth $L$ RAC computation, which
we show in the next section to motivate the combinatorial lower bound on the Start-End separation rank of deep RACs, given in Conjecture~\ref{conjecture:high_L}.

\section{A Formal Motivation for Conjecture~\ref{conjecture:high_L}}
\label{app:rac_tns:conjecture}

The above presented construction of TNs which correspond to deep RACs, allows us to further investigate the effect of network depth on its ability to model long-term temporal dependencies. We present below a formal motivation for the lower bound on the Start-End separation rank of deep recurrent networks, given in Conjecture~\ref{conjecture:high_L}. Though our analysis employs TNs visualizations, it is formal nonetheless -- these graphs represent the computation in a well-defined manner (see Appendices~\ref{app:rac_tns:tns_intro}-\ref{app:rac_tns:deep} above).

Our conjecture relies on the fact that it is sufficient
to find a specific instance of the network parameters $\Theta\times\h^{0,l}$ for which
$\mat{\A(y^{T,L, \Theta}_c)}_{S,E}$ achieves a certain rank, in order for this rank to
be a lower bound on the Start-End separation rank of the network. This follows from combining Claim~\ref{claim:grid_sep_deep} and Lemma~\ref{lemma:poly_full_rank}. Claim~\ref{claim:grid_sep_deep} assures us that the Start-End separation rank of the function realized by an RAC of any depth $L$, is lower bounded by the rank of the matrix obtained by the corresponding grid tensor matricization: $\sep{S,E}{y^{T,L, \Theta}_c}\geq \rank{\mat{\A(y^{T,L, \Theta}_c)}_{S,E}}$. Thus, one must show that $\rank{\mat{\A(y^{T,L, \Theta}_c)}_{S,E}} \geq${\tiny{$\multiset{\min\{M,R\}}{\multiset{T/2}{L-1}}$}} for all of the values of parameters $\Theta\times\h^{0,l}$ but a set of Lebesgue measure zero, in order to establish the lower
bound in Conjecture~\ref{conjecture:high_L}.
Next, we rely on Lemma~\ref{lemma:poly_full_rank}, which states that since the entries of
$\A(y^{T,L, \Theta}_c)$ are polynomials in the deep recurrent network's
weights, it suffices to find a single example for which the rank of the
matricized grid tensor is greater than the desired lower bound. Finding
such an example would indeed imply that for almost all of the values of
the network parameters, the desired inequality holds.

\begin{figure}
	\centering
	\includegraphics[width=1\linewidth]{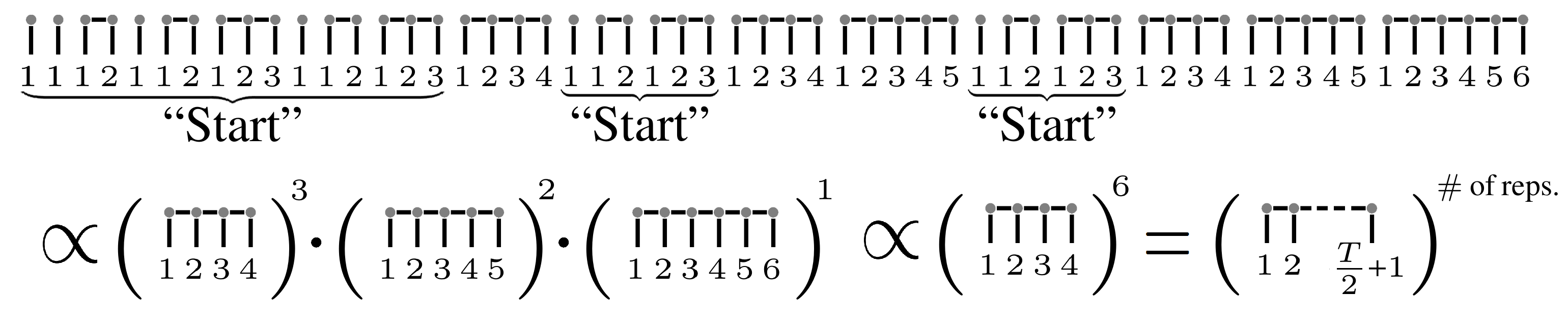}
	\caption{Above: TN representing the computation of a depth $L=3$ RAC after $T=6$ time-steps, when choosing $\WI{2}$ to be of rank-1. See full TN, for general values of the weight matrices, in Figure~\ref{fig:conjecture}. Below: Reduction of this TN to the factors affecting the Start-End matricization of the grid tensor represented by the TN.}
	\label{fig:conjecture_app}
\end{figure}

In the following, we choose a weight assignment that effectively `separates' between the first layer and higher layers, in the sense that $\WI{2}$ is of rank-1. This is done in similar spirit to the assignment used in the proof of Theorem~\ref{theorem:main_result}, in which $\WI{2}_{ij} \equiv \delta_{i1}$ (see Appendix~\ref{app:proofs:main_result}). Under this simplifying assignment, which suffices for our purposes according to the above discussion, the entire computation performed in deeper layers contributes only a constant factor to the matricized grid tensor. In this case, the example of the TN corresponding to an RAC of depth $L=3$ after $T=6$ time-steps, which is shown in full in Figure~\ref{fig:conjecture}, takes the form shown in the upper half of Figure~\ref{fig:conjecture_app}.
Next, in order to evaluate $\rank{\mat{\A(y^{T,L, \Theta}_c)}_{S,E}}$, we note that graph segments which involve only indices from the ``Start'' set, will not affect the rank of the matrix under mild conditions on $\WI{1}, \WH{1}$ (for example, this holds if $\WI{1}$ is fully ranked and does not have vanishing elements, and $\WH{1}=I$). Specifically, under the Start-End matricization these segments will amount to a different constant multiplying each row of the matrix. For the example of the RAC of depth $L=3$ after $T=6$ time-steps, this amounts to the effective TN given in the bottom left side of Figure~\ref{fig:conjecture_app}.
Finally, the dependence of this TN on the indices of time-steps $\{\nicefrac{T}{2}+2,\ldots ,T\}$, namely those outside of the basic unit involving indices of time-steps $\{1,\ldots ,\nicefrac{T}{2}+1\}$, may only increase the resulting Start-End matricization rank (this holds due to the temporal invariance of the recurrent network's weights). Thus, we are left with an effective TN resembling the one shown in Section~\ref{sec:results:conjecture}, where the basic unit separating ``Start" and ``End" indices is raised to the power of the number of its repetitions in the graph.
In the following, we prove a claim according to which the number of repetitions of this basic unit in the TN graph increases combinatorially with the depth of the RAC:

\begin{claim}
	Let $\phi(T,L,R)$ be the TN representing the computation performed after
	$T$ time-steps by an RAC with $L$ layers and $R$ hidden channels per layer. Then, the number of occurrences in layer $L=1$ of the basic unit connecting ``Start" and ``End" indices (bottom right in Figure~\ref{fig:conjecture_app}), is exactly {\tiny{\multiset{T/2}{L-1}}}.
\end{claim}
\begin{proof}
	Let $y^{T,L, \Theta}_c$ be the function computing the output after
	$T$ time-steps of an RAC with $L$ layers, $R$ hidden channels per layer and
	weights denoted by $\Theta$. In order to focus on repetitions in layer $L=1$, we assign $\WI{2}_{ij} \equiv \delta_{i1}$ for which the following upholds (see a similar and more detailed derivation in Appendix~\ref{app:proofs:main_result}):
	\begin{align*}
	\A(y^{T,L, \Theta}_c)_{d_1,\ldots,d_T}
	= \left(Const.\right)&\prod_{t_L=1}^T\prod_{t_{L-1}=1}^{t_L}\cdots\prod_{t_2=1}^{t_3} \sum_{r_1,\ldots,r_{t_2}=1}^R\left( \prod_{j=1}^{t_2}
	\WI{1}_{r_jd_j}\prod_{j=1}^{t_2-1}\WH{1}_{r_jr_{j+1}}\right)\\
	= \left(Const.\right)(V_{d_1\ldots d_{\nicefrac{T}{2}}})\prod_{t_L=\nicefrac{T}{2}+1}^T&\prod_{t_{L-1}=\nicefrac{T}{2}+1}^{t_L}\cdots\prod_{t_2=\nicefrac{T}{2}+1}^{t_3} \sum_{r_1,\ldots,r_{t_2}=1}^R\left( \prod_{j=1}^{t_2}
	\WI{1}_{r_jd_j}\prod_{j=1}^{t_2-1}\WH{1}_{r_jr_{j+1}}\right),
	\end{align*}
	where the constant term in the first line is the contribution of the deeper layers under this assignment, and the tensor $V_{d_1\ldots d_{\nicefrac{T}{2}}}$, which becomes a vector under the Start-End matricization, reflects the contribution of the ``Start'' set indices. Observing the argument of the chain of products in the above expression, $\sum_{r_1,\ldots,r_{t_2}=1}^R\left( \prod_{j=1}^{t_2}
	\WI{1}_{r_jd_j}\prod_{j=1}^{t_2-1}\WH{1}_{r_jr_{j+1}}\right)$, it is an order $t_2$ tensor, exactly given by the TN representing the computation of a depth $L=1$ RAC after $t_2$ time-steps.
	Specifically, for $t_2=\nicefrac{T}{2}+1$, it is exactly equal to the basic TN unit connecting ``Start" and ``End" indices, and for $\nicefrac{T}{2}+1<t_2\leq T$ it contains this basic unit. This means that in order to obtain the number of repetition of this basic unit in $\phi$, we must count the number of multiplications implemented by the chain of products in the above expression. Indeed this number is equal to:
	\begin{equation*}
	\sum_{t_L=\nicefrac{T}{2} +1}^T\sum_{t_{L-1}=\nicefrac{T}{2} +1}^{t_L}\cdots \sum_{t_2=\nicefrac{T}{2} +1}^{t_3}t_2 =\multiset{T/2}{L-1}
	\end{equation*}
	
\end{proof}

Finally, the form of the lower bound presented in Conjecture~\ref{conjecture:high_L} is obtained by considering a rank $R$ matrix, such as the one obtained by the Start-End matricization of the TN basic unit discussed above, raised to the Hadamard power of $\multiset{\nicefrac{T}{2}}{L-1}$. The rank of the resultant matrix, is upper bounded by {\tiny{$\multiset{R}{\multiset{\nicefrac{T}{2}}{L-1}}$}} as shown for example in~\citep{amini2012low}. We leave it as an open problem to prove Conjecture~\ref{conjecture:high_L}, by proving that the upper bound is indeed tight in this case.

\clearpage
%
%

\chapter{Deferred proof of Theorem~\ref{theorem:main_result}}
\label{app:proofs:main_result}


	

\initial{I}n this appendix, we follow the proof strategy that is outlined in
Section~\ref{sec:theory:bounds}, and prove Theorem~\ref{theorem:main_result}, which
shows a combinatorial advantage of deep recurrent networks over shallow ones in
the ability to model long-term dependencies, as measured by the Start-End
separation rank (see Section~\ref{sec:corr:sep}).
In Appendices~\ref{app:proofs:main_result:shallow} and~\ref{app:proofs:main_result:deep}, we prove the bounds on the
Start-End separation rank of the shallow and deep RACs, respectively,
while more technical lemmas which are employed during the proof are relegated to
Appendix~\ref{app:proofs:main_result:technical}.

\section{The Start-End Separation Rank of Shallow RACs}
\label{app:proofs:main_result:shallow}

We consider the Tensor Network construction of the calculation carried
out by a shallow RAC, given in Figure~\ref{fig:shallow_rac_tn}.
According to the presented construction, the shallow RAC weights
tensor [Equations~\eqref{eq:score} and~\eqref{eq:tt_decomp}] is represented by
a Matrix Product State (MPS) Tensor Network
\citep{orus2014practical}, with the following order-3 tensor
building block:
$M_{k_{t-1}d_tk_t}=W^\textrm{I}_{k_td_t}W^\textrm{H}_{k_tk_{t-1}}$,
where $d_t\in[M]$ is the input index and $k_{t-1},k_t\in[R]$ are the
internal indices (see
Figure~\ref{fig:shallow_rac_tn}c). In
TN terms, this means that the bond dimension of this MPS is equal to
$R$. We apply the result of \citep{levine2018deep}, who state that the
rank of the matrix obtained by matricizing any tensor according to a
partition $(S,E)$ is equal to a min-cut separating $S$ from $E$ in the
Tensor Network graph representing this tensor, for all of the values of the TN
parameters but a set of Lebesgue measure zero. In this MPS Tensor
Network, the minimal cut \wrt~the partition $(S,E)$ is equal to the
bond dimension $R$, unless $R > M^{\nicefrac{T}{2}}$, in which
case the minimal cut contains the external legs instead. Thus, in the
TN representing $\A^{T,1, \Theta}_c$, the minimal cut \wrt~the partition
$(S,E)$ is equal to $\min\{R,M^{\nicefrac{T}{2}}\}$, implying
$\rank{\mat{\A^{T,1, \Theta}_c})_{S,E}} = \min\{R,M^{\nicefrac{T}{2}}\}$ for all values of the parameters but a set of Lebesgue measure zero.
The first half of the
theorem follows from applying Claim~\ref{claim:grid_sep_shallow},
which assures us that the Start-End separation rank of the function
realized by a shallow ($L=1$) RAC is equal to
$\rank{\mat{\A^{T,1, \Theta}_c})_{S,E}}$.

\hfill $\square$ 

\section{Lower-bound on the Start-End Separation Rank of Deep RACs}
\label{app:proofs:main_result:deep}

For a deep network, Claim~\ref{claim:grid_sep_deep} assures us that
the Start-End separation rank of the function realized by a depth
$L =2$ RAC is lower bounded by the rank of the matrix obtained by the
corresponding grid tensor matricization, for any choice of template
vectors. Specifically:
\begin{equation*}
\sep{S,E}{y^{T,L, \Theta}_c}
\geq \rank{\mat{\A(y^{T,L, \Theta}_c)}_{S,E}}.
\end{equation*}
Thus, proving that
$\rank{\mat{\A(y^{T,L, \Theta}_c)}_{S,E}} \geq \multiset{\min\{R,M\}}{T/2}$
for all of the values of parameters $\Theta\times\h^{0,l}$ but a set of Lebesgue measure
zero, would satisfy the theorem.

In the following, we provide an assignment of weight matrices and initial hidden states for
which $\rank{\mat{\A(y^{T,L, \Theta}_c)}_{S,E}} = \multiset{\min\{R,M\}}{T/2}$.
In accordance with Claim~\ref{claim:rank_everywhere}, this will
suffice as such an assignment implies this rank is achieved for all configurations of
the recurrent network weights but a set of Lebesgue measure zero.

We begin by choosing a specific set of template vectors
$\x^{(1)}, \ldots, \x^{(M)} \in \X$. Let $F \in \R^{M \times M}$ be a
matrix with entries defined by $F_{ij} \equiv f_j(\x^{(i)})$.
According to \citep{cohen2016convolutional}, since $\{f_d\}_{d=1}^M$
are linearly independent, then there is a choice of template vectors
for which $F$ is non-singular.

Next, we describe our assignment. In the
expressions below we use the notation $\delta_{ij} = \begin{cases}
1 & i = j \\ 0 & i \neq j\end{cases}$. Let $z \in \R \setminus \{0\}$
be an arbitrary non-zero real number, let $\Omega \in \R_+$ be an arbitrary
positive real number, and let $Z \in\R^{R \times M}$
be a matrix with entries
$Z_{ij} \equiv \begin{cases} z^{\Omega^i \delta_{ij}} &
i \leq M \\ 0 & i > M \end{cases}$.

We set
$\WI{1} \equiv Z \cdot (F^T)^{-1}$ and set $\WI{2}$ such that its entries
are $\WI{2}_{ij} \equiv \delta_{i1}$. 
We set
$\WH{1} \equiv \WH{2} \equiv I$, i.e. to the identity matrix, and
additionally we set the entries of $\WO$ to $\WO_{ij} = \delta_{1j}$. Finally, we choose the
initial hidden state values so they bear no effect on the calculation, namely $\h^{0,l}=\left(\WH{l}\right)^{-1}\1=\1$ for $l=1,2$.

Under the above assignment, the output for the corresponding class $c$
after $T$ time-steps is equal to:
\begin{align*}
y^{T,L, \Theta}_c(\x^1, \ldots, \x^T)
&= \left(\WO \h^{T,2}\right)_c \\
(\WO_{ij} \equiv \delta_{1j}) \Rightarrow &= (\h^{T,2})_1 \\
[\text{Equation~\eqref{eq:deep_rn}}] \Rightarrow
&= \left((\WH{2} \h^{T-1,2}) \odot (\WI{2}\h^{T,1})\right)_1 \\
(\WH{2} \equiv I) \Rightarrow
&= \left((\h^{T-1,2}) \odot (\WI{2}\h^{T,1})\right)_1\\
(\h^{0,2}=\1) \Rightarrow
&= \prod_{t=1}^T \left(\WI{2}\h^{t,1}\right)_1 \\
(\WI{2}_{ij} \equiv \delta_{1i}) \Rightarrow
&= \prod_{t=1}^T \sum_{r=1}^R \left(\h^{t,1}\right)_r \\
[\text{Equation~\eqref{eq:deep_rn}}] \Rightarrow
&= \prod_{t=1}^T \sum_{r=1}^R
\left((\WH{1} \h^{t-1,1}) \odot (\WI{1} \ff(\x^t))\right)_r \\
(\WH{1} \equiv I) \Rightarrow
&= \prod_{t=1}^T \sum_{r=1}^R
\left((\h^{t-1,1}) \odot (\WI{1} \ff(\x^t))\right)_r\\
(\h^{0,1}=\1) \Rightarrow
&= \prod_{t=1}^T \sum_{r=1}^R \prod_{j=1}^t
\left(\WI{1} \ff(\x^j)\right)_r.
\end{align*}
When evaluating the grid tensor for our chosen set of template
vectors, i.e. $\A(y^{T,L, \Theta}_c)_{d_1,\ldots,d_T} =
y^{T,L, \Theta}_c(\x^{(d_1)}, \ldots, \x^{(d_T)})$, we can
substitute $f_j(\x^{(i)}) \equiv F_{ij}$, and thus
\begin{equation*}
(\WI{1} \ff(\x^{(d)}))_r = (\WI{1} F^T)_{rd} = (Z \cdot (F^T)^{-1}F^T)_{rd} = Z_{rd}.
\end{equation*}
Since we defined $Z$ such that for $r \geq \min\{R,M\}$ $Z_{rd} = 0$, and denoting $\bar{R}\equiv \min\{R,M\} $ for brevity of notation, the
grid tensor takes the following form:
\begin{equation*}
\A(y^{T,L, \Theta}_c)_{d_1,\ldots,d_T} =
\prod_{t=1}^T \sum_{r=1}^{\bar{R}} \prod_{j=1}^t Z_{r d_j}
= \left(\prod_{t=1}^{\nicefrac{T}{2}} \sum_{r=1}^{\bar{R}} \prod_{j=1}^t Z_{r d_j}\right)
\cdot \left(\prod_{t=\nicefrac{T}{2}+1}^{T} \sum_{r=1}^{\bar{R}} \prod_{j=1}^t Z_{r d_j}\right),
\end{equation*}
where we split the product into two expressions, the left
part that contains only the indices in the start set $S$, i.e.
$d_1,\ldots,d_{\nicefrac{T}{2}}$, and the right part which contains
all external indices (in the start set $S$ and the end set $E$). Thus, under matricization w.r.t. the Start-End
partition, the left part is mapped to a vector
$\aaa \equiv \matflex{\prod_{t=1}^{\nicefrac{T}{2}} \sum_{r=1}^{\bar{R}}
	\prod_{j=1}^t Z_{r d_j}}_{S,E}$ containing only non-zero entries per the definition of $Z$, and
the right part is mapped to a matrix $B \equiv
\matflex{\prod_{t=\nicefrac{T}{2}+1}^T
	\sum_{r=1}^{\bar{R}} \prod_{j=1}^t Z_{r d_j}}_{S,E}$, where each entry
of $\uu$ multiplies the corresponding row of $B$. This results in:
\begin{equation*}
\mat{\A(y^{T,L, \Theta}_c)_{d_1,\ldots,d_T}}_{S,E}
= \mathrm{diag}(\aaa) \cdot B.
\end{equation*}
Since $\aaa$ contains only non-zero entries, $\mathrm{diag}(\aaa)$ is
of full rank, and so
$\rank{\mat{\A(y^{T,L, \Theta}_c)_{d_1,\ldots,d_T}}_{S,E}}=\rank{B}$, leaving us to prove that $\rank{B} =
\multiset{{\bar{R}}}{\nicefrac{T}{2}}$ . For brevity of notation, we define $N \equiv \multiset{{\bar{R}}}{\nicefrac{T}{2}}$.

To prove the above, it is sufficient to show that $B$
can be written as a sum of $N$
rank-1 matrices, i.e. $B=\sum_{i=1}^N \uu^{(i)} \otimes \vv^{(i)}$,
and that $\{\uu^{(i)}\}_{i=1}^N$ and $\{\vv^{(i)}\}_{i=1}^N$ are two
sets of linearly independent vectors. Indeed, applying
Claim~\ref{claim:decomp} on the entries of $B$, specified w.r.t. the
row $(d_1,\ldots,d_{\nicefrac{T}{2}})$ and column
$(d_{\nicefrac{T}{2}+1},\ldots,d_T)$, yields the following form:
\begin{equation*}
B_{(S,E)}=
\sum_{\substack{\p^{(\nicefrac{T}{2})} \in \state{{\bar{R}}}{\nicefrac{T}{2}}}}
\left(
\prod_{r=1}^{{\bar{R}}} \prod_{j=1}^{\nicefrac{T}{2}} Z_{r d_j}^{p^{(\nicefrac{T}{2})}_r}
\vphantom{\sum_{\substack{(\p^{(\nicefrac{T}{2}-1)},\ldots,\p^{(1)}) \\ \in \trajectory{\p^{(\nicefrac{T}{2})}}}}}
\right) \cdot
\left(
\sum_{\substack{(\p^{(\nicefrac{T}{2}-1)},\ldots,\p^{(1)}) \\ \in \trajectory{\p^{(\nicefrac{T}{2})}}}}
\prod_{r=1}^{{\bar{R}}}{\prod_{j=\nicefrac{T}{2}+1}^T} Z_{r d_j}^{p_r^{(T-j+1)}}
\right),
\end{equation*}
where for all $k$, $\p^{(k)}$ is $\bar{R}$-dimensional vector of non-negative integer
numbers which sum to $k$, and we explicitly define
$\state{\bar{R}}{\nicefrac{T}{2}}$ and $\trajectory{\p^{(\nicefrac{T}{2})}}$
in Claim~\ref{claim:decomp}, providing a softer more intuitive definition hereinafter. $\state{\bar{R}}{\nicefrac{T}{2}}$
can be viewed as the set of all possible states of a bucket containing $\nicefrac{T}{2}$
balls of ${\bar{R}}$ colors, where $p^{(\nicefrac{T}{2})}_r$ for $r\in[\bar{R}]$ specifies the number of balls
of the $r$'th color.
$\trajectory{\p^{(\nicefrac{T}{2})}}$ can be viewed
as all possible trajectories from a given state to an empty bucket, i.e. $(0,\ldots,0)$,
where at each step we remove a single ball from the bucket.
We note that the number of all initial states of the bucket is exactly
$\abs{\state{\bar{R}}{\nicefrac{T}{2}}} = N \equiv \multiset{{\bar{R}}}{\nicefrac{T}{2}}$.
Moreover, since the expression in the left parentheses
contains solely indices from the start set $S$, i.e. $d_1,\ldots,d_{\nicefrac{T}{2}}$,
while the right contains solely indices from the end set $E$, i.e.
$d_{\nicefrac{T}{2}+1},\ldots,d_T$, then each summand is in fact a rank-1 matrix. Specifically,
it can be written as $\uu^{\p^{(\nicefrac{T}{2})}} \otimes \vv^{\p^{(\nicefrac{T}{2})}}$,
where the entries of $\uu^{\p^{(\nicefrac{T}{2})}}$ are represented by the expression
in the left parentheses, and those of $\vv^{\p^{(\nicefrac{T}{2})}}$ by the expression
in the right parentheses.

We prove that the set $\left\{ \uu^{\p^{(\nicefrac{T}{2})}} \in \R^{M^{\nicefrac{T}{2}}}
\right\}_{\p^{(\nicefrac{T}{2})} \in \state{\bar{R}}{\nicefrac{T}{2}}}$ is linearly
independent by arranging it as the columns of the matrix
$U \in \R^{M^{\nicefrac{T}{2}} \times N}$, and showing that its rank equals to $N$.
Specifically, we observe the sub-matrix defined by the subset of the rows of $U$,
such that we select the row $\dd \equiv (d_1,\ldots,d_{\nicefrac{T}{2}})$ only if it holds
that $\forall j, d_j \leq d_{j+1}$. 
Note that there are exactly $N$ such rows, similarly to the number of columns, which can be intuitively understood since for the imaginary `bucket states' defining the columns $\p^{(\nicefrac{T}{2})}$ there is no meaning of order in the balls, and having imposed the restriction $\forall j, d_j \leq d_{j+1}$ on the $\nicefrac{T}{2}$ length tuple $\dd$, there is no longer a degree of freedom to order the `colors' in $\dd$, reducing the number of rows from $M^{\nicefrac{T}{2}}$ to $N$ (note that by definition $N\leq\multiset{M}{\nicefrac{T}{2}}<M^{\nicefrac{T}{2}}$ ). 
Thus, in the resulting
sub-matrix, denoted by $\bar{U} \in \R^{N \times N}$, not only do the columns correspond
to the vectors of $\state{\bar{R}}{\nicefrac{T}{2}}$, but also its rows, where
the row specified by the tuple $\dd$, corresponds to the
vector $\q^{(\nicefrac{T}{2})} \in \state{\bar{R}}{\nicefrac{T}{2}}$, such that for $r\in[\bar{R}]:~q^{(\nicefrac{T}{2})}_r \equiv \abs{\{j \in [\nicefrac{T}{2}] | d_j = r\}}$ specifies the amount of
repetitions of the number (`color') $r$ in the given tuple.

Accordingly, for each element of
$\bar{U}$ the following holds:
\begin{align*}
\bar{U}_{\q^{(\nicefrac{T}{2})},\p^{(\nicefrac{T}{2})}}
&= \prod_{r=1}^{{\bar{R}}} \prod_{j=1}^{\nicefrac{T}{2}} Z_{r d_j}^{p^{(\nicefrac{T}{2})}_r}\\
(Z_{ij} = z^{\Omega^i \delta_{ij}}) \Rightarrow
&= z^{\sum_{j=1}^{\nicefrac{T}{2}} \sum_{r=1}^{\bar{R}} p^{(\nicefrac{T}{2})}_r\Omega^r\delta_{r d_j}} \\
(\text{definition of }\delta_{ij}) \Rightarrow
&= z^{\sum_{j=1}^{\nicefrac{T}{2}} \Omega^{d_j} p^{(\nicefrac{T}{2})}_{d_j}} \\
(\text{Grouping identical summands}) \Rightarrow
&= z^{\sum_{r=1}^{\bar{R}} \Omega^r \abs{\{j \in [\nicefrac{T}{2}] | d_j = r\}} p^{(\nicefrac{T}{2})}_r} \\
(q^{(\nicefrac{T}{2})}_r \equiv \abs{\{j \in [\nicefrac{T}{2}] | d_j = r\}}) \Rightarrow
&= z^{\sum_{r=1}^{\bar{R}} \Omega^r q^{(\nicefrac{T}{2})}_r p^{(\nicefrac{T}{2})}_r} \\
\left(
\begin{matrix}
\bar{q}^{(\nicefrac{T}{2})}_r {\equiv} \Omega^{\nicefrac{r}{2}} q^{(\nicefrac{T}{2})}_r \\
\bar{p}^{(\nicefrac{T}{2})}_r {\equiv} \Omega^{\nicefrac{r}{2}} p^{(\nicefrac{T}{2})}_r
\end{matrix}
\right) \Rightarrow
&= z^{\inprod{\bar{\q}^{(\nicefrac{T}{2})}}{\bar{\p}^{(\nicefrac{T}{2})}}}.
\end{align*}
Since the elements of $\bar{U}$ are polynomial in $z$, then as we prove in
Lemma~\ref{lemma:poly_full_rank}, it is sufficient to show that there exists
a single contributor to the determinant of $\bar{U}$ that has the highest degree
of $z$ in order to ensure that the matrix is fully ranked for all values of $z$
but a finite set. Observing the summands of the determinant, i.e.
$ z^{\sum_{\q^{(\nicefrac{T}{2})} \in \state{\bar{R}}{\nicefrac{T}{2}}}
	\inprod{\bar{\q}{(\nicefrac{T}{2})}}{\sigma(\bar{\q}^{(\nicefrac{T}{2})})}}$,
where $\sigma$ is a permutation on the rows of $\bar{U}$, and noting that $\state{\bar{R}}{\nicefrac{T}{2}}$
is a set of non-negative numbers by definition,
Lemma~\ref{lemma:rearrange} assures us the existence of a strictly maximal
contributor, satisfying the conditions of Lemma~\ref{lemma:poly_full_rank}, thus the set $\left\{ \uu^{\p^{(\nicefrac{T}{2})}} 
\right\}_{\p^{(\nicefrac{T}{2})} \in \state{\bar{R}}{\nicefrac{T}{2}}}$ is linearly
independent.

We prove that the set $\left\{ \vv^{\p^{(\nicefrac{T}{2})}} \in \R^{M^{\nicefrac{T}{2}}}
\right\}_{\p^{(\nicefrac{T}{2})} \in \state{\bar{R}}{\nicefrac{T}{2}}}$ is linearly
independent by arranging it as the columns of the matrix
$V \in \R^{M^{\nicefrac{T}{2}} \times N}$, and showing that its rank equals to $N$.
As in the case of $U$, we select the same sub-set of rows to form the sub-matrix
$\bar{V} \in \R^{N \times N}$. We show that each of the diagonal elements of $\bar{V}$
is a polynomial function whose degree is strictly larger than the degree of all other
elements in its row. As an immediate consequence, the product of the diagonal elements,
i.e. $\prod_{i=1}^N \bar{V}_{ii}(z)$, has degree strictly larger than any other summand
of the determinant $\det(\bar{V})$, and by employing Lemma~\ref{lemma:poly_full_rank},
$\bar{V}$ has full-rank for all values of $z$
but a finite set. The degree of the polynomial function in each entry of $\bar{V}$
is given by:
\begin{align*}
\deg\left(\bar{V}_{\dd,\p^{(\nicefrac{T}{2})}} \right)
&= \max_{\substack{(\p^{(\nicefrac{T}{2}-1)},\ldots,\p^{(1)}) \\ \in \trajectory{\p^{(\nicefrac{T}{2})}}}}
\deg\left( \prod_{r=1}^{{\bar{R}}}{\prod_{j=\nicefrac{T}{2}+1}^T} Z_{r d_j}^{p_r^{(T-j+1)}}\right) \\
&= \max_{\substack{(\p^{(\nicefrac{T}{2}-1)},\ldots,\p^{(1)}) \\ \in \trajectory{\p^{(\nicefrac{T}{2})}}}}
\deg\left( z^{
	\sum_{j=\nicefrac{T}{2}+1}^T \sum_{r=1}^{\bar{R}}
	\Omega^r p^{(T-j+1)}_r \delta_{r d_j}
}\right) \\
&= \max_{\substack{(\p^{(\nicefrac{T}{2}-1)},\ldots,\p^{(1)}) \\ \in \trajectory{\p^{(\nicefrac{T}{2})}}}}
\sum_{j=\nicefrac{T}{2}+1}^T
\Omega^{d_j} p^{(T-j+1)}_{d_j}.
\end{align*}
The above can be formulated as the following combinatorial optimization problem. We are
given an initial state $\p^{(\nicefrac{T}{2})}$ of the bucket of $\nicefrac{T}{2}$ balls
of $\bar{R}$ colors and a sequence of colors $\dd = (d_{\nicefrac{T}{2}+1},\ldots,d_T)$.
At time-step $j$ one ball is taken out of the bucket and yields a reward of
$\Omega^{d_j} p^{(T-j+1)}_{d_j}$, i.e. the number of remaining balls
of color $d_j$ times the weight $\Omega^{d_j}$. Finally,
$\deg(\bar{V}_{\dd, \p^{(\nicefrac{T}{2})}})$ is the accumulated reward of the optimal
strategy of emptying the bucket. In Lemma~\ref{lemma:combinatoric_induction} we prove
that there exists a value of $\Omega$ such that for every sequence of colors $\dd$,
i.e. a row of $\bar{V}$, the maximal reward over all possible initial states is solely attained at the state $\q^{(\nicefrac{T}{2})}$
corresponding to $\dd$, i.e.
$q^{(\nicefrac{T}{2})}_r = \abs{\{j\in\{\nicefrac{T}{2}+1,\ldots,T\}|d_j = r\}}$.
Hence, $\deg(\bar{V}_{ii})$ is indeed strictly larger than the degree of all
other elements in the $i$'th row $\forall i \in [N]$.

Having proved that both $U$ and $V$ have rank $N \equiv \multiset{\bar{R}}{\nicefrac{T}{2}}$ for all values of $z$
but a finite set, we know there exists a value of $z$ for which $\rank{B} =\multiset{\bar{R}}{\nicefrac{T}{2}}$, and the theorem follows.

\hfill $\square$ 

\section{Technical Lemmas and Claims}
\label{app:proofs:main_result:technical}

In this section we prove a series of useful technical lemmas, that we have
employed in our proof for the case of deep RACs, as described in
Appendix~\ref{app:proofs:main_result:deep}.
We begin by quoting a claim regarding the prevalence of the maximal matrix rank
for matrices whose entries are polynomial functions:
\begin{claim} \label{claim:rank_everywhere}
	Let $M, N, K \in \N$, $1 \leq r \leq \min\{M,N\}$ and a polynomial mapping
	$A:\R^K \to \R^{M \times N}$, i.e. for every $i \in [M]$ and $j\in [N]$ it
	holds that $A_{ij}:\R^K \to \R$ is a polynomial function. If there exists a
	point $\x \in \R ^K$ s.t. ${\textrm {rank}}{(A(\x))} \geq r$, then the set
	$\{\x \in \R^K : \textrm{rank}{(A(\x))} < r\}$ has zero measure (w.r.t. the
	Lebesgue measure over $\R^K$).
\end{claim}
\begin{proof}
	See \citep{sharirtractable}.
\end{proof}

Claim~\ref{claim:rank_everywhere} implies that it suffices to show a specific
assignment of the recurrent network weights for which the corresponding grid
tensor matricization achieves a certain rank, in order to show this is a lower
bound on its rank for all configurations of the network weights but a set of
Lebesgue measure zero. Essentially, this means that it is enough to provide a
specific assignment that achieves the required bound in
Theorem~\ref{theorem:main_result} in order to prove the theorem. Next, we show
that for a matrix with entries that are polynomials in $x$, if a single
contributor to the determinant has the highest degree of $x$, then the matrix is
fully ranked for all values of $x$ but a finite set:

\begin{lemma}\label{lemma:poly_full_rank}
	Let $A\in\R^{N\times N}$ be a matrix whose entries are polynomials in
	$x\in\R$. In this case, its determinant may be written as
	$\det(A)=\sum_{\sigma\in S_N}sgn(\sigma)p_\sigma(x)$, where $S_N$ is the
	symmetric group on $N$ elements and $p_\sigma(x)$ are polynomials defined by
	$p_\sigma(x)\equiv\prod_{i=1}^{N} A_{i\sigma(i)}(x),~\forall{\sigma\in S_n}$.
	Additionally, assume there exist $\bar{\sigma}$ such that
	$\deg(p_{\bar{\sigma}}(x)) > \deg(p_{\sigma}(x)) ~ \forall \sigma
	\neq \bar{\sigma}$. Then, for all values of $x$ but a finite set, $A$ is
	fully ranked.
\end{lemma}
\begin{proof}
	We show that in this case $\det(A)$, which is a polynomial in $x$ by its
	definition, is not the zero polynomial. Accordingly, $\det(A)\neq 0$ for all
	values of $x$ but a finite set. Denoting $t\equiv\deg(p_{\bar{\sigma}}(x))$,
	since $t>\deg(p_{\sigma}(x))~\forall\sigma\neq\bar{\sigma}$, a monomial of
	the form $c\cdot x^t,c\in\R \setminus \{0\}$ exists in $p_{\bar{\sigma}}(x)$ and
	doesn't exist in any $p_\sigma(x),~\sigma\neq\bar{\sigma}$. This implies
	that $\det(A)$ is not the zero polynomial, since its leading term has a
	non-vanishing coefficient $sgn(\bar{\sigma})\cdot c\neq 0$, and the lemma
	follows from the basic identity: $\det(A)\neq 0 \iff$ $A$ is fully ranked.
\end{proof}

The above lemma assisted us in confirming that the assignment provided for
the recurrent network weights indeed achieves the required grid tensor
matricization rank of $\multiset{\bar{R}}{T/2}$. The following lemma, establishes a
useful relation we refer to as the \emph{vector rearrangement inequality}:
\begin{lemma}\label{lemma:rearrange}
	Let $\{\vv^{(i)}\}_{i=1}^{N}$ be a set of $N$ different vectors in $\R^{\bar{R}}$
	such that $\forall i\in[N],~j\in[\bar{R}]:~v^{(i)}_j\geq 0$. Then, for all
	$\sigma\in S_N$ such that $\sigma\neq\mathbb{I}_N$, where $S_N$ is the
	symmetric group on $N$, it holds that:
	\begin{equation*}
	\sum_{i=1}^N \inprod{\vv^{(i)}}{\vv^{(\sigma(i))}} < \sum_{i=1}^{N} \norm{\vv^{(i)}}^2.
	\end{equation*}
\end{lemma}
\begin{proof}
	We rely on theorem 368 in~\citep{hardy1952inequalities}, which implies that
	for a set of non-negative numbers $\{a^{(1)},\ldots,a^{(N)}\}$ the following
	holds for all $\sigma\in S_N$:
	\begin{equation}\label{eq:rearrange}
	\sum_{i=1}^{N}a^{(i)}a^{({\sigma(i)})}\leq\sum_{i=1}^{N}(a^{(i)})^2,
	\end{equation}
	with equality obtained only for $\sigma$ which upholds
	$\sigma(i)=j\iff a^{(i)}=a^{(j)}$. The above relation, referred to as the
	\emph{rearrangement inequality}, holds separately for each component
	$j\in[\bar{R}]$ of the given vectors:
	\begin{equation*}
	\sum_{i=1}^{N}v_j^{(i)}v_j^{(\sigma(i))}\leq\sum_{i=1}^{N}(v_j^{(i)})^2.
	\end{equation*}
	We now prove that for all $\sigma\in S_N$ such that
	$\sigma\neq\mathbb{I}_N$, $\exists \hat{j}\in[\bar{R}]$ for which the above inequality
	is hard, \ie:
	\begin{equation}\label{hard_ineq}
	\sum_{i=1}^{N}v_{\hat{j}}^{(i)}v_{\hat{j}}^{(\sigma(i))}<\sum_{i=1}^{N}(v_{\hat{j}}^{(i)})^2.
	\end{equation}
	By contradiction, assume that $\exists\hat{\sigma}\neq\mathbb{I}_N$ for
	which $\forall j \in [\bar{R}]$:
	\begin{equation*}
	\sum_{i=1}^{N}v_j^{(i)}v_j^{(\hat{\sigma}(i))}=\sum_{i=1}^{N}(v_j^{(i)})^2.
	\end{equation*}
	From the conditions of achieving equality in the rearrangement inequality
	defined in Equation~\eqref{eq:rearrange}, it holds that
	$\forall j \in [\bar{R}]:~v_j^{(\hat{\sigma}(i))}= v_j^{(i)}$, trivially
	entailing: $\vv^{(\hat{\sigma}(i))}=\vv^{(i)}$. Thus,
	$\hat{\sigma}\neq\mathbb{I}_N$ would yield a contradiction to
	$\{\vv^{(i)}\}_{i=1}^{N}$ being a set of $N$ different vectors in $\R^{\bar{R}}$.
	Finally, the hard inequality of the lemma for $\sigma\neq\mathbb{I}_N$ is
	implied from Equation~\eqref{hard_ineq}:
	\begin{equation*}
	\sum_{i=1}^N \inprod{\vv^{(i)}}{\vv^{(\sigma(i))}}
	\equiv \sum_{i=1}^N \left(\sum_{j=1}^{\bar{R}} v_j^{(i)} v_j^{(\sigma(i))}\right)
	= \sum_{j=1}^{\bar{R}} \left(\sum_{i=1}^N v_j^{(i)} v_j^{(\sigma(i))}\right)
	< \sum_{j=1}^{\bar{R}} \left(\sum_{i=1}^N (v_j^{(i)})^2 \right)
	= \sum_{i=1}^N \norm{\vv^{(i)}}^2.
	\end{equation*}
\end{proof}

The vector rearrangement inequality in Lemma~\ref{lemma:rearrange}, helped us
ensure that our matrix of interest denoted $\bar{U}$ upholds the conditions of
Lemma~\ref{lemma:poly_full_rank} and is thus fully ranked. Below, we show an identity that allowed us to
make combinatoric sense of a convoluted expression:
\begin{claim}\label{claim:decomp}
	Let $\bar{R}$ and $M$ be positive integers, let $Z \in \R^{\bar{R}\times M}$
	be a matrix, and let $\A$ be a tensor with $T$
	modes, each of dimension $M$, defined by $\A_{d_1,\ldots,d_T} \equiv
	\prod_{t=\nicefrac{T}{2}+1}^{T}\sum_{r=1}^{\bar{R}}\prod_{j=1}^{t}Z_{rd_j}$,
	where $d_1,\ldots,d_T \in [M]$. Then, the following identity holds:
	\begin{align*}
	\A_{d_1,\ldots,d_T} &=
	\sum_{\substack{\p^{(\nicefrac{T}{2})} \\ \in \state{\bar{R}}{\nicefrac{T}{2}}}}
	\sum_{\substack{(\p^{(\nicefrac{T}{2}-1)},\ldots,\p^{(1)}) \\ \in \trajectory{\p^{(\nicefrac{T}{2})}}}}
	\prod_{r=1}^{\bar{R}}
	\left(\prod_{j=1}^{\nicefrac{T}{2}} Z_{r d_j}^{p^{(\nicefrac{T}{2})}_r}\right)
	\left(\prod_{j=\nicefrac{T}{2}+1}^T Z_{r d_j}^{p_r^{(T-j+1)}}\right),
	\end{align*}
	where $\state{\bar{R}}{K} \equiv \{\p^{(K)} \in (\N \cup \{0\})^{\bar{R}} |
	\sum_{i=1}^{\bar{R}} p_i = K\}$,
	and $\trajectory{\p^{(K)}} \equiv \{ (\p^{(K-1)}, \ldots, \p^{(1)}) |
	\forall k \in [K-1], (\p^{(k)} \in \state{\bar{R}}{k} \wedge
	\forall r \in [\bar{R}], p^{(k)}_r \leq p_r^{(k+1)})\}$.
	\footnote{See Appendix~\ref{app:proofs:main_result:deep} for a more intuitive
		definition of the sets $\state{\bar{R}}{K}$ and
		$\trajectory{\p^{(T-k+1)}}$.}
\end{claim}
\begin{proof}
	We will prove the following more general identity by induction. For any
	$k \in [T]$, define $\A^{(k)}_{d_1,\ldots,d_T} \equiv
	\prod_{t=k}^{T}\sum_{r=1}^{\bar{R}}\prod_{j=1}^{t}Z_{rd_j}$, then the
	following identity holds:
	\begin{align*}
	\A^{(k)}_{d_1,\ldots,d_T} &=
	\smashoperator[l]{\sum_{\substack{\p^{(T-k+1)} \\ \in \state{\bar{R}}{T-k+1}}}}
	\sum_{\substack{(\p^{(T-k)},\ldots,\p^{(1)}) \\ \in \trajectory{\p^{(T-k+1)}}}}
	\prod_{r=1}^{\bar{R}}
	\left(\prod_{j=1}^{k-1} Z_{r d_j}^{p^{(T-k+1)}_r}\right)
	\left(\smashoperator[r]{\prod_{j=k}^T} Z_{r d_j}^{p_r^{(T-j+1)}}\right).
	\end{align*}
	
	The above identity coincides with our claim for $k=\nicefrac{T}{2}+1$
	We begin with the base case of $k=T$, for which the set $\state{\bar{R}}{1}$
	simply equals to the unit vectors of $(\N\cup\{0\})^{\bar{R}}$, i.e. for
	each such $\p^{(1)}$ there exists $\bar{r} \in [\bar{R}]$ such that
	$p^{(1)}_r = \delta_{\bar{r} r} \equiv
	\begin{cases} 1 & \bar{r} = r \\ 0 & \bar{r} \neq r \end{cases}$. Thus,
	the following equalities hold:
	\begin{align*}
	\sum_{\p^{(1)} \in \state{\bar{R}}{1}}
	\prod_{r=1}^{\bar{R}} \prod_{j=1}^T Z_{r d_j}^{p^{(1)}_r}
	&= \sum_{\bar{r}=1}^{\bar{R}} \prod_{r=1}^{\bar{R}} \prod_{j=1}^T
	Z_{r d_j}^{\delta_{\bar{r} r}}
	= \sum_{\bar{r}=1}^{\bar{R}} \prod_{j=1}^T Z_{\bar{r} d_j}
	= \A^{(T)}_{d_1,\ldots,d_T}.
	\end{align*}
	
	By induction on $k$, we assume that the claim holds for $\A^{(k+1)}$ and
	prove it on $\A^{(k)}$. First notice that we can rewrite our claim for
	$k < T$ as:
\begin{align}\label{eq:proof:decomp:alternative}
	\A^{(k)}_{d_1,\ldots,d_T} &=
			\sum_{\substack{\p^{(T-k+1)} \\ \in \state{\bar{R}}{T-k+1}}}
	\sum_{\substack{(\p^{(T-k)},\ldots,\p^{(1)}) \\ \in \trajectory{\p^{(T-k+1)}}}}
	\prod_{r=1}^{\bar{R}}
	\left(\prod_{j=1}^{k} Z_{r d_j}^{p^{(T-k+1)}_r}\right)
	\left(\smashoperator[r]{\prod_{j=k+1}^T} Z_{r d_j}^{p_r^{(T-j+1)}}\right),
\end{align}
	where we simply moved the $k$'th term $Z^{p^{(k)}_r}_{r d_k}$ in the right
	product expression to the left product. We can also can rewrite $\A^{(k)}$
	as a recursive formula:
	\begin{align*}
	\A^{(k)}_{d_1,\ldots,d_T} &=
	\left(\sum_{r=1}^{\bar{R}} \prod_{j=1}^k Z_{r d_j}\right)
	\cdot \A^{(k+1)}_{d_1,\ldots,d_T}
	= \left(\sum_{\bar{r}=1}^{\bar{R}} \prod_{r=1}^{\bar{R}} \prod_{j=1}^k
	Z_{r d_j}^{\delta_{\bar{r} r}}\right)
	\cdot \A^{(k+1)}_{d_1,\ldots,d_T}
	\end{align*}.
	Then, employing our induction assumption for $\A^{(k+1)}$, results in:
	\begin{align}
	\A^{(k)}_{d_1,\ldots,d_T}
	&=
	\left(\sum_{\bar{r}=1}^{\bar{R}} \prod_{r=1}^{\bar{R}} \prod_{j=1}^k
	Z_{r d_j}^{\delta_{\bar{r} r}}\right)\quad
	\smashoperator[l]{\sum_{\substack{\p^{(T-k)} \\ \in \state{\bar{R}}{T-k}}}}
	\sum_{\substack{(\p^{(T-k-1)},\ldots,\p^{(1)}) \\ \in \trajectory{\p^{(T-k)}}}}
	\prod_{r=1}^{\bar{R}}
	\left(\prod_{j=1}^k Z_{r d_j}^{p^{(T-k)}_r}\right)
	\left(\smashoperator[r]{\prod_{j=k+1}^T} Z_{r d_j}^{p_r^{(T-j+1)}}\right)
	\nonumber \\
	&=
	\sum_{\bar{r}=1}^{\bar{R}}
	{\sum_{\substack{\p^{(T-k)} \\ \in \state{\bar{R}}{T-k}}}}
	\sum_{\substack{(\p^{(T-k-1)},\ldots,\p^{(1)}) \\ \in \trajectory{\p^{(T-k)}}}}
	\prod_{r=1}^{\bar{R}}
	\left(\prod_{j=1}^k Z_{r d_j}^{p^{(T-k)}_r + \delta_{\bar{r} r}}\right)
	\left(\smashoperator[r]{\prod_{j=k+1}^T} Z_{r d_j}^{p_r^{(T-j+1)}}\right)
	\label{eq:proof:decomp:almost}
	\end{align}
	To prove that the right hand side of Equation~\eqref{eq:proof:decomp:almost} is
	equal to our alternative form of our claim given by
	Equation~\eqref{eq:proof:decomp:alternative}, it is sufficient to show a
	bijective mapping from the terms in the sum of
	Equation~\eqref{eq:proof:decomp:almost}, each specified by a sequence
	$(\bar{r}, \p^{(T-k)}, \ldots, \p^{(1)})$, where $\bar{r} \in [\bar{R}]$,
	$\p^{(T-k)} \in \state{\bar{R}}{T-k}$, and $(\p^{(T-k-1)}, \ldots, \p^{(1)})
	\in \trajectory{\p^{(T-k)}}$, to the terms in the sum of
	Equation~\eqref{eq:proof:decomp:alternative}, each specified by a similar sequence
	$(\p^{(T-k+1)}, \p^{(T-k)}, \ldots, \p^{(1)})$, where $\p^{(T-k+1)} \in
	\state{\bar{R}}{T-k+1}$ and $(\p^{(T-k)},\ldots,\p^{(1)}) \in
	\trajectory{\p^{(T-k+1)}}$.
	
	Let $\phi$ be a mapping such that
	$(\bar{r}, \p^{(T-k)}, \ldots, \p^{(1)}) \overset{\phi}{\mapsto}
	(\p^{(T-k+1)}, \p^{(T-k)}, \ldots, \p^{(1)})$, where
	$p^{(T-k+1)}_r \equiv p^{(T-k)}_r + \delta_{\bar{r} r}$.
	$\phi$ is injective, because if $\phi(\bar{r}_1, \p^{(T-k, 1)}, \ldots,
	\p^{(1, 1)}) = \phi(\bar{r}_2, \p^{(T-k, 2)}, \ldots, \p^{(1, 2)})$ then
	for all $j \in \{1,\ldots,T-k+1\}$ it holds that $\p^{(j,1)} = \p^{(j,2)}$,
	and specifically for $\p^{(T-k+1, 1)} = \p^{(T-k+1, 2)}$ it entails that
	$\delta_{\bar{r}_1 r} = \delta_{\bar{r}_2 r}$, and thus
	$\bar{r}_1 = \bar{r}_2$.
	$\phi$ is surjective, because for any sequence
	$(\p^{(T-k+1)}, \p^{(T-k)}, \ldots, \p^{(1)})$, for which it holds that
	$\forall j, \p^{(j)} \in (\N \cup \{0\})^{\bar{R}}$,
	$\sum_{r=1}^{\bar{R}} p^{(j)}_r = j$, and
	$\forall r, p^{(j)}_r \leq p^{(j+1)}_r$, then it must also holds that
	$p^{(T-k+1)}_r - p^{(T-k)}_r = \delta_{\bar{r} r}$ for some $\bar{r}$,
	since $\sum_{r=1}^{\bar{R}} (p^{(T-k+1)}_r -p^{(T-k)}_r) = (T-k+1)-(T-k) =1$
	and every summand is a non-negative integer.
	
\end{proof}

Finally, Lemma~\ref{lemma:combinatoric_induction} assists us in ensuring that our matrix of interest denoted $\bar{V}$ upholds the conditions of
Lemma~\ref{lemma:poly_full_rank} and is thus fully ranked:
\begin{lemma}\label{lemma:combinatoric_induction}
	Let $\Omega \in \R_+$ be a positive real number.
	For every $\p^{(\nicefrac{T}{2})} \in \state{\bar{R}}{\nicefrac{T}{2}}$
	(see definition in Claim~\ref{claim:decomp}) and every
	$\dd = (d_{\nicefrac{T}{2}+1},\ldots,d_T) \in [\bar{R}]^{\nicefrac{T}{2}}$,
	where $\forall j, d_j \leq d_{j+1}$, we define the following optimization
	problem:
	\begin{equation*}
	f(\dd,\p^{(\nicefrac{T}{2})}) = \max_{\substack{(\p^{(\nicefrac{T}{2}-1)},\ldots,\p^{(1)}) \\ \in \trajectory{\p^{(\nicefrac{T}{2})}}}}
	\sum_{j=\nicefrac{T}{2}+1}^T
	\Omega^{d_j} p^{(T-j+1)}_{d_j},
	\end{equation*}
	where $\trajectory{\p^{(\nicefrac{T}{2})}}$ is defined as in Claim~\ref{claim:decomp}.
	Then, there exists $\Omega$ such that for every such
	$\dd$ the maximal value of $f(\dd,\p^{(\nicefrac{T}{2})})$ over all $\p^{(\nicefrac{T}{2})} \in \state{\bar{R}}{\nicefrac{T}{2}}$ is
	strictly attained at $\hat{\p}^{(\nicefrac{T}{2})}$ defined by
	$\hat{p}^{(\nicefrac{T}{2})}_r = \abs{\{j \in \{\nicefrac{T}{2}+1,\ldots,T\} | d_j = r\}}$.
\end{lemma}
\begin{proof}
	We will prove the lemma by first considering a simple strategy for choosing
	the trajectory for the case of $f(\dd, \hat{\p}^{(\nicefrac{T}{2})})$, achieving
	a certain reward $\rho^*$, and then showing that it is strictly larger than the
	rewards attained for all of the possible trajectories of any other
	$\p^{(\nicefrac{T}{2})} \neq \hat{\p}^{(\nicefrac{T}{2})}$.
	
	Our basic strategy is to always pick the ball of the lowest available color $r$.
	More specifically, if $\hat{p}^{(\nicefrac{T}{2})}_1 > 0$, then in the first
	$\hat{p}^{(\nicefrac{T}{2})}_1$ time-steps we remove balls of the color $1$,
	in the process of which we accept a reward of
	$\Omega^1 \hat{p}^{(\nicefrac{T}{2})}_1$ in the first time-step,
	$\Omega^1 (\hat{p}^{(\nicefrac{T}{2})}_1 - 1)$ in the second time-step,
	and so on to a total reward of
	$\Omega^1 \sum_{i=1}^{\hat{p}^{(\nicefrac{T}{2})}_1} i$.
	Then, we proceed to removing $\hat{p}^{(\nicefrac{T}{2})}_2$ balls of color $2$, and
	so forth. This strategy will result in an accumulated reward of:
	\begin{equation*}
	\rho^* \equiv \sum_{r=1}^{\bar{R}} \Omega^r
	\sum_{i=1}^{\hat{p}^{(\nicefrac{T}{2})}_r} i.
	\end{equation*}
	
	Next, we assume by contradiction that there exists
	$\p^{(\nicefrac{T}{2})} \neq \hat{\p}^{(\nicefrac{T}{2})}$ such that
	$\rho \equiv f(\dd, \p^{(\nicefrac{T}{2})}) \geq \rho^*$. We show by induction that this
	implies $\forall r, p^{(\nicefrac{T}{2})}_r \geq \hat{p}^{(\nicefrac{T}{2})}_r$,
	which would result in a contradiction, since per our assumption $\p^{(\nicefrac{T}{2})} \neq \hat{\p}^{(\nicefrac{T}{2})}$ this means that there is $r$ such
	that $p^{(\nicefrac{T}{2})}_r > \hat{p}^{(\nicefrac{T}{2})}_r$, but since
	$\p^{(\nicefrac{T}{2})}, \hat{\p}^{(\nicefrac{T}{2})} \in \state{\bar{R}}{\nicefrac{T}{2}}$ then the following contradiction arises
	$\nicefrac{T}{2} = \sum_{r=1}^{\bar{R}} p^{(\nicefrac{T}{2})}_r >
	\sum_{r=1}^{\bar{R}} \hat{p}^{(\nicefrac{T}{2})}_r = \nicefrac{T}{2}$.
	More specifically, we show that our assumption entails that for all $r$ starting with $r=\bar{R}$ and
	down to $r=1$, it holds that
	$p^{(\nicefrac{T}{2})}_r \geq \hat{p}^{(\nicefrac{T}{2})}_r$.
	
	Before we begin proving the induction, we choose a value for $\Omega$ that upholds
	$\Omega>(\nicefrac{T}{2})^{2}$ such that the following condition holds: for any $r \in [\bar{R}]$,
	the corresponding weight for the color $r$, i.e. $\Omega^r$, is strictly
	greater than $\Omega^{r-1} (\nicefrac{T}{2})^{2}$. Thus, adding the
	reward of even a single ball of color $r$ is always preferable over
	any possible amount of balls of color $r' < r$.
	
	We begin with the base case of $r=\bar{R}$. If $\hat{p}^{(\nicefrac{T}{2})} = 0$
	the claim is trivially satisfied. Otherwise, we assume by contradiction that
	$p^{(\nicefrac{T}{2})}_{\bar{R}} < \hat{p}^{(\nicefrac{T}{2})}_{\bar{R}}$.
	If $p^{(\nicefrac{T}{2})}_{\bar{R}} = 0$, then the weight of the color
	$\bar{R}$ is not part of the total reward $\rho$, and per our choice of $\Omega$
	it must hold that $\rho < \rho^*$ since $\rho^*$ does include a term of $\Omega^{\bar{R}}$ by definition. Now, we examine the last state of the
	trajectory $\p^{(1)}$, where there is a single ball left in the bucket.
	Per our choice of $\Omega$, if $p^{(1)}_{\bar{R}} = 0$, then once again
	$\rho < \rho^*$, implying that $p^{(1)}_{\bar{R}} = 1$. Following the
	same logic, for $j \in [p^{(\nicefrac{T}{2})}_{\bar{R}}]$, it holds that
	$p^{(j)}_{\bar{R}} = j$. Thus the total contribution of the $\bar{R}$'th
	weight is at most:
	\begin{equation}\label{eq:combinatoric_induction:base}
	\Omega^{\bar{R}} \left((\hat{p}^{(\nicefrac{T}{2})}_{\bar{R}} - p^{(\nicefrac{T}{2})}_{\bar{R}})
	\cdot p^{(\nicefrac{T}{2})}_{\bar{R}}
	+ \sum_{i=1}^{p^{(\nicefrac{T}{2})}_{\bar{R}}} i\right).
	\end{equation}
	This is because before spending
	all of the $p^{(\nicefrac{T}{2})}_{\bar{R}}$ balls of color $\bar{R}$ at the end,
	there are another $(\hat{p}^{(\nicefrac{T}{2})}_{\bar{R}} - p^{(\nicefrac{T}{2})}_{\bar{R}})$
	time-steps at which we add to the reward a value of $p^{(\nicefrac{T}{2})}_{\bar{R}}$.
	However, since Equation~\eqref{eq:combinatoric_induction:base} is strictly less than the corresponding contribution of $\Omega^{\bar{R}}$ in $\rho^*$:
	$\Omega^{\bar{R}} \sum_{i=1}^{\hat{p}^{(\nicefrac{T}{2})}} i$, then it follows that
	$\rho < \rho^*$, in contradiction to our assumption, which implies that to uphold the assumption the following must hold:
	$p^{(\nicefrac{T}{2})}_{\bar{R}} \geq \hat{p}^{(\nicefrac{T}{2})}_{\bar{R}}$, proving the induction base.
	
	Assuming our induction hypothesis holds for all $r' > r$, we show it also holds
	for $r$. Similar to our base case, if $\hat{p}^{(\nicefrac{T}{2})}_r = 0$ then
	our claim is trivially satisfied, and likewise if $p^{(\nicefrac{T}{2})}_r =0$,
	hence it remains to show that the case of
	$p^{(\nicefrac{T}{2})}_{\bar{R}} < \hat{p}^{(\nicefrac{T}{2})}_{\bar{R}}$ is
	not possible. First, according to our hypothesis, $\forall r' > r,
	p^{(\nicefrac{T}{2})}_{r'} \geq \hat{p}^{(\nicefrac{T}{2})}_{r'}$, and per our choice
	of $\Omega$, the contributions to the reward of all of the weights for $r' > r$,
	are at most $\sum_{r'=r+1}^{\bar{R}} \Omega^{r'}
	\sum_{i=1}^{\hat{p}^{(\nicefrac{T}{2})}_{r'}} i$, which is exactly equal to the
	corresponding contributions in $\rho^*$. This means that per our choice of
	$\Omega$ it suffices to show that the contributions originating in the color $r$
	are strictly less than the ones in $\rho^*$ to prove our hypothesis.
	In this optimal setting, the state of the bucket at time-step
	$j = \nicefrac{T}{2} - \sum_{r'=r+1}^{\bar{R}} \hat{p}^{(\nicefrac{T}{2})}_{r'}$
	must upholds $p^{(j)}_{r'} = \hat{p}^{(\nicefrac{T}{2})}_{r'}$ for $r' > r$, and
	zero otherwise. At this point, employing exactly the same logic as in
	our base case, the total contribution to the reward of the weight for the
	$r$'th color is at most:
	\begin{equation}\label{eq:combinatoric_induction:induction}
	\Omega^{r} \left((\hat{p}^{(\nicefrac{T}{2})}_{r} - p^{(\nicefrac{T}{2})}_{r})
	\cdot p^{(\nicefrac{T}{2})}_{r}
	+ \sum_{i=1}^{p^{(\nicefrac{T}{2})}_{r}} i\right),
	\end{equation}
	which is strictly less than the respective contribution in $\rho^*$.
	
\end{proof}



\clearpage
%
%

\chapter{Experiments with Natural Data}
\label{app:natural_data_experiments}
\textit{All experiments presented in this appendix can be reproduced using the source code publicly available in \url{https://github.com/HUJI-Deep/Long-Term-Memory-of-Deep-RNNs}.}
\initial{I}n this appendix, we present experimental evaluations on problems with natural data. This appendix tends to complement the main experiments, presented in Chapter~\ref{chap:experiments}. Though the main experiments systematically demonstrate the trends proved in the theoretical analysis (Chapter \ref{chapter:theory}), they are performed only on synthetic data. Moreover, since the synthetic training data is effectively unlimited, these experiments are lacking any discussion of how depth influence generalization on unseen data.

In order to fill these shortages, we present hereby an additional experiment on a natural dataset of a limited size. This experiment deals with classification of hand-written digits taken from the well known MNIST database (\citep{lecun1998mnist}).

\section{Permuted Pixel-by-Pixel MNIST}\label{sec:exp:perm_mnist}
In the Pixel-by-Pixel MNIST problem, the $28\times 28$ input image is fed sequentially, pixel by pixel, rather than exposing it at once to the predictor as typically done with CNNs. Here the input is essentially a temporal sequence of $784$ time-steps, in each a single pixel is being used as the input. Natural images are characterized with locality of the meaningful features. Consequently, the order in which the pixels are fed to the predictor affects the length of meaningful correlations needed to be captured for a successful learning. In the basic form of the problem the pixels are fed in a raster scan manner, row-by-row from top to bottom, preserving the locality of features, that are now local in time rather than local in space. Conversely, in the harder variant of the task, on which we would focus, a fixed permutation is applied to the dataset in advance, effectively making it a problem of long-term correlations. This variant of the problem is referred to as the Permuted Pixel-by-Pixel MNIST. 

Similarly to the main experiments, we use RMSprop~\citep{tieleman2012lecture} with $\gamma=0.9$, a batch size of $128$ and a learning rate in the range $[1e-5,1e-3]$ where the best of several values is taken. We use a portion of 5K examples taken out from the original training data as our validation set, which leaves us with a training set of 55K examples. We train each configuration for a maximum of 100K iterations, and we use an early stopping criteria defined by the loss on the validation set. The final evaluation is done on the conventional MNIST test data, composed of a held-out set of 10K examples.

The results, presented in Table \ref{table:permuted_mnist}~show that deep RNN architectures are more suitable for learning long-term correlations in natural data than shallow RNNs with the same number of resources. Moreover, the fact that the training data is limited, and the evaluation is done on a held-out test set demonstrates that deeper RNNs also generalize to unseen examples better than shallower ones. These empirical observations complement the main experiments of Chapter \ref{chap:experiments}~, support the main findings of Chapter \ref{chapter:theory}~and extend it to the regime of generalization.

\begin{table*}[t!]
\begin{center}
\caption{Results of the Permuted Pixel-by-Pixel MNIST Task, as presented in Section \ref{sec:exp:perm_mnist}. The rightmost column presents the test accuracy achieved by an scoRNN of depth and width defined by the first and second columns (from left) respectively. The size of each architecture, measured in the number of parameters, appears in the third column. Two results are emphasized in bold, marking deep networks that performed better than shallow networks of a larger size. For example, a \emph{depth-3} scoRNN with 167K parameters achieved a test accuracy of $97.2\%$ while a \emph{depth-1} scoRNN of size 187K achieved only an accuracy of $96.8\%$. Similarly, the accuracy of a \emph{depth-2} scoRNN with 26K parameters was $96.3\%$ while the accuracy of \emph{depth-1} scoRNNs of sizes 36K and 69K was no more than $96.2\%$. \label{table:permuted_mnist}}
~\\
\label{tbl:roc1}
\small{
\begin{tabular}{|c|c|c|c|} 
 \hline
 depth & \#channels & \#params & Test Accuracy \\ [0.5ex] 
 \hline
 2 & 128 & 26k & \ \ \textbf{96.3} \\
 \hline
 1 & 256 & 36K & \ \ 96.1 \\
 \hline
 1 & 360 & 69K & *96.2 \\
 \hline
 1 & 512 & 137K & *96.6 \\
 \hline
 3 & 256 & 167K	& \ \  \textbf{97.2} \\
 \hline
 1 & 600 & 187K & \ \  96.8 \\
 \hline
\end{tabular}
}
\end{center}
\begin{center}
\footnotesize{* as reported in \citep{pmlr-v80-helfrich18a}.}~
\end{center}
\end{table*}
\clearemptydoublepage
%
\backmatter
\bibliographystyle{siam}
\refstepcounter{chapter}
\bibliography{thesisbiblio}

\begin{thebibliography}{10}

\bibitem{amini2012low}
{\sc A.~Amini, A.~Karbasi, and F.~Marvasti}, {\em Low-rank matrix approximation
  using point-wise operators}, IEEE Transactions on Information Theory, 58
  (2012), pp.~302--310.

\bibitem{amodei2016deep}
{\sc D.~Amodei, S.~Ananthanarayanan, R.~Anubhai, J.~Bai, E.~Battenberg,
  C.~Case, J.~Casper, B.~Catanzaro, Q.~Cheng, G.~Chen, et~al.}, {\em Deep
  speech 2: End-to-end speech recognition in english and mandarin}, in
  International Conference on Machine Learning, 2016, pp.~173--182.

\bibitem{arjovsky2016unitary}
{\sc M.~Arjovsky, A.~Shah, and Y.~Bengio}, {\em Unitary evolution recurrent
  neural networks}, in International Conference on Machine Learning, 2016,
  pp.~1120--1128.

\bibitem{bahdanau2014neural}
{\sc D.~Bahdanau, K.~Cho, and Y.~Bengio}, {\em Neural machine translation by
  jointly learning to align and translate}, arXiv preprint arXiv:1409.0473,
  (2014).

\bibitem{beylkin2009multivariate}
{\sc G.~Beylkin, J.~Garcke, and M.~J. Mohlenkamp}, {\em Multivariate regression
  and machine learning with sums of separable functions}, SIAM Journal on
  Scientific Computing, 31 (2009), pp.~1840--1857.

\bibitem{beylkin2002numerical}
{\sc G.~Beylkin and M.~J. Mohlenkamp}, {\em Numerical operator calculus in
  higher dimensions}, Proceedings of the National Academy of Sciences, 99
  (2002), pp.~10246--10251.

\bibitem{brock2016_neuralPhotoEditing}
{\sc A.~Brock, T.~Lim, J.~Ritchie, and N.~Weston}, {\em Neural photo editing
  with introspective adversarial networks}, in 5th International Conference on
  Learning Representations (ICLR), 2017.

\bibitem{cho2014learning}
{\sc K.~Cho, B.~Van~Merri{\"e}nboer, C.~Gulcehre, D.~Bahdanau, F.~Bougares,
  H.~Schwenk, and Y.~Bengio}, {\em Learning phrase representations using rnn
  encoder-decoder for statistical machine translation}, arXiv preprint
  arXiv:1406.1078,  (2014).

\bibitem{cirecsan2010deep}
{\sc D.~C. Cire{\c{s}}an, U.~Meier, L.~M. Gambardella, and J.~Schmidhuber},
  {\em Deep, big, simple neural nets for handwritten digit recognition}, Neural
  computation, 22 (2010), pp.~3207--3220.

\bibitem{cohen2016deep}
{\sc N.~Cohen, O.~Sharir, and A.~Shashua}, {\em Deep simnets}, IEEE Conference
  on Computer Vision and Pattern Recognition (CVPR),  (2016).

\bibitem{cohen2016expressive}
\leavevmode\vrule height 2pt depth -1.6pt width 23pt, {\em On the expressive
  power of deep learning: A tensor analysis}, Conference On Learning Theory
  (COLT),  (2016).

\bibitem{cohen2016convolutional}
{\sc N.~Cohen and A.~Shashua}, {\em Convolutional rectifier networks as
  generalized tensor decompositions}, International Conference on Machine
  Learning (ICML),  (2016).

\bibitem{cohen2017inductive}
\leavevmode\vrule height 2pt depth -1.6pt width 23pt, {\em Inductive bias of
  deep convolutional networks through pooling geometry}, in 5th International
  Conference on Learning Representations (ICLR), 2017.

\bibitem{cohen2017boosting}
{\sc N.~Cohen, R.~Tamari, and A.~Shashua}, {\em Boosting dilated convolutional
  networks with mixed tensor decompositions}, arXiv preprint arXiv:1703.06846,
  (2017).

\bibitem{NIPS2011_4350}
{\sc O.~Delalleau and Y.~Bengio}, {\em Shallow vs. deep sum-product networks},
  in Advances in Neural Information Processing Systems 24, J.~Shawe-Taylor,
  R.~S. Zemel, P.~L. Bartlett, F.~Pereira, and K.~Q. Weinberger, eds., Curran
  Associates, Inc., 2011, pp.~666--674.

\bibitem{elhihi1996}
{\sc S.~El~Hihi and Y.~Bengio}, {\em Hierarchical recurrent neural networks for
  long-term dependencies}, NIPS 8. MIT Press,  (1996).

\bibitem{eldan2016power}
{\sc R.~Eldan and O.~Shamir}, {\em The power of depth for feedforward neural
  networks}, in Conference on Learning Theory, 2016, pp.~907--940.

\bibitem{gers2000recurrent}
{\sc F.~A. Gers and J.~Schmidhuber}, {\em Recurrent nets that time and count},
  in Neural Networks, 2000. IJCNN 2000, Proceedings of the IEEE-INNS-ENNS
  International Joint Conference on, vol.~3, IEEE, 2000, pp.~189--194.

\bibitem{graves2013generating}
{\sc A.~Graves}, {\em Generating sequences with recurrent neural networks},
  arXiv preprint arXiv:1308.0850,  (2013).

\bibitem{graves2009novel}
{\sc A.~Graves, M.~Liwicki, S.~Fern{\'a}ndez, R.~Bertolami, H.~Bunke, and
  J.~Schmidhuber}, {\em A novel connectionist system for unconstrained
  handwriting recognition}, IEEE transactions on pattern analysis and machine
  intelligence, 31 (2009), pp.~855--868.

\bibitem{graves2013speech}
{\sc A.~Graves, A.-r. Mohamed, and G.~Hinton}, {\em Speech recognition with
  deep recurrent neural networks}, in Acoustics, speech and signal processing
  (icassp), 2013 ieee international conference on, IEEE, 2013, pp.~6645--6649.

\bibitem{hackbusch2006efficient}
{\sc W.~Hackbusch}, {\em On the efficient evaluation of coalescence integrals
  in population balance models}, Computing, 78 (2006), pp.~145--159.

\bibitem{hackbusch2012tensor}
\leavevmode\vrule height 2pt depth -1.6pt width 23pt, {\em Tensor spaces and
  numerical tensor calculus}, vol.~42, Springer Science \& Business Media,
  2012.

\bibitem{hardy1952inequalities}
{\sc G.~H. Hardy, J.~E. Littlewood, and G.~P{\'o}lya}, {\em Inequalities},
  Cambridge university press, 1952.

\bibitem{harrison2003multiresolution}
{\sc R.~J. Harrison, G.~I. Fann, T.~Yanai, and G.~Beylkin}, {\em
  Multiresolution quantum chemistry in multiwavelet bases}, in Computational
  Science-ICCS 2003, Springer, 2003, pp.~103--110.

\bibitem{he2016deep}
{\sc K.~He, X.~Zhang, S.~Ren, and J.~Sun}, {\em Deep residual learning for
  image recognition}, in Proceedings of the IEEE Conference on Computer Vision
  and Pattern Recognition, 2016, pp.~770--778.

\bibitem{pmlr-v80-helfrich18a}
{\sc K.~Helfrich, D.~Willmott, and Q.~Ye}, {\em Orthogonal recurrent neural
  networks with scaled {C}ayley transform}, in Proceedings of the 35th
  International Conference on Machine Learning, J.~Dy and A.~Krause, eds.,
  vol.~80 of Proceedings of Machine Learning Research, Stockholmsmassan,
  Stockholm Sweden, 10--15 Jul 2018, PMLR, pp.~1969--1978.

\bibitem{henaff_OrthogonalRNNs}
{\sc M.~Henaff, A.~Szlam, and Y.~LeCun}, {\em Orthogonal rnns and long- memory
  tasks}, arXiv preprint arXiv:1602.06662,  (2016).

\bibitem{hermans2013training}
{\sc M.~Hermans and B.~Schrauwen}, {\em Training and analysing deep recurrent
  neural networks}, in Advances in Neural Information Processing Systems, 2013,
  pp.~190--198.

\bibitem{hochreiter1997long}
{\sc S.~Hochreiter and J.~Schmidhuber}, {\em Long short-term memory}, Neural
  computation, 9 (1997), pp.~1735--1780.

\bibitem{jing2016tunable}
{\sc L.~Jing, Y.~Shen, T.~Dub{\v{c}}ek, J.~Peurifoy, S.~Skirlo, Y.~LeCun,
  M.~Tegmark, and M.~Solja{\v{c}}i{\'c}}, {\em Tunable efficient unitary neural
  networks (eunn) and their application to rnns}, arXiv preprint
  arXiv:1612.05231,  (2016).

\bibitem{khrulkov2018expressive}
{\sc V.~Khrulkov, A.~Novikov, and I.~Oseledets}, {\em Expressive power of
  recurrent neural networks}, in 6th International Conference on Learning
  Representations (ICLR), 2018.

\bibitem{Krizhevsky:2012wl}
{\sc A.~Krizhevsky, I.~Sutskever, and G.~E. Hinton}, {\em {ImageNet
  Classification with Deep Convolutional Neural Networks.}}, Advances in Neural
  Information Processing Systems,  (2012), pp.~1106--1114.

\bibitem{lecun1998mnist}
{\sc Y.~LeCun, C.~Cortes, and C.~J. Burges}, {\em The mnist database of
  handwritten digits}, 1998.

\bibitem{RAC_paper}
{\sc Y.~Levine, O.~Sharir, A.~Ziv, and A.~Shashua}, {\em On the long-term
  memory of deep recurrent networks}, arXiv preprint arXiv:1710.09431,  (2017).

\bibitem{levine2018deep}
{\sc Y.~Levine, D.~Yakira, N.~Cohen, and A.~Shashua}, {\em Deep learning and
  quantum entanglement: Fundamental connections with implications to network
  design}, in 6th International Conference on Learning Representations (ICLR),
  2018.

\bibitem{martens2011learning}
{\sc J.~Martens and I.~Sutskever}, {\em Learning recurrent neural networks with
  hessian-free optimization}, in Proceedings of the 28th International
  Conference on Machine Learning (ICML-11), Citeseer, 2011, pp.~1033--1040.

\bibitem{mohamed2012acoustic}
{\sc A.-r. Mohamed, G.~E. Dahl, and G.~Hinton}, {\em Acoustic modeling using
  deep belief networks}, IEEE Transactions on Audio, Speech, and Language
  Processing, 20 (2012), pp.~14--22.

\bibitem{nair2010rectified}
{\sc V.~Nair and G.~E. Hinton}, {\em Rectified linear units improve restricted
  boltzmann machines}, in Proceedings of the 27th International Conference on
  Machine Learning (ICML-10), 2010, pp.~807--814.

\bibitem{novikov2015tensorizingNN}
{\sc A.~Novikov, D.~Podoprikhin, A.~Osokin, and D.~P. Vetrov}, {\em Tensorizing
  neural networks}, in Advances in Neural Information Processing Systems 28,
  C.~Cortes, N.~D. Lawrence, D.~D. Lee, M.~Sugiyama, and R.~Garnett, eds.,
  Curran Associates, Inc., 2015, pp.~442--450.

\bibitem{orus2014practical}
{\sc R.~Or{\'u}s}, {\em A practical introduction to tensor networks: Matrix
  product states and projected entangled pair states}, Annals of Physics, 349
  (2014), pp.~117--158.

\bibitem{oseledets2011tensor}
{\sc I.~V. Oseledets}, {\em Tensor-train decomposition}, SIAM Journal on
  Scientific Computing, 33 (2011), pp.~2295--2317.

\bibitem{pascanu2013construct}
{\sc R.~Pascanu, C.~Gulcehre, K.~Cho, and Y.~Bengio}, {\em How to construct
  deep recurrent neural networks}, arXiv preprint arXiv:1312.6026,  (2013).

\bibitem{pascanu2013difficulty}
{\sc R.~Pascanu, T.~Mikolov, and Y.~Bengio}, {\em On the difficulty of training
  recurrent neural networks}, in International Conference on Machine Learning,
  2013, pp.~1310--1318.

\bibitem{poon2011sum}
{\sc H.~Poon and P.~Domingos}, {\em Sum-product networks: A new deep
  architecture}, in Computer Vision Workshops (ICCV Workshops), 2011 IEEE
  International Conference on, IEEE, 2011, pp.~689--690.

\bibitem{schmidhuber1992learning}
{\sc J.~H. Schmidhuber}, {\em Learning complex, extended sequences using the
  principle of history compression.}, Neural Computation,  (1992).

\bibitem{sharir2018expressive}
{\sc O.~Sharir and A.~Shashua}, {\em On the expressive power of overlapping
  architectures of deep learning}, in 6th International Conference on Learning
  Representations (ICLR), 2018.

\bibitem{sharirtractable}
{\sc O.~Sharir, R.~Tamari, N.~Cohen, and A.~Shashua}, {\em Tractable generative
  convolutional arithmetic circuits},  (2016).

\bibitem{NIPS2016_6211}
{\sc E.~Stoudenmire and D.~J. Schwab}, {\em Supervised learning with tensor
  networks}, in Advances in Neural Information Processing Systems 29, D.~D.
  Lee, M.~Sugiyama, U.~V. Luxburg, I.~Guyon, and R.~Garnett, eds., Curran
  Associates, Inc., 2016, pp.~4799--4807.

\bibitem{sutskever2011generating}
{\sc I.~Sutskever, J.~Martens, and G.~E. Hinton}, {\em Generating text with
  recurrent neural networks}, in Proceedings of the 28th International
  Conference on Machine Learning (ICML-11), 2011, pp.~1017--1024.

\bibitem{tagare2011}
{\sc D.~H. Tagare}, {\em Notes on optimization on stiefel manifolds}, Yale
  University, New Haven,  (2011).

\bibitem{telgarsky2015representation}
{\sc M.~Telgarsky}, {\em Representation benefits of deep feedforward networks},
  arXiv preprint arXiv:1509.08101,  (2015).

\bibitem{tieleman2012lecture}
{\sc T.~Tieleman and G.~Hinton}, {\em Lecture 6.5-rmsprop: Divide the gradient
  by a running average of its recent magnitude}, COURSERA: Neural networks for
  machine learning, 4 (2012), pp.~26--31.

\bibitem{leSimpleWayToInitialize}
{\sc Q.~V.~Le, N.~Jaitly, and G.~E.~Hinton}, {\em A simple way to initialize
  recurrent networks of rectified linear units}, arXiv preprint
  arXiv:1504.00941,  (2015).

\bibitem{wisdom2016full}
{\sc S.~Wisdom, T.~Powers, J.~Hershey, J.~Le~Roux, and L.~Atlas}, {\em
  Full-capacity unitary recurrent neural networks}, in Advances in Neural
  Information Processing Systems, 2016, pp.~4880--4888.

\bibitem{wu2016multiplicative}
{\sc Y.~Wu, S.~Zhang, Y.~Zhang, Y.~Bengio, and R.~R. Salakhutdinov}, {\em On
  multiplicative integration with recurrent neural networks}, in Advances in
  Neural Information Processing Systems, 2016, pp.~2856--2864.

\bibitem{zhang2016architectural}
{\sc S.~Zhang, Y.~Wu, T.~Che, Z.~Lin, R.~Memisevic, R.~R. Salakhutdinov, and
  Y.~Bengio}, {\em Architectural complexity measures of recurrent neural
  networks}, in Advances in Neural Information Processing Systems, 2016,
  pp.~1822--1830.

\end{thebibliography}
\clearemptydoublepage
%
%
\end{document}